\documentclass{article}
\usepackage[preprint]{custom_2020}
\usepackage{amsmath}
\usepackage{stmaryrd}
\usepackage{tgtermes}

\usepackage[dvipsnames]{xcolor}

\usepackage{bm}
\usepackage{endnotes}

\usepackage{amssymb,amsfonts}

\usepackage{caption}
\usepackage{subcaption}
\usepackage{graphicx}

\usepackage{url}
\usepackage{enumerate}
\usepackage{multicol}
\usepackage{multirow}
\usepackage{mathtools}
\usepackage{tabularray}
\usepackage{cals}
\usepackage{algorithm}
\usepackage{algpseudocode}
\usepackage{tikz}
\usepackage{color}
\usepackage{array}
\usepackage{multirow}
\usepackage{eufrak}
\usepackage{comment}
\usepackage{tikz}
\usetikzlibrary{arrows.meta, positioning}
\usetikzlibrary{shapes}
\usetikzlibrary{calc}

\DeclareMathOperator{\vc}{vec}

\DeclareMathOperator{\card}{card}

\DeclareMathOperator{\diag}{diag}
\DeclareMathOperator{\ReLU}{ReLU}

\newcommand{\blue}[1]{{\textcolor{black}{#1}}}
\newcommand{\Red}[1]{{\textcolor{black}{#1}}}

\newcommand{\mathbbm}[1]{\text{\usefont{U}{bbm}{m}{n}#1}} % from mathbbm.sty
\usepackage{natbib}
 \bibpunct[, ]{(}{)}{,}{a}{}{,}%

\usepackage{amsthm}
\theoremstyle{plain}

\newtheorem{lemma}{Lemma}

\newtheorem{proposition}{Proposition}

\newtheorem{assumption}{Assumption}
\newtheorem{example}{Example}

\title{Wasserstein Distributionally Robust Shallow Convex Neural Networks}

\author{
Julien Pallage\\
Department of Electrical Engineering\\
Polytechnique Montréal, GERAD, \& Mila \\
\And
Antoine Lesage-Landry\\
Department of Electrical Engineering\\
Polytechnique Montréal, GERAD, \& Mila\\
}

\begin{document}
%%%%%%%%%%%%%%%%
\maketitle

\begin{abstract}
% Enter your abstract
In this work, we propose Wasserstein distributionally robust shallow convex neural networks (WaDiRo-SCNNs) to provide reliable nonlinear predictions when subject to adverse and corrupted datasets. Our approach is based on the reformulation of a new convex training program for $\ReLU$\blue{-based} shallow neural networks, which allows us to cast the problem into the order-1 Wasserstein distributionally robust optimization framework. Our training procedure is conservative, has low stochasticity, is solvable with open-source solvers, and is scalable to large industrial deployments. We provide out-of-sample performance guarantees, show that hard convex physical constraints can be enforced in the training program, and propose a mixed-integer convex post-training verification program to evaluate model stability. WaDiRo-SCNN aims to make neural networks safer for critical applications, such as in the energy sector. Finally, we numerically demonstrate our model's performance through both a synthetic experiment and a real-world power system application, viz., the prediction of hourly energy consumption in non-residential buildings within the context of virtual power plants, and evaluate its stability across standard regression benchmark datasets. The experimental results are convincing and showcase the strengths of the proposed model.

\end{abstract}

%%%%%%%%%%%%%%%%%%%%%%%%%%%%%%%%%%%%%%%%%%%%%%%%%%%%%%%%%%%%%%%%%%%%%%

% Text of your paper here

\section{Introduction}\label{sec:intro}
\color{black}
Consider a general feedforward neural network (NN) training problem for a network of $h\in \mathbb{N}$ hidden layers under a loss function $\mathcal{L}:\mathbb{R}^{N \times m_{h+1} }\times \mathbb{R}^{N \times m_{h+1}}\to \mathbb{R}$ which penalizes the difference between the $N\in \mathbb{N}$ training labels and NN predictions \Red{of dimension $m_{h+1}$, the number of hidden neurons in the output layer}. Let $\mathbf{x}_j \in \mathbb{R}^{m_0}, \mathbf{y}_j \in \mathbb{R}^{m_{h+1}}, \ \text{and} \ \hat{\mathbf{y}}_j \in \mathbb{R}^{m_{h+1}}$ define training features, training labels, and training predictions, respectively, $\forall j \in \Red{\{1, 2, 3, \ldots, N \} \eqqcolon} \llbracket N \rrbracket$. Matrices $\mathbf{Y}, \hat{\mathbf{Y}}\in \mathbb{R}^{N \times m_{h+1}}$ denote the collection of all labels and predictions. The problem is as follows:
\begin{alignat*}{3}
    &\min_{\mathbf{W}_1, \mathbf{W}_2, \ldots, \mathbf{W}_h, \mathbf{b}_1, \mathbf{b}_2, \ldots, \mathbf{b}_h } && \mathcal{L}\left(  \hat{\mathbf{Y}}, \mathbf{Y}\right) \tag{\texttt{NNT}}\label{eq:NNT}\\
    &\qquad \quad  \;\;\;\, \text{s.t.} &&\hat{\mathbf{y}}_j =  \phi^h( \mathbf{W}_h^\top(\ldots \phi^2(\mathbf{W}_2^\top\phi^1(\mathbf{W}_1^\top \mathbf{x}_j + \mathbf{b}_1) + \mathbf{b}_2) \ldots) + \mathbf{b}_h ) \quad && \forall j \in \llbracket N \rrbracket,
\end{alignat*}
where $\Red{\phi^i} : \mathbb{R}^{m_i} \to \mathbb{R}^{m_{i+1}}$ is the activation function of the hidden layer $i$, $m_i\in \Red{\mathbb{N}_{>0}}$ is the number of neurons in layer $i$, $\mathbf{W}_i \in \mathbb{R}^{m_{i-1} \times m_i}$ are the weights in layer $i$, and $\mathbf{b}_i \in \mathbb{R}^{m_i}$ are the bias weights of layer $i$ for all $i \in \llbracket h \rrbracket $. In our notation, we assume that NN inputs and outputs are vectors. Let $\mathbf{0}_{m_i}$ be the null vector of dimension $m_i$, a common activation function \Red{$\phi^i$} for regression is the element-wise rectified linear unit ($\ReLU$) operator: $$\ReLU (\cdot) = \max\{\mathbf{0}_{m_i}, \cdot\},$$ as it is accurate when modelling nonlinear patterns and biomimics the required excitation threshold for brain neurons to \textit{activate}\Red{~\citep{householder1941theory}}. The acquisition of features is done through the input layer \Red{while the resulting labels are obtained through the output layer}. \Red{Thus,} a neural network of $h$ hidden layers has effectively $h+2$ layers. To solve \eqref{eq:NNT}, the \Red{gradient of the} loss is propagated iteratively, one layer at a time, from the output neurons to the input neurons, by computing gradient estimators of the NN architecture at each layer and updating weights with methods derived from non-convex optimization, e.g., stochastic gradient descent (SGD) \citep{amari1993backpropagation}. This procedure is referred to as backpropagation \citep{bishop2023deep}.

In part because~\eqref{eq:NNT} is typically non-convex and only solvable through stochastic heuristics with many hyperparameters, NNs have traditionally been disregarded in critical sectors, e.g., energy~\citep{Chatzivasileiadis2022trust}, healthcare~\citep{rasheed2022explainable}, and finance~\citep{GIUDICI2023104088}. Even though they tend to be successful nonlinear predictors, their lack of interpretability, limited out-of-the-box performance guarantees, significant data requirements, as well as high susceptibility to data corruption and other adversarial attacks, have labelled them as unreliable for such applications.  

In the past decade, many solutions have been proposed to guide and certify NN training as much as possible to mitigate these inherent flaws before deployment. 

\color{black}
\blue{As a first example,} blackbox optimization (BBO) algorithms with provable convergence properties \blue{can provide some local convergence guarantees in the} hyperparameter tuning \blue{phase}~\citep{articleBBO, kawaguchi2021recipe} \blue{because NNs' loss landscape, with respect to their hyperparameters, can be fractal~\citep{sohl2024boundary}. However, when paired with highly stochastic training programs, their relevance is diminished. }
\color{black}

Because their high complexity and opacity make them unpredictable, complete performance assessment is hard to obtain during training. As such, post-training verification frameworks have emerged as a way to certify formal guarantees on trained machine learning (ML) models \citep{huang2017safety}. These procedures are conceptually similar to quality control tests in factories. By using a model's trained weights as parameters and representing the model's internal mechanisms in an optimization problem, these frameworks aim to find theoretical worst-case scenarios, e.g., worst-case constraint violation in a specific critical application \citep{venzke2020WorstCase}, worst-case instability \citep{Huang2021_lipschitzNN}, and more \citep{albarghouthi2021introduction}. A general feedforward neural network \Red{post-training} verification \Red{(PTV)} problem takes the form:
\begin{alignat*}{3}
    &\max_{\mathbf{x}, \hat{\mathbf{y}}, \boldsymbol{\sigma}  } \ && \mathcal{V}\left(\hat{\mathbf{y}}, \mathbf{x}, \boldsymbol{\sigma} \right) \tag{\texttt{PTV}}\label{eq:PTV}\\
    &\;\;\text{s.t.} &&\hat{\mathbf{y}} =  \phi^h( \mathbf{W}_h^\top(\ldots \phi^2(\mathbf{W}_2^\top\phi^1(\mathbf{W}_1^\top \mathbf{x} + \mathbf{b}_1) + \mathbf{b}_2) \ldots) + \mathbf{b}_h ) \\
    & && f_i(\hat{\mathbf{y}}, \mathbf{x}, \boldsymbol{\sigma}) \leq 0 && \forall i \in \llbracket n \rrbracket\\
    & && g_i(\hat{\mathbf{y}}, \mathbf{x}, \boldsymbol{\sigma}) = 0 && \forall i \in \llbracket l \rrbracket\\
    & &&\mathbf{x} \in \mathcal{X},
\end{alignat*}
where $\mathbf{x}\in \mathcal{X}$ is any possible input from the (bounded) feature space $\mathcal{X}\subseteq\mathbb{R}^{m_0}$, $\mathcal{V}:\mathbb{R} \times \mathbb{R}^{m_{h+1}} \times \mathbb{R}^{\card (\boldsymbol{\sigma})} \to \mathbb{R}$ is an objective function which measures a specific guarantee, $\boldsymbol{\sigma}$ represents all of the problem's other variables used in the verification process, and $f_1, f_2, \dots, f_n, g_1, g_2, \dots, g_l$ are the family of inequality and equality constraints needed for the certification. Note that a convex representation of the internal NN mechanisms, with respect to the inputs, is desirable for such problems because the convergence guarantees to global optimality of convex optimization eliminate any doubt regarding the validity of the verification.

When a neural network is used in a dynamical setting, in which a system evolves over time on a state-space and follows a set of physical rules \citep{wiggins2003systems}, the introduction of physical knowledge from the setting into its training procedure can increase its overall reliability and greatly reduce the data requirements to obtain good empirical performances~\citep{Karniadakis2021, xu2023physics}. This type of training is often referred to as physics-informed, physics-guided, or physics-constrained depending on how physical knowledge is embedded in the training process. We refer interested readers to \cite{cuomo2022scientific} for an overview of different techniques. We will cover the most common one in Section~\ref{subsec:related}.

Typically, as it is hard to guarantee NN out-of-sample performance over time and their performance may degrade (drift), the aforementioned techniques are combined in complex pre-deployment and post-deployment loops~\citep{stiasny2022closing} with tight MLOps (contraction of machine learning, software development, and operations) procedures for lifecycle management~\citep{kreuzberger2023mlops}. \Red{Extensive} MLOps practices help monitor and assert, theoretically and empirically, the quality of ML models from training to deployment, but \Red{require substantial} computational infrastructures to be well implemented. Such infrastructures are not democratized in every sector, especially energy and healthcare. We present an example of a safe MLOps pipeline inspired by~\cite{stiasny2022closing} in Figure \ref{fig:deployMlOps}. A safe MLOps pipeline can be divided into three parts: (i) definition of initial conditions, (ii) hyperparameter optimization, and (iii) model verification. Initial conditions englobe the choice of dataset, ML model, hyperparameter tuning algorithm, and the definition of the hyperparameter search space. Hyperparameter optimization covers the typical training and validation phases: hyperparameters are chosen for each run to pre-process the training dataset and train the ML model, the trained model is then evaluated on the validation dataset, and different metrics are logged. This part ends when a stopping criterion is met, e.g., the maximum number of runs is reached. Finally, the best models in the validation phase are verified through post-training certification methods and evaluated on the testing dataset. If the performance expectations are not met by any model, initial conditions must be redefined. Otherwise, the best ML model can be deployed. 

Most of the computational burden stems from hyperparameter optimization. \Red{As such, there is a growing interest in designing complex predictive models with low-stochasticity training procedures, e.g., involving fewer hyperparameters and exhibiting convexity with respect to the weights.} \Red{There is also an interest in designing training procedures that can withstand imperfect training data and thus have} lower data pre-processing requirements, e.g., by being distributionally robust. These models simplify the hyperparameter optimization phase and are more likely than standard models to meet the requirements of the verification phase.
\color{black}
\begin{figure}[t]
    \centering

\resizebox{\textwidth}{!}{
\begin{tikzpicture}[
    node distance=2cm,
    every node/.style={ rounded corners, align=center, text width=8em,},
    decision/.style={diamond, draw, text width=3em, align=center},
    arrow/.style={-{Stealth[scale=1.2]}, dotted}
]
\node[draw,] (dataset) {Dataset};
\node[draw,above=2cm of dataset] (modelSel) {Model selection};
\node[draw,above=2cm of modelSel] (hyperparamSel) {Hyperparameter tuning algorithm\\ and search space};

\node[draw,below right=0.1cm and 2cm of dataset] (featureSel) {Feature selection};
\node[draw,below=0.5cm of featureSel] (informedSamp) {Informed sampling};
\node[draw,above=0.5cm of featureSel] (outlierFilt) {Outlier filtering};
\node[draw,above=0.5cm of outlierFilt] (scaling) {Scaling};
\node[below=0.6cm of informedSamp, black] (dataprep) {Data preprocessing};

\node[draw,above right=0.1cm and 2cm of featureSel] (train) {Training };
\node[draw,right=of train] (valid) {Validation };
\node[draw,above=1.5of valid ](logging) {Log run};
\node[draw,above=3.95cm of train ](hyperparam) {Hyperparameter selection};
\node[draw,above=1.5cm of logging, diamond, aspect=2, text width=6em](stop) {Stop?};
\draw[] node[above = 0.05 cm of stop.west, label distance = 0mm,inner sep=3pt] {no};
\draw[] node[above = 0.05 cm of stop.east, label distance = 0mm,inner sep=3pt] {yes};

\node[draw,below right=0.1cm and 2cm of valid] (test) {Testing};
\node[draw,above=0.5cm of test] (constrVio) {Theoretical certifications};
\node[below=0.6cm of test, black,text width=12em] (verify) {(iii) Model verification};
\node[draw,above right=-0.2cm and 3cm of test, diamond, aspect=2, text width=6em] (yayNay) {Good enough?};
\draw[] node[above right= -0.05cm and -1.4cm  of yayNay.north, label distance = 0mm,inner sep=3pt] {no};
\draw[] node[above = 0.05 cm  of yayNay.east, label distance = 0mm,inner sep=3pt] {yes};
\node[draw,right=of yayNay, ForestGreen] (deploy) {Deploy model};
\node[draw, above=2cm of yayNay, red, text width=12em] (restart) {Redefine initial conditions};

\draw[arrow] (dataset.east)  to [out=0,in=180] (scaling.west) ;
\draw[arrow] (dataset.east)  to [out=0,in=180] (outlierFilt.west);
\draw[arrow] (dataset.east)  to [out=0,in=180](informedSamp.west);
\draw[arrow] (dataset.east)  to [out=0,in=180] (featureSel.west);

\draw[arrow] (scaling.east) to [out=0,in=180](train.west);
\draw[arrow] (outlierFilt.east)  to [out=0,in=180] (train.west);
\draw[arrow] (informedSamp.east)  to [out=0,in=180] (train.west);
\draw[arrow] (featureSel.east)  to [out=0,in=180] (train.west);
\draw[arrow] (train.east) -- (valid.west) node[midway, above] {model};
\draw[arrow] (valid.north) -- (logging.south) node[midway, ] {val. loss};

\draw[arrow] (logging.north) -- (stop.south);
\draw[arrow] (stop.west) -- (hyperparam.east);
\draw[arrow] (yayNay.north) -- (restart.south);
\draw[arrow] (hyperparam.south) -- (train.north) node[midway, above=0.5cm, text width=12em,] {new training\\ hyperparameters};
\draw[arrow] (stop.east) -- ($(constrVio.north) + (0,0.5)$) node[midway, text width=12em,] {best model \\in validation};
\draw[arrow] (hyperparam.west) -- ($(scaling.north) + (0,0.5)$) node[midway, above=0.1cm, text width=12em,] {new data\\ hyperparameters};
%\draw [arrow,black, dashed] (yayNay.north) to [out=90,in=30] (modelSel.north) ;
%\draw [arrow,black, dashed] (yayNay.north) to [out=90,in=30] (hyperparamSel.north);
\draw[arrow] (constrVio.east)  to [out=0,in=180] (yayNay.west);
\draw[arrow] (test.east)  to [out=0,in=180] (yayNay.west);
\draw[arrow] (yayNay.east) -- (deploy.west);
\draw[arrow] (hyperparamSel.east) to [out=45,in=135] (hyperparam.north) ;
\draw[arrow] (modelSel.east) to [out=30,in=90] (train.north);

\draw[black,thick, rounded corners] ($(scaling.north west)+(-1,0.5)$)  rectangle ($(informedSamp.south east)+(1,-0.5)$);
\draw[black,thick, rounded corners] ($(constrVio.north west)+(-1,0.5)$)  rectangle ($(test.south east)+(1,-0.5)$);
\draw[black,thick, rounded corners] ($(stop.north east)+(1.5,0.7)$)  rectangle ($(dataprep.south west)+(-1.2,-0.5)$) node[midway, below=5.2cm, text width=16em,] {(ii) Hyperparameter optimization};
\draw[black,thick, rounded corners] ($(hyperparamSel.north east)+(0.5,0.5)$)  rectangle ($(dataset.south west)+(-0.5,-0.5)$) node[midway, below=4.2cm, text width=12em,] {(i) Initial conditions};
\end{tikzpicture}
}

    \caption{Example of safe pre-deployment MLOps pipeline}
    \label{fig:deployMlOps}
\end{figure}
\color{black}
In this work, our motivation is to design a nonlinear predictive ML model with a simple low-stochasticity training program that can be guided by physical constraints and that has inherent performance guarantees. We also aim to design this model with an architecture that can be easily included in post-training verification frameworks or data-driven control applications, \Red{if required}. Shallow convex neural networks, as we will show, are \Red{well-suited} for the task. In doing so, we hope to lower computational infrastructure needs for safe deployment in large-scale critical sectors where a lack of expertise and budget constraints do not permit state-of-the-art MLOps pipelines at all stages, and yet accurate nonlinear predictions are required. We specifically target virtual power plants (VPPs)~\citep{pudjianto2007virtual} applications where reliable forecasts of distributed assets are crucial for the safe operations of the electrical power grid. 
\color{black}

\blue{As an example of a standard VPP application, we utilize our model to} predict the \blue{hourly energy consumption} of non-residential buildings enrolled in demand response programs \citep{siano2014demand}. Because of their \blue{significant impact on distribution grids and their} diverse activities, non-residential buildings have complex \blue{nonlinear} consumption patterns that \blue{must be predicted accurately to enhance grid operation \citep{MASSANA2015322}}. As each building needs its model \blue{and thus possibly thousands of models must be deployed,} scalability issues arise when utilizing complex \blue{stochastic} ML training procedures \blue{in complete safe MLOps pipelines}. Finally, the sensitivity of the application, and the risk associated with poor predictions, promote models with theoretical guarantees, \blue{mechanisms to respect physical constraints}, and robustness to possible anomalies in the training data, e.g., numerical errors, noise in sensor readings, telemetry issues, meters outage, and unusual extreme events~\citep{s21175834}. 
\color{black}
\subsection{Notation}
We use the following notation throughout the paper:

\begin{itemize}
    \item Bold lowercase letters, e.g., $\mathbf{a}$, refer to vectors, bold uppercase letters, e.g., $\mathbf{A}$, refer to matrices, and normal lowercase or uppercase letters refer to scalars, e.g., $a$, or $A$. Bold Greek letters may refer to matrices depending on context.
    \item The sets of real, natural, and integer numbers are denoted by $\mathbb{R}$, $\mathbb{N}$, and $\mathbb{Z}$, respectively. \Red{We add the subscript $\{\cdot\}_{>0}$, e.g., $\mathbb{N}_{>0}$, to represent the respective subset of strictly positive numbers.}
    \item \Red{We use the following combinations of font style and letters for distributions: $\mathbbm{D}, \mathbbm{F}, \mathbbm{P}, \mathbbm{Q}, \mathbbm{U}$, and $\mathbbm{V}$.}
    \item Calligraphic uppercase letters refer to sets, e.g., $\mathcal{X}$ and $ \mathcal{Z}$, or to functions depending on context, e.g., $\mathcal{L}(\cdot)$ and $ \mathcal{V}(\cdot)$.
    \item Letters $\mathbf{x},\ \mathbf{y}, \text{ and } \mathbf{z}$ are reserved for features, labels, and samples, respectively, while $\hat{\mathbf{y}}$ refers to model predictions.
    \item Let $\mathbf{a}\in\mathbb{R}^n$ be a general vector of dimension $n\in\mathbb{N}_{>0}$, column vectors are compactly written $\mathbf{a} = \left( a_1, a_2, \ldots, a_n\right)$ and row vectors as $\mathbf{a}^\top = \left( a_1, a_2, \ldots, a_n\right)^\top = \left( a_1\ a_2\ \ldots \ a_n\right)$.
    \item Bracket notation $\llbracket \cdot \rrbracket$ is used as a concise alternative to $\{1,2, \ldots, \cdot \}$. 
    \item $\{\cdot\}\succeq 0$ refers to semidefinite positiveness if $\{\cdot\}$ is a matrix and is an element-wise $\geq 0$ if $\{\cdot\}$ is a vector.
    \item $\mathbbm{1}(\cdot) : \mathbb{R}^n \to \{0,1\}^{n}$ is an element-wise indicator function of arbitrary dimension $n$ which returns 1 or 0 element-wise if the condition is true or false, respectively, for each separate element.
    \item $\vc (\cdot) : \mathbb{R}^{m\times n} \to \mathbb{R}^{mn}$ denotes the vectorization operator which flattens matrices. If $\mathbf{A} = (\mathbf{a}_1 \ \mathbf{a_2} \ \ldots \ \mathbf{a_n})$ where $\mathbf{a}_i \in \mathbb{R}^m,\ n,m \in \mathbb{N}_{>0}, \ \forall i \in \llbracket n \rrbracket$, then $\vc (\mathbf{A}) = (\mathbf{a}_1^\top \ \mathbf{a}_2^\top \ \ldots \ \mathbf{a}_n^\top)^\top $.
    \item $\card(\{\cdot\})$ refers to the cardinality of the set $\{ \cdot \}$.
    \item $\mathbf{0}_n \in \mathbb{R}^n$ refers to the null vector of size $n \in \mathbb{N}_{>0}$.
    \item $\|\cdot\|_1$ refers to the $\ell_1$-norm , $\|\cdot\|_2$ refers to the $\ell_2$-norm, $\|\mathbf{y} - \hat{\mathbf{y}}\|_1$ refers to the $\ell_1$-loss, and $\|\mathbf{y} - \hat{\mathbf{y}}\|_2^2$ refers to the $\ell_2$-loss where $\mathbf{y}$ and $\hat{\mathbf{y}}$ are the vectors of all training labels and model predictions, respectively.
\end{itemize}
\color{black}
\subsection{Related Work}\label{subsec:related}

We now review the strands of literature closest to ours. \blue{We refer interested readers to \cite{bishop2006pattern} and \cite{bishop2023deep} for a full horizon of classical ML and modern deep learning as we will not cover concepts such as backpropagation and iterative methods to optimize non-convex neural networks, e.g., SGD \citep{amari1993backpropagation}.}
\paragraph{Convex training of neural networks.}
Convex optimization for machine learning (ML) training \Red{procedures is desirable as it} guarantees convergence to the minimal loss. In recent years,~\cite{pilanci2020neural} have shown that training a feedforward shallow neural network (SNN), i.e., a one-hidden-layer NN, with $\ReLU$ activation functions is equivalent to a convex training procedure. They also extended their results to convolutional neural networks (CNNs). More importantly, they fundamentally linked the $\ReLU$-SNN training problem to a regularized convex program. Following \blue{this work,}~\cite{mishkin2022scnn} have obtained an efficient convex training procedure of SNNs with graphics processing unit (GPU) acceleration, and~\cite{kuelbs2024adversarialtrainingtwolayerpolynomial} have designed an adversarial training \blue{procedure} by controlling the worst-case output of the neural network in a neighbourhood around each training sample. In this work, we generalize the regularization procedure from the convex training of \cite{pilanci2020neural} through the lens of distributionally robust optimization with the Wasserstein metric. \blue{The metric is introduced in Section \ref{subsec:dro}}.

\paragraph{Distributionally robust neural networks.}
Distributionally robust optimization (DRO), in \blue{the} ML \blue{setting}, aims to minimize the expected training loss of an \blue{ML} model over the most adverse data distribution within a distance of the training \blue{set, i.e., the distribution inside a specified ambiguity set, constructed from the available training data, that generates the biggest expected loss.} \blue{As demonstrated by~\cite{kuhn2019wasserstein}, DRO} minimizes the worst-case \blue{optimal} risk associated with the training of the model. Different authors have worked on distributionally robust neural networks. \cite{NEURIPS2023_53be3798} \blue{propose to train} deep neural networks using the first-order approximation of the Wasserstein DRO loss function \blue{in the} backpropagation \blue{but leave the numerical evaluation for future work. Wasserstein DRO uses the definition of the Wasserstein distance to construct the ambiguity set. We cover it in Section \ref{subsec:dro}.} The authors of~\cite{levy2020large} have developed a DRO gradient estimator that integrates into neural networks' \blue{non-convex} optimization pipeline similarly to the widely used SGD optimizer. In~\cite{sagawa2020distributionally}, the authors propose a method to train deep NNs in a DRO fashion by defining the ambiguity set with groups of the training data and increasing regularization. These methods integrate distributional robustness in the training procedure of deep NNs but do not directly tackle the Wasserstein DRO problem, or an equivalent reformulation. \blue{They also all suffer from the lack of global optimality guarantees, making them harder to deploy in large-scale critical applications.} Our proposition, while only applying to shallow networks, \Red{builds on} the tractable reformulation of the Wasserstein DRO problem proposed by~\cite{mohajerin2018data} \Red{and the convergence to a global optimum is ensured by its convex structure.}

\paragraph{Physics-informed neural networks.}
The training approach for physics-informed neural networks (PINNs) primarily involves a weighted multi-objective function designed to narrow the gap between labels and model predictions while penalizing physical constraint violations. This is often achieved by simultaneously minimizing the residuals of a partial differential equation (PDE) and its boundary conditions~\citep{cuomo2022scientific}. \color{black} Suppose all NN training weights are collected in $\boldsymbol{\Theta}$, a typical PINN training program takes the form:
\begin{alignat*}{2}
    &\min_{\boldsymbol{\Theta}} && \ \lambda_\mathcal{F}\mathcal{L_F}\left(\Phi(\mathbf{X}, \boldsymbol{\Theta})\right) + \lambda_\mathcal{B} \mathcal{L_B}\left(\Phi(\mathbf{X}, \boldsymbol{\Theta})\right) +  \lambda_{\text{data}} \mathcal{L}\left(\Phi(\mathbf{X}, \boldsymbol{\Theta}), \mathbf{Y}\right)
\end{alignat*}
where $\mathcal{L_F}$ and $\mathcal{L_B}$ are the functions enforcing the differential equations and boundary conditions, respectively, of the governing physical system, $\mathcal{L}$ is the regression objective between labelled data and model predictions, and $\hat{\mathbf{Y}} = \Phi(\mathbf{X}, \boldsymbol{\Theta})$ \citep{cuomo2022scientific}. The scalar weights $\lambda_\mathcal{F}$, $\lambda_\mathcal{B}$, and $\lambda_{\text{data}}$ must be tuned empirically. 

\Red{We note that the constraints are softly enforced by penalizing their violation in the objective. It is similar to considering the Lagrangian relaxation of the constrained training problem.} 
\color{black}
The loss terms related to the PDE are derived through the automatic differentiation of the neural network's outputs~\citep{baydin2017AD}. This complex loss function is then iteratively optimized through backpropagation within the network. While~\cite{cuomo2022scientific} note that boundary conditions can be strictly imposed during training by forcing a part of the network to satisfy these conditions, see~\cite{zhu2021machine}, they are typically integrated as penalty terms within the loss function. \blue{By using the augmented Lagrangian in the objective function and increasing the penalty terms related to constraint violations iteratively, it is theoretically possible to converge to a PINN which imposes hard inequality constraints during training \citep{LuPINN_hard}.} Physics-informed neural networks are characterized by their extensive modelling capabilities, facilitated by automatic differentiation, which allows for the approximation of continuous PDEs across various neural network architectures, e.g., CNNs~\citep{gao2021phygeonet}, long short-term memory networks~\citep{zhang2020physics}, Bayesian NNs~\citep{YANG2021109913}, and generative adversarial networks~\citep{yang2020pinngan}. Although limited to convex constraints, in this work, we demonstrate that our training methodology can enforce hard constraints on the shallow network's output during training \blue{without complexifying its architecture or objective function}. These constraints may include, but are not limited to, boundary constraints, ramping constraints, \Red{discrete PDEs}, and constraints derived from discrete-time dynamical systems.

\paragraph{Application.}
As stated \blue{previously}, prediction algorithms used \blue{in large-scale VPP applications} must respect some inherent constraints. They need to be computationally scalable, they must be able to model nonlinearities, and the sensibility of the application requires trustworthiness in their predictions, e.g., robustness, interpretability, and \Red{performance} guarantees. \blue{Regarding the forecast of non-residential buildings' energy consumption, because grid operators do not generally have access to information on building activity, occupancy, and control, the system is approached as a blackbox and data-driven approaches tend to be the default option}~\citep{AMASYALI20181192}. Authors have used support vector regression (SVR)~\citep{DONG2005545} and other hybrid methods leveraging both classification and linear regression to satisfy the aforementioned constraints~\citep{AMASYALI20181192}. These methods are linear by parts and can approximate nonlinear distributions adequately. Statistical methods such as \Red{autoregressive integrated moving average (ARIMA)}~\citep{newsham2010arimabuild} and Gaussian processes have also proved to be good contenders as they even offer empirical uncertainty bounds~\citep{weng2015GP}. Many authors have proposed deep-learning methods, yet these propositions do not have much \Red{deployment} acceptability as they fail to meet the industrial constraints \blue{and bloat the requirements for a safe ML pipeline}. In this work, we show that our proposition is interesting for virtual power plant applications where the forecast of many distributed assets is used for critical decision-making, as we satisfy all industrial constraints better than other comparable models.  

\color{black}
\paragraph{Outlier generation.} Distributionally robust learning, in general, aims at protecting machine learning models from out-of-distribution outliers in the training set. These outliers tend to be adversarial to the prediction quality of ML models in validation and testing sets and being able to ignore them should increase models' out-of-sample performance. As such, to correctly benchmark distributionally robust machine learning models, one must be able to generate outliers that are indeed out-of-distribution with respect to the original distribution and be able to do it regardless of the form of the original distribution. In the literature, we found that methodologies to achieve these two objectives in regression tasks are limited. For example, \cite{Chen2018_JMLR} create a training set by randomly drawing samples from two distributions, the inlying and outlying ones, under probability $p \in (0,1)$ and $1-p$, respectively. They used Gaussian distributions for both. In \cite{qi2022distributionally}, the authors task their model with the forecast of a real-world dataset. To control the adversity of the setting, they discard the real labels and create a synthetic version of them by using the prediction of a linear regression trained on the original dataset. They then tune an additive noise over the labels to simulate different levels of outlier corruption. In \cite{liu2022distributionally}, they notably propose generating corrupted synthetic data by integrating attack mechanisms into the generation to bias their dataset, i.e., selection bias and anti-causal bias. While their methodology is interesting, they must rely on multiple Gaussian distributions to design their attacks as intended. By leveraging the sliced-Wasserstein distance \citep{bonnotte2013unidimensional}, a tractable approximation of the Wasserstein distance, our methodology generates outliers, does not rely on any assumptions of the original distribution, making it quite flexible, and ensures that outliers do not create tangible patterns. 
\color{black}

\subsection{Contributions}
\color{black}
In this work, we propose a new distributionally robust convex training procedure for $\ReLU$ shallow neural networks under the Wasserstein metric. Under some conditions, we show that the standard $\ReLU$ shallow convex neural network (SCNN) training can be decoupled and formulated as a constrained linear regression over a modified sample space. SCNN training requires a set of \textit{sampling} vectors, which are known a priori, to model possible activation patterns. We condition over this set to construct a new independent and identically distributed conditional empirical distribution based on the modified sample space. Decoupling the training and working with this new empirical distribution allows us to easily adapt the SCNN training to the tractable formulation of the Wasserstein-DRO problem proposed by \cite{mohajerin2018data}, derive out-of-sample performance guarantees stemming from its Rademacher complexity, and enforce hard convex physical constraints during training. 

Additionally, we propose a first post-training verification program specifically tailored to SCNNs. We model the inner workings of SCNNs as a mixed-integer convex optimization problem to certify the worst-case local instability in the feature space of a trained SCNN. This post-verification framework enables us to evaluate how different regularization paradigms, such as distributionally robust learning, influence the stability of trained SCNNs without having to map their weights to their non-convex equivalents. This program also opens research opportunities for bilevel programs where SCNN certification and training are tackled simultaneously.

We propose a new methodology to benchmark distributionally robust (and regularized) machine learning models in a controlled synthetic environment. We design a method to introduce real out-of-sample outliers under the sliced-Wasserstein distance \citep{bonnotte2013unidimensional}, a computationally tractable approximation of the Wasserstein distance. We combine this method with noise to corrupt machine learning datasets generated with standard optimization benchmark functions typically reserved for non-convex solver benchmarking. We show that our WaDiRo-SCNN training performs well empirically in highly corrupted environments: it is robust and conservative while being able to model non-convex functions. Finally, we test our model on a real-world VPP application, viz., the hourly energy consumption prediction of non-residential buildings in Montréal, Canada. For this application, we enforce the physical constraint that energy consumption is non-negative. We observe that this constraint greatly reduces the number of constraint violations during testing while having no significant precision drop with the original non-physics-constrained WaDiRo-SCNN.
\color{black}

To the best of our knowledge, our specific contributions are as follows:
\begin{itemize}
    \item We demonstrate that the SCNN with $\ell_1$-loss training problem is similar to a \blue{constrained} linear regression over a modified sample space.
    \item We formulate an exact, tractable, and low-stochasticity Wasserstein distributionally robust training program for shallow convex neural networks under both $\ell_1$-norm and $\ell_2$-norm order-1 Wasserstein distances \blue{by exploiting this constrained linear regression formulation and linking it with past literature}. 
    \item We derive out-of-sample guarantees for WaDiRo-SCNNs with $\ell_1$-loss training via the Rademacher complexity \blue{under its original sample space and its modified one}.
    \item We show that hard convex physical constraints can be easily included in the training procedure of SCNNs with $\ell_1$-norm loss and their WaDiRo counterpart.
    \item \blue{We design a mixed-integer convex program (MICP) for the post-training verification of SCNNs. We utilize it to show how distributionally robust and regularized training methods increase model stability empirically.}
    \item \blue{We propose a new benchmarking procedure with synthetic data to empirically evaluate the robustness of ML models regarding out-of-distribution outliers.}
    \item We illustrate the conservatism of WaDiRo-SCNNs in predicting synthetic benchmark functions with controlled data corruption and its performance in predicting real-world non-residential buildings' consumption patterns.
\end{itemize}
\color{black}
The rest of the paper is organized as follows. In Section \ref{sec:pre}, we introduce Wasserstein distributionally robust optimization and its tractable formulation for convex problems (Section \ref{subsec:dro}) as well as shallow convex neural networks (Section \ref{subsec:scnn}). In Section \ref{sec:wdroscnn} we present the main results of this paper, \Red{namely}, WaDiRo-SCNNs. In Sections \ref{sec:osp}, \ref{sec:pcscnn}, and \ref{sec:post} we address the theoretical out-of-sample performance of our proposition, physics-constrained SCNNs, and SCNN post-training verification, respectively. Finally, in Section \ref{sec:outliers} we present a methodology to generate outliers and in Section \ref{sec:numerical}, we present all numerical experiments.
\color{black}
\section{Preliminaries}\label{sec:pre}
We start by covering the fundamentals of Wasserstein distributionally robust optimization before introducing shallow convex neural networks. 
\subsection{Distributionally Robust Optimization}\label{subsec:dro}

Distributionally robust optimization (DRO) minimizes a worst-case expected loss function over an ambiguity set, i.e., an uncertainty set in the space of probability distributions~\citep{kuhn2019wasserstein}. The problem can be formulated as:
\begin{alignat}{1}
   &\inf_{\boldsymbol{\beta}} \sup_{\mathbbm{Q} \in \Omega} \  \mathbb{E}^\mathbbm{Q}[h_{\boldsymbol{\beta}}(\mathbf{z})],   \label{eq:DRO} \tag{\texttt{DRO}}
\end{alignat}
where $\mathbf{z} \in \mathcal{Z} \subseteq \mathbb{R}^d, d \in \mathbb{N}$, is a random \Red{vector}, $\mathcal{Z}$ is the set of all possible values of $\mathbf{z}$, $\mathbbm{Q}$ is a distribution from the ambiguity set $\Omega$ \Red{that characterizes the distribution of the random vector $\mathbf{z}$}, $\boldsymbol{\beta} \in \mathbb{R}^p$ is the $p$-dimension decision vector, and $h_{\boldsymbol{\beta}}(\mathbf{z}): \mathcal{Z}\times \mathbb{R}^p \to \mathbb{R}$ is the \blue{objective} function when applying $\boldsymbol{\beta}$ on a sample~\citep{Chen2020_DRO,mohajerin2018data}.

We are interested in a data-driven branch of DRO which defines the ambiguity set as a Wasserstein ball\blue{, a ball which contains all distributions within a Wasserstein distance from a reference distribution \citep{yue2022linear},} centred on an empirical distribution containing $N$ samples: $\hat{\mathbbm{P}}_N = \frac{1}{N} \sum_{i=1}^N \delta_{\mathbf{z}_i}$, where $\delta_{\mathbf{z}_i}$ is a Dirac delta function assigning a probability mass of one at each known sample $\mathbf{z}_i$. \Red{The empirical distribution is constructed with historical realizations of the random vector $\mathbf{z}$.} Note that as the number of samples goes to infinity the empirical distribution \Red{weakly} converges to the true distribution~$\mathbbm{P}$\blue{~\citep[Chapter 19.2]{vandervaart_stats}}. The Wasserstein distance \Red{(WD)}, {also called \blue{earth mover's distance and closely related to the optimal transport plan \citep{kuhn2019wasserstein}}}, can be pictured as the {minimal effort} that it would take to displace a given pile of a weighted resource, e.g., dirt, to shape a second specific pile. The effort of a trip is given by the distance travelled and the weight of the carried resource. Specifically, it provides a sense of similarity between the two distributions. Formally, the \blue{probabilistic definition of the} order-$t$ Wasserstein distance between \blue{probability distributions} $\mathbbm{U}$ and $\mathbbm{V}$ \blue{on $\mathcal{Z}$} is: \color{black}
$$W_{\|\cdot\|,t}(\mathbbm{U}, \mathbbm{V}) = \inf_{\substack{U \sim \mathbbm{U},\\ V \sim \mathbbm{V}}} \left( \mathbb{E}[\|U - V \|^t] \right)^{\frac{1}{t}}, \quad t \geq 1,$$
%$$W_{\|\cdot\|,t}(\mathbbm{U}, \mathbbm{V}) = \left( \min_{\pi \in \mathcal{P}(\mathbbm{U}, \mathbbm{V})} \int_{\mathcal{Z}\times\mathcal{Z}} \|\mathbf{z}_1 - \mathbf{z}_2\|^t \text{d}\pi(\mathbf{z}_1,\mathbf{z}_2)\right)^{1/t},$$
where the infimum is taken over all possible couplings of random vectors $U$ and $V$ which are marginally distributed as $\mathbbm{U}$ and $\mathbbm{V}$, respectively \citep{panaretos2019statistical}. \blue{As such, the Wasserstein ball defines a space containing all possible distributions close to the distribution in its center. We refer interested readers to the open-access book of \cite{panaretos2020invitation} which covers the Wasserstein distance thoroughly while offering interesting statistical insights.}\color{black}

In this context,~\eqref{eq:DRO} is expressed as:
\begin{alignat}{2}
   &\inf_{\beta} \sup_{\mathbbm{Q} \in \Omega} \ && \mathbb{E}^\mathbbm{Q}[h_{\beta}(\mathbf{z})] \tag{\texttt{WDRO}}\label{eq:WDRO}\\
   &\text{with} \ \Omega = && \{ \mathbbm{Q} \in \mathcal{P}(\mathcal{Z}) : W_{\|\cdot\|,t}(\mathbbm{Q}, \hat{\mathbbm{P}}_N) \leq \epsilon\}, \nonumber
\end{alignat}
where $\Omega$ is the order-$t$ Wasserstein \blue{ball} of radius $\epsilon > 0$ centred on $\hat{\mathbbm{P}}_N$, and $\mathcal{P}(\mathcal{Z})$ is the \Red{set} of all probability distributions with support on $\mathcal{Z}$. In a way, $\epsilon$ can be interpreted as a hyperparameter that controls our conservatism towards $\hat{\mathbbm{P}}_N$ \blue{because a greater radius admits distributions farther from $\hat{\mathbbm{P}}_N$ into the ambiguity set}.  This problem is {not tractable}~\citep{mohajerin2018data}.

Using the strong duality theorem and the order-1 Wasserstein metric,~\cite{mohajerin2018data} show that an exact tractable reformulation of \eqref{eq:WDRO} is obtainable if $h_{\boldsymbol{\beta}}(\mathbf{z})$ is convex in $\mathbf{z} \in \mathcal{Z}$ \blue{and if $\mathcal{Z} = \mathbb{R}^d$}, where $d$ is the dimension of the samples. The problem to consider then becomes:
\begin{alignat}{2}
   &\inf_{\boldsymbol{\beta}} \sup_{\mathbbm{Q} \in \Omega} \  \mathbb{E}^\mathbbm{Q}[h_{\boldsymbol{\beta}}(\mathbf{z})] = &&\inf_{\boldsymbol{\beta}} \kappa\epsilon + \frac{1}{N} \sum_{i=1}^{N} h_{\boldsymbol{\beta}}(\mathbf{z}_i) \tag{\texttt{TWDRO}} \label{eq:CWDRO}\\
   &\ &&\,\text{s.t.} \ \kappa =  \sup \{ \| \boldsymbol{\theta} \|_* : h_{\boldsymbol{\beta}}^*(\boldsymbol{\theta}) < +\infty \}, \nonumber
\end{alignat}
where $\| \cdot \|_*$ is the dual norm, $\| \boldsymbol{\theta} \|_* \equiv \sup_{\|\mathbf{z}\|\leq 1} \boldsymbol{\theta}^\top \mathbf{z}$, and $h_{\boldsymbol{\beta}}^*(\boldsymbol{\theta}) $ is the convex conjugate of $h_{\boldsymbol{\beta}}(\mathbf{z})$, $h_{\boldsymbol{\beta}}^*(\boldsymbol{\theta}) = \sup_\mathbf{z} \{ \boldsymbol{\theta}^\top \mathbf{z} - h_{\boldsymbol{\beta}}(\mathbf{z})\}$~\citep{boyd2004convex}. 

Training machine learning models with \eqref{eq:CWDRO} has been proved to be a generalization of regularization~\citep{shafieezadeh2019regularization}. By training over the worst-case expected distribution in a Wasserstein ball centred on the training dataset, we can expect to avoid overfitting on the empirical distribution, thus \blue{yielding increased robustness} against outliers, \blue{noise}, and some adversarial attacks.

\subsection{Shallow Convex Neural Networks}\label{subsec:scnn}
Let $\mathbf{X} \in \mathbb{R}^{N \times d}$ be a data matrix of $N$ feature vectors $\mathbf{x}_j \in \mathbb{R}^d \ \forall j \in  \llbracket N \rrbracket$, and $\mathbf{y} \in \mathbb{R}^N$ be the corresponding one-dimensional labels. Define data samples as a feature-label pair: $\mathbf{z}_j = (\mathbf{x}_j, y_j) \ \forall j \in \llbracket N \rrbracket$. The non-convex training problem of a single-output feedforward shallow neural network (SNN) with $\ReLU$ activation functions, \Red{without regularization, and without bias weights} is as follows:
\begin{alignat}{1}
    p^\star &= \min_{\mathbf{W}_{1},\mathbf{w}_2 } \mathcal{L}\left( \sum_{i=1}^{m}\{(\mathbf{X}\mathbf{W}_{1i})_{+} w_{2i}\}, \mathbf{y}   \right), \tag{\texttt{SNNT}} \label{eq:SNNT}
\end{alignat}
where $m$ is the number of hidden neurons,  $\mathbf{W}_1 \in \mathbb{R}^{d \times m}$, and $\mathbf{w}_2 \in \mathbb{R}^m$ are the weights of the first and second layers, $\mathcal{L}: \mathbb{R}^N \rightarrow \mathbb{R}$ is a convex loss function, and $(\cdot)_+$ is an element-wise $\max\{0, \cdot \}$ operator ($\ReLU$).

The convex reformulation of the SNN training problem \eqref{eq:SNNT} is based on an enumeration process of all the possible {$\ReLU$ activation patterns} in the hidden layer for a {fixed training dataset}~$\mathbf{X}$~\citep{pilanci2020neural}. The enumeration is equivalent to generating all possible hyperplane arrangements passing through the origin and clustering the data with them. The set of all \Red{$\Bar{P}$} possible activation patterns a $\ReLU$ SNN can take for a training set $\mathbf{X}$ is given by:
\begin{equation}\label{eq:dmatrices}
    \mathcal{D}_{\mathbf{X}} = \Red{\{\mathbf{D}_1, \mathbf{D}_2, \ldots, \mathbf{D}_{\Red{\Bar{P}}} : \mathbf{D}_i \neq \mathbf{D}_{j\neq i}, \ i,j \in \llbracket \Red{\Bar{P}} \rrbracket \} \subset} \left\{ \mathbf{D} = \diag(\mathbbm{1}(\mathbf{Xs}\succeq \mathbf{0})) : \mathbf{s}\in \mathbb{R}^d \right\}, 
\end{equation}
where $\mathbf{D}$ is a diagonal matrix with a possible activation pattern encoded on the diagonal \Red{and $\mathbf{s}$ is the vector used to encode the activation pattern with respect to the dataset $\mathbf{X}$}.

By introducing the cone-constrained variables $\boldsymbol{\nu}, \boldsymbol{\omega}\in \mathbb{R}^{\Red{\Bar{P}} \times d}$ and the diagonal matrices from $\mathcal{D}_\mathbf{X}$,~\cite{pilanci2020neural} have obtained a convex training procedure from which we can \blue{retrieve} the previous non-convex weights under some conditions. The optimization problem is now defined as
\begin{alignat}{2}
  p^\star = &\min_{\boldsymbol{\nu}, \boldsymbol{\omega}} && \mathcal{L}\left( \sum_{i = 1}^{\Red{\Bar{P}}} \mathbf{D}_i \mathbf{X}(\boldsymbol{\nu}_i - \boldsymbol{\omega}_i), \mathbf{y} \right) \tag{\texttt{SCNNT}}\label{eq:SCNNT}\\
&\ \text{s.t.}\quad  \quad &&\boldsymbol{\nu}_i,\boldsymbol{\omega}_i \in \mathcal{K}_i \quad  \forall i \in \llbracket \Red{\Bar{P}} \rrbracket, \nonumber
\end{alignat}
where  $\mathcal{K}_i = \{ \mathbf{u} \in \mathbb{R}^d : (2\mathbf{D}_i - \mathbf{I})\mathbf{X}\mathbf{u} \succeq \mathbf{0}\}$ and $\Red{\Bar{P}} = \card (\mathcal{D}_{\mathbf{X}})$. 
For $m\geq m^\star=\sum_{i:\boldsymbol{\nu}_i \neq \mathbf{0}}^{\Red{\Bar{P}}} 1 + \sum_{i:,\boldsymbol{\omega}_i \neq \mathbf{0}}^{\Red{\Bar{P}}} 1 $, \blue{with $1\leq m^\star \leq N + 1$, }problems \eqref{eq:SNNT} and \eqref{eq:SCNNT} have identical optimal values~\citep{pilanci2020neural}.  \blue{At least one optimal non-convex NN from \eqref{eq:SNNT} can be assembled, using $m^\star$ neurons, with the optimal solution of  \eqref{eq:SCNNT} and the following mapping} between the optimal weights of each problem:
\begin{alignat}{2}
& \left(\mathbf{W}_{1i}^\star, \mathbf{w}_{2i}^\star\right)=\left(\frac{\boldsymbol{\nu}_i^\star}{\sqrt{\left\|\boldsymbol{\nu}_i^\star\right\|_2}}, \sqrt{\left\|\boldsymbol{\nu}_i^\star\right\|_2}\right), \quad &&\text { if } \quad \boldsymbol{\nu}_i^\star \neq \mathbf{0}, \forall i \in \llbracket \Red{\Bar{P}} \rrbracket \label{eq:map1}\\
&  \left(\mathbf{W}_{1(i+\Red{\Bar{P}})}^\star, \mathbf{w}_{2(i+\Red{\Bar{P}})}^\star\right)=\left(\frac{\boldsymbol{\omega}_i^\star}{\sqrt{\left\|\boldsymbol{\omega}_i^\star\right\|_2}},-\sqrt{\left\|\boldsymbol{\omega}_i^\star\right\|_2}\right), \quad &&\text { if } \quad \boldsymbol{\omega}_i^\star \neq\mathbf{ 0}, \forall i \in \llbracket \Red{\Bar{P}} \rrbracket. \label{eq:map2}
\end{alignat}
\blue{We note that \cite{wang2021hidden} demonstrate how to recover the full optimal set of non-convex networks.} 

To make the problem computationally tractable, we can \blue{approximate} the full hyperplane arrangements by sampling random vectors~\citep{mishkin2022scnn}. For example, using a Gaussian sampling, we \Red{approximate} $\mathcal{D}_{\mathbf{X}}$ by $\Tilde{\mathcal{D}}_{\mathbf{X}} = \{ \mathbf{D}_i = \diag(\mathbbm{1}(\mathbf{X S} \succeq \mathbf{0})) : \mathbf{S}  \thicksim \mathcal{N}_d(\boldsymbol{\mu}, \boldsymbol{\Sigma}), \mathbf{D}_i \neq \mathbf{D}_{j \neq i},\ i,j \in \llbracket P \rrbracket \}$ where \Red{$P \leq \Bar{P}$}, $\mathbf{S}$ is a random Gaussian vector of mean $\boldsymbol{\mu} \in \mathbb{R}^d$ and of covariance $\boldsymbol{\Sigma}\in \mathbb{R}^{d \times d}$,  and \blue{such that $\mathbf{s}_i \in \mathbb{R}^d$ is \Red{the realization generating matrix $\mathbf{D}_i$},} $ \forall i \in \llbracket P \rrbracket$. \Red{We note that other sampling methods can be used, e.g., uniform sampling, and they may possibly lead to richer sets $\Tilde{\mathcal{D}}_{\mathbf{X}}$.} This \blue{approximation} has the same minima as \eqref{eq:SCNNT} if 
\begin{equation*} m \geq b \coloneq \sum_{\mathbf{D}_i \in \Tilde{\mathcal{D}}_\mathbf{X}} \blue{\card} \left(\{ \boldsymbol{\nu}_i^\star :\boldsymbol{\nu}_i^\star \neq \mathbf{0}\} \cup \{ \boldsymbol{\omega}_i^\star :\boldsymbol{\omega}_i^\star \neq \mathbf{0}\} \right), \end{equation*}
where in this case $\boldsymbol{\nu}^\star, \boldsymbol{\omega}^\star$ are global optima of \blue{the sub-sampled convex problem} and if the optimal activations are in the convex model\blue{, i.e., $\left\{ \diag (\mathbf{X W}_{1i}^\star \geq 0) : i \in \llbracket m \rrbracket\right\} \subseteq \Tilde{\mathcal{D}}_{\mathbf{X}}$} \citep{mishkin2022scnn}. Interested readers are referred to~\cite{pilanci2020neural} and \cite{mishkin2022scnn} for the detailed coverage of SCNNs. 

We remark that if these conditions are met, for a feature sample $\mathbf{x}$ not included in the training dataset $\mathbf{X}$, then the non-convex formulation of the neural network gives the same output as the convex formulation when using the same set of realized sampling vectors \blue{$\mathcal{S} = \{ \mathbf{s}_1,  \mathbf{s}_2, \ldots,  \mathbf{s}_P \} $}.  \Red{We also observe empirically that as the cardinality of $\Tilde{\mathcal{D}}_{\mathbf{X}}$ grows, both formulation converges to the same outputs for a given input.} This notion will be useful in the next section. \color{black}

If these conditions are not met, i.e., the SNN is not wide enough, we still obtain an approximately optimal network that can outperform conventional training methods as highlighted by \cite{mishkin2022scnn}. By \Red{approximate optimality}, we mean that it is optimized by a convex program that guarantees optimality while not having an exact mapping to the optimal non-convex neural network formulation. Thus, \Red{imposing} the maximal width of the SCNN, \Red{which depends on the cardinality of $\Tilde{\mathcal{D}}_{\mathbf{X}}$}, is a way to control the tractability of the convex training with respect to the available computational power and the size of the training dataset. \Red{Because generating the full hyperplane arrangement is impractical in most settings, the notation of the subsequent sections is aligned with the use of $\Tilde{\mathcal{D}}_{\mathbf{X}}$.}\color{black}

\section{Wasserstein Distributionally Robust Shallow Convex Neural Networks}\label{sec:wdroscnn}
We now introduce WaDiRo-SCNNs and provide tractable procedures to train them. First, we reexpress the training \blue{objective} function of the SCNN in a form compatible with the tractable formulation of the Wasserstein DRO problem from~\cite{mohajerin2018data} to obtain performance guarantees for critical applications.

We make the following assumption on samples from $\mathbbm{P}$.
\begin{assumption}\label{ass:z_iid}
    Samples $\mathbf{z} = (\mathbf{x}, y) \in \mathcal{Z}=\mathbb{R}^{d+1}$ are independent and identically distributed (i.i.d.).
\end{assumption}
This is a common assumption in statistical machine learning which is also made by, e.g.,~\cite{Chen2020_DRO} and \cite{mohajerin2018data}. \blue{In practice, in learning settings, we only have access to a finite amount of training samples from which we can indirectly observe~$\mathbbm{P}$. This motivates the use of a finite empirical distribution~$\hat{\mathbbm{P}}_N$ that approximates~$\mathbbm{P}$ \citep{kuhn2019wasserstein}. As such, we assume that our empirical samples are drawn independently from a consistent unknown distribution \Red{with i.i.d.~samples}. When formulating \eqref{eq:CWDRO}, only the independence of the samples is necessary to build $\hat{\mathbbm{P}}_N$. The stronger identically distributed assumption is then only required to bound the out-of-sample performance, see Section \ref{sec:osp}. As a result, the latter part of Assumption~\ref{ass:z_iid} implies that any distribution drift is neglected from the analysis.} \Red{While Assumption \ref{ass:z_iid} is not always preserved in practice, e.g., time-series forecasting, it is required for the theoretical analysis.} Recall~\eqref{eq:dmatrices} and let $d_{ji} = \mathbf{D}_i (j,j) = \mathbbm{1}(\mathbf{x}_j^\top \mathbf{s}_i \geq \mathbf{0}), \ \forall j \in \llbracket N \rrbracket, \ i \in \llbracket P \rrbracket$ such that $\mathbf{d}_j \in \{0,1\}^P$ is the vector encoding the activation state of $\mathbf{x}_j$ for each vector in $\mathcal{S}$. For the remainder of this section, we consider the modified features ${\hat{\mathbf{x}}}_j = \left( \vc(\mathbf{x}_j \mathbf{d}_j^\top), \vc(\mathbf{x}_j \mathbf{d}_j^\top)\right)\ \forall j \in \llbracket N \rrbracket$, which are a \blue{higher order basis expansion} of the original features with their respective possible activation patterns. \Red{This basis expansion is justified in the proof of the following lemma.} 

\blue{We now proceed to decouple the training procedure for distinct samples.}
\begin{lemma}\label{lem:linreg}
Suppose that the training loss function $\mathcal{L}$ is the $\ell_1$-norm, and let $\hat{\mathbf{z}}_j = \left(\hat{\mathbf{x}}_j, y_j\right) \in \hat{\mathcal{Z}}, \ j \in \llbracket N \rrbracket $ be modified samples. Then, the convex training procedure \eqref{eq:SCNNT} is equivalent to a \blue{constrained} linear regression problem.
\end{lemma} 
\textit{Proof:} Define $\mathbf{u}_i = \boldsymbol{\nu}_i - \boldsymbol{\omega}_i $ and let $\mathbf{U}\in \mathbb{R}^{P\times d}$ be the matrix collecting each $\mathbf{u}_i^\top$, $\forall j \in \llbracket N \rrbracket, \forall i \in \llbracket P \rrbracket$. Considering the $\ell_1$-norm loss function, we have:
\begin{align*}
   \mathcal{L}\left( \sum_{i = 1}^P \mathbf{D}_i \mathbf{X}(\boldsymbol{\nu}_i - \boldsymbol{\omega}_i), \mathbf{y} \right) &= \blue{\left\| \left( \sum_{i=1}^{P} d_{1i} \mathbf{x}_1^\top \mathbf{u}_i, \sum_{i=1}^{P} d_{2i} \mathbf{x}_2^\top \mathbf{u}_i,\ldots, \sum_{i=1}^{P} d_{Ni} \mathbf{x}_N^\top \mathbf{u}_i\right) - \mathbf{y} \right\|_1} \\
    &=   \sum_{j=1}^{N} \left| \mathbf{d}_j^\top \mathbf{U} \mathbf{x}_j - y_j   \right| \\
    &=  \sum_{j=1}^{N} | x_{j1}(d_{j1}u_{11} + \ldots + d_{jP}u_{P1}) + \ldots +  \\
    &\qquad \qquad \qquad \qquad \qquad \qquad \qquad \  x_{jd}(d_{j1}u_{1d}  + \ldots + d_{jP}u_{Pd}) - y_j  |  \\
    &=  \sum_{j=1}^{N} \left|  \left( \vc(\mathbf{U}), -1 \right)^\top \left( \vc(\mathbf{x}_j \mathbf{d}_j^\top) , y_j \right)  \right| \\
    &=  \sum_{j=1}^{N} \left|  \left( \vc(\boldsymbol{\nu}),\vc(-\boldsymbol{\omega}),  -1 \right)^\top \left( \vc(\mathbf{x}_j \mathbf{d}_j^\top), \vc(\mathbf{x}_j \mathbf{d}_j^\top), y_j \right)    \right|  \\
    &=   \sum_{j=1}^{N} \left| {\boldsymbol{\beta}}^\top \hat{\mathbf{z}}_j   \right|,  
\end{align*}
where $\boldsymbol{\beta} = \left( \vc(\boldsymbol{\nu}),\vc(-\boldsymbol{\omega}),  -1 \right) $ represent the weights of the linear regression and ${\hat{\mathbf{z}}}_j = \left( \vc(\mathbf{x}_j \mathbf{d}_j^\top), \vc(\mathbf{x}_j \mathbf{d}_j^\top), y_j \right) = \left( \hat{\mathbf{x}}_j, y_j \right) \ \forall j \in \llbracket N \rrbracket$ are the modified samples. The training problem~\eqref{eq:SCNNT} can be written as:
\begin{alignat*}{3}
p^\star_{\ell_1}=&\min_{\boldsymbol{\nu}, \boldsymbol{\omega}: \boldsymbol{\nu}_i,\boldsymbol{\omega}_i \in \mathcal{K}_i \  \forall i \in \llbracket P \rrbracket} && \left\| \mathbf{y} - \blue{\sum_{i=1 }^P} \mathbf{D}_i \mathbf{X}(\boldsymbol{\nu}_i - \boldsymbol{\omega}_i)  \right\|_1 \\
= &\min_{\boldsymbol{\nu}, \boldsymbol{\omega}: \boldsymbol{\nu}_i,\boldsymbol{\omega}_i \in \mathcal{K}_i \ \forall i \in \llbracket P \rrbracket} && \sum_{j=1}^{N} \left| {\boldsymbol{\beta}}^\top {\hat{\mathbf{z}}}_j   \right|\\
&\qquad \ \ \ \text{s.t.} && \boldsymbol{\beta} = \left( \vc(\boldsymbol{\nu}),\vc(-\boldsymbol{\omega}),  -1 \right),
\end{alignat*}
and is equivalent to a constrained linear regression problem with respect to the modified samples.~\Red{$\square$}

\color{black}

Additionally, we note that the LASSO \Red{and ridge-regularized} SCNN training problems are exactly equivalent to their respective linear regression formulation:
\begin{alignat*}{2}
    p^\star_{\ell_1, r}=& \min_{\boldsymbol{\nu}, \boldsymbol{\omega}: \boldsymbol{\nu}_i,\boldsymbol{\omega}_i \in \mathcal{K}_i \  \forall i \in \llbracket P \rrbracket} && \left\| \mathbf{y} - \sum_{i=1}^P \mathbf{D}_i \mathbf{X}(\boldsymbol{\nu}_i - \boldsymbol{\omega}_i)  \right\|_1 + \sum_{i=1}^P \left(r(\boldsymbol{\nu}_i) + r(\boldsymbol{\omega}_i)\right) \\
    =&\min_{\boldsymbol{\nu}, \boldsymbol{\omega}: \boldsymbol{\nu}_i,\boldsymbol{\omega}_i \in \mathcal{K}_i \ \forall i \in \llbracket P \rrbracket} && \sum_{j=1}^{N} \left| y_j -  \left( \vc(\boldsymbol{\nu}),\vc(-\boldsymbol{\omega})\right)^\top {\hat{\mathbf{x}}}_j   \right| + r\left( [ \vc(\boldsymbol{\nu}),\vc(-\boldsymbol{\omega}) ]\right),
\end{alignat*}
where $r(\cdot) = \|\cdot\|_1$ for LASSO regularization and  $r(\cdot) = \|\cdot\|_2^2$ for ridge regularization.

\color{black} The $\ell_1$-norm loss function tends to be more robust against outliers than other regression objectives when training regression models~\citep{Chen2020_DRO}. It is then a \blue{preferable} choice in an application where distributional robustness is desired and will be used for the remainder of this work \blue{as it is also convenient in the WaDiRo reformulation. Nonetheless, we remark that Lemma \ref{lem:linreg} can be adapted to other separable loss functions, i.e., functions $f$ such that $f(\phi_1(x_1) + \phi_2(x_2) + \ldots \phi_N(x_N)) = f(\phi_1(x_1)) + f(\phi_2(x_2)) + \ldots + f(\phi_N(x_N))$ where $x_j \ j \in \llbracket N \rrbracket$ are variables and $\phi_j \ j \in \llbracket N \rrbracket$ are arbitrary functions}. \Red{Other commonly used separable loss functions include the $\ell_p$-norm to the power of $p$, i.e., $\|\cdot\|_p^p$, for $p \in [1,+\infty)$.}

We remark that the transformed feature space preserves the independence of individual samples when conditioned on $\mathcal{S} = \{ \mathbf{s}_1, \mathbf{s}_2, \ldots,  \mathbf{s}_P \} $ \blue{because the mapping from the original feature space to the modified one depends only on individual samples and $\mathcal{S}$.} Whether $\mathcal{S}$ is constructed by random sampling or by a deterministic hyperplane arrangement has no importance as long as these vectors are saved prior to training and are used throughout the training and evaluation. \blue{From this observation, we now introduce our second lemma.}
\begin{lemma}\label{lem:iid}
Suppose Assumption \ref{ass:z_iid} holds, \blue{i.e., samples \Red{$\mathbf{z}  = (\mathbf{x}, y) \in \mathcal{Z}$} are i.i.d., then modified samples \Red{ $\hat{\mathbf{z}} = (\hat{\mathbf{x}}, y)$} given the set $\mathcal{S}$, $\hat{\mathbf{z}}|\mathcal{S} \in \hat{\mathcal{Z}}$,  are also} i.i.d.
\end{lemma}
\textit{Proof:} We observe that: 
\begin{alignat*}{2}
    \mathbf{x}_j\mathbf{d}_j^\top &= \mathbf{x}_j \begin{bmatrix}
        \mathbbm{1}(\mathbf{x}_j^\top \mathbf{s}_1 \geq 0) & \mathbbm{1}(\mathbf{x}_j^\top \mathbf{s}_2\geq 0) & \ldots & \mathbbm{1}(\mathbf{x}_j^\top \mathbf{s}_P\geq 0) 
    \end{bmatrix} \quad &&\forall j \in \llbracket N \rrbracket.
    \intertext{Using Lemma \ref{lem:linreg}, we obtain} 
    g_{\boldsymbol{\beta}}(\hat{\mathbf{z}}_j) &= \left|  \boldsymbol{\beta}^\top \left( \vc(\mathbf{x}_j \mathbf{d}_j^\top), \vc(\mathbf{x}_j \mathbf{d}_j^\top), y_j \right)    \right| && \forall j \in \llbracket N \rrbracket\\
    &= \left| \boldsymbol{\beta}^\top \left(f(\mathbf{x}_j, \mathbf{s}_1, \mathbf{s}_2, \ldots, \mathbf{s}_P), y_j \right)  \right| && \forall j \in \llbracket N \rrbracket \\
    &= \left| \boldsymbol{\beta}^\top \left(\hat{\mathbf{x}}_j, y_j \right)  \right| && \forall j \in \llbracket N \rrbracket ,
\end{alignat*}
where $\hat{\mathbf{x}}_j = f(\mathbf{x}_j, \mathcal{S})$ \blue{is a basis expansion of the original feature space using the definition of the hyperplane generation (or approximation) and $g_{\boldsymbol{\beta}}(\hat{\mathbf{z}}): \hat{\mathcal{Z}}\times \mathbb{R}^{2Pd + 1} \to \mathbb{R}$ is the modified objective function when applying $\boldsymbol{\beta}$ on a modified sample.}
We have showed that each $\hat{\mathbf{x}}_j$ \blue{depends} solely \blue{on} $\mathbf{x}_j$ and $\mathcal{S}$. Because $\mathcal{S}$ is obtained before training\Red{, either by random sampling or by a deterministic hyperplane arrangement which depends on the whole training set $\mathbf{X}$,} and the mapping $f$ is \blue{the same} on every sample, $\hat{\mathbf{z}}|\mathcal{S}$ is also i.i.d.~\hfill $\square$

Following Lemmas \ref{lem:linreg} and \ref{lem:iid}, we consider the new empirical distribution of the modified samples conditioned on $\mathcal{S}$: $\hat{\mathbbm{F}}_N^{\mathcal{S}} = \frac{1}{N} \sum_{j=1}^N \delta_{\hat{\mathbf{z}}_j|\mathcal{S}}$, where $\hat{\mathbf{z}}_j = \left(\hat{\mathbf{x}}_j, y_j\right)^\top$. \blue{As the number of modified samples grows, $\hat{\mathbbm{F}}_N^{\mathcal{S}}$ converges to the true underlying distribution $\mathbbm{F}^\mathcal{S}$.} 

\color{black}
To make our case for the similarity between samples of $\hat{\mathbbm{P}}_N$ and $\hat{\mathbbm{F}}_N^{\mathcal{S}}$, we provide a detailed example of the mapping between the samples of the two distributions. 

\begin{example}\label{ex:mod_samples}
Let $\mathbf{z}_j = (\mathbf{x}_j, y_j)$ be any sample from $\hat{\mathbbm{P}}_N$ and $\hat{\mathbf{z}}_j = (\hat{\mathbf{x}}_j, y_j)$ be the equivalent sample from $\hat{\mathbbm{F}}_N^{\mathcal{S}}$. Suppose that $\card(\mathcal{S}) = 3$ and that $\mathbf{d}_j^\top = \begin{bmatrix}
        \mathbbm{1}(\mathbf{x}_j^\top \mathbf{s}_1 \geq 0) & \mathbbm{1}(\mathbf{x}_j^\top \mathbf{s}_2\geq 0)  & \mathbbm{1}(\mathbf{x}_j^\top \mathbf{s}_3\geq 0)
    \end{bmatrix} = \begin{bmatrix}
        1 & 0 & 1
    \end{bmatrix}$, then:
\begin{alignat*}{1}
\vc (\mathbf{x}_j\mathbf{d}_j^\top)  &= \vc \left( \mathbf{x}_j \begin{bmatrix}
        1 & 0 & 1
    \end{bmatrix}\right)
    = \vc \left(\begin{bmatrix}
        \mathbf{x}_j & \mathbf{0}_d & \mathbf{x}_j
    \end{bmatrix} \right) = \vc \left(\begin{bmatrix}
        {x}_{j1} & 0 & {x}_{j1}\\
        {x}_{j2} & 0 & {x}_{j2}\\
        \vdots & \vdots & \vdots\\
        {x}_{jd} & 0 & {x}_{jd}\\
    \end{bmatrix} \right) = (\mathbf{x}_j, \ \mathbf{0}_d , \ \mathbf{x}_j),
\intertext{and accordingly:}
\hat{\mathbf{z}}_j &= (\hat{\mathbf{x}}_j, y_j) = \left(\vc (\mathbf{x}_j\mathbf{d}_j^\top), \vc (\mathbf{x}_j\mathbf{d}_j^\top), y_j \right)= \left( (\mathbf{x}_j, \ \mathbf{0}_d , \ \mathbf{x}_j), (\mathbf{x}_j, \ \mathbf{0}_d , \ \mathbf{x}_j), y_j \right).
\end{alignat*}
Now, recall that $\mathcal{S}$ is known a priori and can be generated randomly to approximate the hyperplane arrangement of $\hat{\mathbbm{P}}_N$. As such, $\mathcal{S}$ do not bring additional information to $\mathbf{x}_j$; indeed $\hat{\mathbf{x}}_j$ is purely an expansion of $\mathbf{x}_j$ with a possible $\ReLU$ activation pattern. Effectively, we map the original samples to a higher dimension and apply the Wasserstein distributionally robust regularization on this higher representation, which is closely related to the original feature space. \hfill \Red{$\square$}
\end{example}
\color{black}
We formulate the \Red{intractable} Wasserstein distributionally robust training problem of the SCNN by centering the Wasserstein ball on $\hat{\mathbbm{F}}_N^{\mathcal{S}}$:
\begin{alignat}{2}
   &\inf_{\boldsymbol{\beta}} \sup_{\mathbbm{Q} \in \Omega} \  &&\mathbb{E}^\mathbbm{Q}[g_{\boldsymbol{\beta}}(\hat{\mathbf{z}})|\mathcal{S}] \label{eq:W-DRO SCNN}\tag{\texttt{WDR-SCNNT}}\\
   &\text{s.t.}\  &&\boldsymbol{\nu}_i,\boldsymbol{\omega}_i \in \mathcal{K}_i \quad  \forall i \in \llbracket P \rrbracket \nonumber\\
  & \ && \blue{\boldsymbol{\beta} = \left( \vc(\boldsymbol{\nu}),\vc(-\boldsymbol{\omega}),  -1 \right) } \nonumber
\end{alignat}
where $\hat{\mathbf{z}} \in \hat{\mathcal{Z}}$ and $\mathcal{P}(\hat{\mathcal{Z}})$ is the space of all possible distributions with support on $ \hat{\mathcal{Z}}=\mathbb{R}^{2Pd + 1}$ and where $\Omega$ is defined as:\begin{equation*}
    \Omega = \{ \mathbbm{Q} \in \mathcal{P}(\hat{\mathcal{Z}}) : W_{\|\cdot\|,t}(\mathbbm{Q}, \hat{\mathbbm{F}}_N^\mathcal{S}) \leq \epsilon\}.
\end{equation*} \blue{Note that $g_{\boldsymbol{\beta}}(\hat{\mathbf{z}}) = \left|  \boldsymbol{\beta}^\top \left( \vc(\mathbf{x} \mathbf{d}^\top), \vc(\mathbf{x} \mathbf{d}^\top), y \right)   \right|$ as defined in the proof of Lemma \ref{lem:iid}.}
\begin{proposition}\label{pr:l1}
Consider the $\ell_1$-norm training loss function and suppose Assumption \ref{ass:z_iid} holds, then the convex reformulation of \eqref{eq:W-DRO SCNN} is given by the following:
\begin{enumerate}[(i)]
\item If the $\ell_1$-norm is used in the definition of the order-1 Wasserstein distance $\left(W_{\|\cdot\|_1,1}\right)$, we have:
\begin{alignat*}{4}
   &\min_{\boldsymbol{\nu}, \boldsymbol{\omega}, a, \mathbf{c}} \quad &&\epsilon  a  \ &&+ \frac{1}{N} \sum_{j=1}^{N} c_j \tag{\texttt{WDR1-SCNNT}}\label{eq:WDR1-SCNNT}\\
   &\textup{s.t.} \quad&&{{\beta}}_k &&\leq a \quad &&\forall k \in \llbracket 2P d + 1 \rrbracket\\
    & &&-{{\beta}}_k &&\leq a \quad &&\forall k \in \llbracket 2P  d + 1 \rrbracket\\
   & &&{\boldsymbol{\beta}}^\top \hat{\mathbf{z}}_j &&\leq c_j \quad &&\forall j \in \llbracket N\rrbracket\\
   & &&-{\boldsymbol{\beta}}^\top \hat{\mathbf{z}}_j &&\leq c_j \quad &&\forall j \in \llbracket N\rrbracket\\
    & &&\blue{\boldsymbol{\beta}} && \blue{= \left( \vc(\boldsymbol{\nu}),\vc(-\boldsymbol{\omega}),  -1 \right) }\\
    & &&\boldsymbol{\nu}_i, \boldsymbol{\omega}_i &&\in \mathcal{K}_i \quad  &&\forall i \in \llbracket P\rrbracket.
\end{alignat*}
\item If the $\ell_2$-norm is used in the definition of the order-1 Wasserstein distance $\left(W_{\|\cdot\|_2,1}\right)$, we obtain:
\begin{alignat*}{4}
   &\min_{\boldsymbol{\nu}, \boldsymbol{\omega}, a, \mathbf{c}} \quad && \epsilon  a &&+ \frac{1}{N} \sum_{j=1}^{N} c_j \tag{\texttt{WDR2-SCNNT}}\label{eq:WDR2-SCNNT}\\
   &\textup{s.t.} \quad &&\|{\boldsymbol{\beta}}\|_2^2  &&\leq a^2 \quad \quad &&\\
   & \ &&a &&\geq 0  \\
   & \ &&{\boldsymbol{\beta}}^\top \hat{\mathbf{z}}_j &&\leq c_j \quad &&\forall j \in \llbracket N\rrbracket\\
   & \ &&-{\boldsymbol{\beta}}^\top \hat{\mathbf{z}}_j &&\leq c_j \quad &&\forall j \in \llbracket N\rrbracket\\
   & &&\blue{\boldsymbol{\beta}} && \blue{= \left( \vc(\boldsymbol{\nu}),\vc(-\boldsymbol{\omega}),  -1 \right) }\\
    &\ &&\boldsymbol{\nu}_i, \boldsymbol{\omega}_i &&\in \mathcal{K}_i \quad  &&\forall i \in \llbracket P\rrbracket.
\end{alignat*}
\end{enumerate}
\end{proposition}
\textit{Proof:} Lemma \ref{lem:linreg} states that the training problem, with respect to the modified samples $\hat{\mathbf{z}}$, is similar to a constrained linear regression \blue{over a basis expansion of the original feature space. This problem is convex in its objective and its constraints.} Lemma~\ref{lem:iid} shows that the distribution of the modified samples is i.i.d. when conditioned on $\mathcal{S}$. Because \eqref{eq:W-DRO SCNN} \blue{centres the Wasserstein ball on} $\hat{\mathbbm{F}}_N^{\mathcal{S}}$, by invoking Lemmas~\ref{lem:linreg} and \ref{lem:iid} \blue{regarding the modified sample space}, we \blue{can reexpress}~\eqref{eq:W-DRO SCNN} \blue{to the tractable DRO formulation}~\eqref{eq:CWDRO}. The final forms then follow from Theorem 6.3 of ~\cite{mohajerin2018data} and \blue{are inherited from WaDiRo linear regressions}~\citep[Chapter 4]{Chen2020_DRO}. The detailed proof, \blue{adapted to shallow convex neural networks}, is provided in Appendix \ref{sec:AnnA} for completeness. \hfill $\square$

We remark that~\eqref{eq:WDR1-SCNNT} and~\eqref{eq:WDR2-SCNNT} can be easily modified to take into account the bias weights of the hidden and output layers of the SNN. The process is as follows: (i) extend $\mathbf{X}$ with a column of ones which is equivalent to adding bias weights in the hidden layer; (ii) let $\boldsymbol{\beta}$ be defined as $ \left( \vc(\boldsymbol{\nu}),\vc(-\boldsymbol{\omega}), b,  -1 \right)$ and $ \hat{\mathbf{z}}_j$ as $\left( \vc(\mathbf{x}_j \mathbf{d}_j^\top), \vc(\mathbf{x}_j \mathbf{d}_j^\top), 1, y_j \right) \ \forall j \in \llbracket N \rrbracket$ where $b \in \mathbb{R}$ is the bias of the sole output neuron. We also remark that both formulations of Proposition~\ref{pr:l1} are fairly fast to solve with open-source \blue{convex} solvers, e.g., \texttt{Clarabel}~\citep{Clarabel_2024}\blue{, if the maximal number of neurons dictated by $\Tilde{\mathcal{D}}_{\mathbf{X}}$ and the size of the training set are \textit{reasonable} in regards to the available computational power, e.g., in our case less than 300 neurons and 3000 samples were required. We provide time comparisons in Section \ref{subsec:compute} and identify acceleration methods from the literature that could be adapted to our formulation in Section \ref{sec:conclusion}. Moreover, with only two hyperparameters, i.e., the maximal number of neurons and the Wasserstein ball radius, the hyperparameter search space is significantly reduced compared to standard non-convex training programs. As such, they contribute to a lighter MLOps pipeline by diminishing the computational effort required during hyperparameter tuning.} Recall that the weights from the standard non-convex SNN training problem can be retrieved via \eqref{eq:map1} and~\eqref{eq:map2} \blue{even though they are not required to obtain accurate predictions as the SCNN representation is sufficient}.

\section{Out-of-sample performance}\label{sec:osp}
Next, we \blue{obtain formal out-of-sample performance guarantees for WaDiRo-SCNNs}. By first computing the Rademacher complexity \citep{Bartlett2003RademacherAG}, it is then possible to bound the {expected out-of-sample error} of our model. We make the following additional assumptions similarly to~\cite{Chen2020_DRO}.

\begin{assumption}\label{as:boundZ}
The norm of modified samples from $\mathbbm{F}^\mathcal{S}$ are bounded above: $ \| {\hat{\mathbf{z}}} \| \leq \hat{R} < +\infty$.
\end{assumption}

\begin{assumption}\label{as:boundBeta}
The dual norm of the \Red{trained} weights, \Red{$\boldsymbol{\beta}^\star$,} is bounded above: \Red{$ \| \boldsymbol{\beta}^\star \|_{*} \leq {B}_* < +\infty$}.
\end{assumption}
These assumptions are reasonable as we aim for real-world applications which, by nature, generally satisfy them, \Red{and trained weights are finite.}

As shown previously, the SCNN training problem is similar to a linear regression. We can follow the procedure used in~\cite{Chen2020_DRO} for linear regressions to bound the performance of WaDiRo-SCNNs. 

\begin{proposition}\label{pr:osp1}
    Under Assumptions \ref{ass:z_iid}, \ref{as:boundZ}, and \ref{as:boundBeta}, and considering an $\ell_1$-loss function, with probability at least $1-\delta$ with respect to the sampling, the expected {out-of-sample error} of WaDiRo-SCNNs on $\mathbbm{F}^\mathcal{S}$ is:
    \begin{alignat*}{1}
    &\mathbb{E}^{\mathbbm{F}^{\mathcal{S}}}\left[\left|{\boldsymbol{\beta}}^{\star\top } {\hat{\mathbf{z}}} \right|  \middle| \mathcal{S} \right] \leq \frac{1}{N} \sum_{j=1}^N\left|{\boldsymbol{\beta}}^{\star \top} {\hat{\mathbf{z}}}_j\right|+\frac{2 {B}_* \hat{R} }{\sqrt{N}}+ B_* \hat{R} \sqrt{\frac{8 \ln (2 / \delta)}{N}},%\tag{\texttt{WaDiRo-OSP}}
    \end{alignat*}
    for a modified sample $\hat{\mathbf{z}}$ from the true distribution, that was not available for training, where ${\boldsymbol{\beta}}^{\star}$ are \blue{trained} weights obtained from \blue{either}~\eqref{eq:WDR1-SCNNT}or~\eqref{eq:WDR2-SCNNT}). 

Moreover, for any $\zeta>\frac{2 {B}_* \hat{R} }{\sqrt{N}}+{B}_* \hat{R} \sqrt{\frac{8 \ln (2 / \delta)}{N}}$, we have:
 \begin{equation*} \mathrm{Pr}\left(\left|{\boldsymbol{\beta}}^{\star\top } {\hat{\mathbf{z}}} \right| \geq \frac{1}{N} \sum_{j=1}^N\left|{\boldsymbol{\beta}}^{\star\top } {\hat{\mathbf{z}}}_j \right|  +\zeta \middle| \mathcal{S}\right) \leq  \frac{\frac{1}{N} \sum_{j=1}^N\left|{\boldsymbol{\beta}}^{\star\top } {\hat{\mathbf{z}}}_j \right|  +\frac{2 {B}_* \hat{R} }{\sqrt{N}}+  B_* \hat{R} \sqrt{\frac{8 \ln (2 / \delta)}{N}}}{\frac{1}{N} \sum_{j=1}^N\left|{\boldsymbol{\beta}}^{\star\top } {\hat{\mathbf{z}}}_j \right|  +\zeta}.\end{equation*}
\end{proposition}

\textit{Proof:}  By Lemma \ref{lem:linreg}, the WaDiRo-SCNN's training loss function with respect to $\hat{\mathbbm{F}}^\mathcal{S}_N$ belongs to the following family: 
\begin{equation*} \mathcal{H}=\left\{(\mathbf{x}, y) \to h_{\boldsymbol{\beta}} (\mathbf{x}, y): h_{\boldsymbol{\beta}} (\mathbf{x}, y) = \left|y-\mathbf{x}^{\top} \boldsymbol{\beta} \right|\right\}.\end{equation*}
Thus, the bounds follow from Theorem 4.3.3 of \cite{Chen2020_DRO}. \hfill $\square$

\color{black}
Interestingly, we can obtain similar performance bounds with respect to the original distribution~$\mathbbm{P}$ instead of the modified one. We now make further assumptions.

\begin{assumption}\label{as:boundss}
    The norm of the original samples, features, and labels are bounded above: $\|\mathbf{z}\| \leq R < +\infty, \ \|\mathbf{x}\| \leq S < +\infty, \text{and }\|\mathbf{y}\| \leq T < +\infty$.
\end{assumption}

Consider the \textit{general} family of loss functions for arbitrary weights $\boldsymbol{\beta}$:
\begin{equation}
    \mathcal{G}=\left\{(\mathbf{x}, y) \mapsto g_{\boldsymbol{\beta}} (\mathbf{x}, y): g_{\boldsymbol{\beta}} (\mathbf{x}, y) = \left|y - {\boldsymbol{\beta}}^\top \hat{\mathbf{x}}\right|, \ \hat{\mathbf{x}} = \left( \vc(\mathbf{x} \mathbf{d}^\top)^\top, \vc(\mathbf{x} \mathbf{d}^\top)^\top\right) \right\},
\end{equation}
where $\mathbf{d}^\top = \begin{bmatrix}
        \mathbbm{1}(\mathbf{x}^\top \mathbf{s}_1 \geq 0) & \mathbbm{1}(\mathbf{x}^\top \mathbf{s}_2\geq 0)  & \ldots & \mathbbm{1}(\mathbf{x}^\top \mathbf{s}_P\geq 0)
    \end{bmatrix}$ and $\mathbf{s}_i \in \mathcal{S} \ \forall i \in \llbracket P \rrbracket$ are \textit{general} sampling vectors.

Consider $\psi_1 = B_* 2PR$ for \eqref{eq:WDR1-SCNNT} and $\psi_2 = B_* \sqrt{2P}R$ for~\eqref{eq:WDR2-SCNNT}, upper bounds on the prediction error of the WaDiRo-SCNN for any sample from the sample space. Then, we can rewrite the out-of-sample performance of trained WaDiRo-SCNNs with respect to  $\mathbbm{P}^\mathcal{S}$, the conditioned underlying distribution, as follows.

\begin{proposition}\label{pr:osp2}
    Under Assumptions \ref{ass:z_iid}, \ref{as:boundBeta}, and \ref{as:boundss}, and considering an $\ell_1$-loss function, with probability at least $1-\delta$ with respect to the sampling, the expected {out-of-sample error} of WaDiRo-SCNNs on $\mathbbm{P}^\mathcal{S}$ is:
    \begin{alignat}{1}
    &\mathbb{E}^{\mathbbm{P}^{\mathcal{S}}}\left[\left|{\boldsymbol{\beta}}^{\star\top } {\hat{\mathbf{z}}} \right|  \middle| \mathcal{S} \right] \leq \frac{1}{N} \sum_{j=1}^N\left|{\boldsymbol{\beta}}^{\star \top} {\hat{\mathbf{z}}}_j\right|+\frac{2 \psi_l }{\sqrt{N}}+ B_* \hat{R} \sqrt{\frac{8 \ln (2 / \delta)}{N}}, \label{eq:rademachE}%\tag{\texttt{WaDiRo-OSP}}
    \end{alignat}
    for a modified sample $\hat{\mathbf{z}}$ from the true distribution given $\mathcal{S}$, that was not available for training, where ${\boldsymbol{\beta}}^{\star}$ are the optimal weights obtained from either~\eqref{eq:WDR1-SCNNT}) or~\eqref{eq:WDR2-SCNNT}) and where $l\in\{1,2\}$. 
\end{proposition}

\textit{Proof:} We first bound the value of the loss function for a single arbitrary sample. For~\eqref{eq:WDR2-SCNNT} the dual norm of the weights is given by $\|\cdot \|_2$, thus:
\begin{alignat*}{2}
    |\boldsymbol{\beta}^\top \hat{\mathbf{z}}| &\leq \|\boldsymbol{\beta}\|_2 \|\hat{\mathbf{z}}\|_2 &\text{(Hölder's inequality)}\\
    &\leq B_* \sqrt{\sum_{i=1}^{P} \sum_{k=1}^d 2 ({x}_k \mathbbm{1}(\mathbf{x}^\top \mathbf{s}_i \geq 0))^2 + y^2} & \\
    &\leq B_* \sqrt{\sum_{i=1}^{P} \sum_{k=1}^d 2 ({x}_k)^2 + y^2} & \text{(Full activation)}\\
    &= B_* \sqrt{2P\|\mathbf{x}\|_2^2 + y^2} & \\
    &\leq B_* \sqrt{2PS^2 + T^2} &\\
    &\leq B_* \sqrt{2P} R &\\
    &=\psi_2. & \text{(Definition)}
\end{alignat*}
For~\eqref{eq:WDR1-SCNNT}, the dual norm is the $\|\cdot\|_\infty$-norm. Hölder's inequality similarly yields:
\begin{alignat*}{2}
    |\boldsymbol{\beta}^\top \hat{\mathbf{z}}| &\leq \|\boldsymbol{\beta}\|_\infty \|\hat{\mathbf{z}}\|_1 &\text{(Hölder's inequality)}\\
    &\leq B_* \sum_{i=1}^{P} \sum_{k=1}^d 2 |({x}_k \mathbbm{1}(\mathbf{x}^\top \mathbf{s}_i \geq 0))| + |y| & \\
    &\leq B_* (2P\|\mathbf{x}\|_1 + |y|) & \\
    &\leq B_* (2PS + T) &\\
    &\leq B_* 2PR &\\
    &= \psi_1. & \text{(Definition)}
\end{alignat*}
We now bound the Rademacher complexity $\mathcal{R}_N$ over $N$ samples for the family of functions $\mathcal{G}$  using the definition provided by \cite{Bartlett2003RademacherAG}:
\begin{alignat*}{2}
    \mathcal{R}_N(\mathcal{G}) &= \mathbb{E} \left[ \mathbb{E} \left[ \sup_{g\in\mathcal{G}} \frac{2}{N} \left|\sum_{j=1}^N \sigma_j g_\beta(\mathbf{x}_j, y_j) \right| \big| (\mathbf{x}_1, y_1),  \ldots, (\mathbf{x}_N, y_N)\right] \right] &\\
    &\leq \mathbb{E}\left[ \frac{2}{N} \left|\sum_{j=1}^N \sigma_j \psi_l \right| \right] &\text{(Upper bound)}\\
    &= \frac{2\psi_l}{N} \mathbb{E}\left[  \left|\sum_{j=1}^N \sigma_j  \right| \right] & \\
    &\leq \frac{2\psi_l}{N} \mathbb{E}\left[  \sqrt{\sum_{j=1}^N \sigma_j^2  } \right] & \text{($\sqrt{N} \geq 0$)}\\
    &= \frac{2\psi_l}{\sqrt{N}},  &
\end{alignat*}
where $\sigma_j \ j \in \llbracket N \rrbracket$ are independent random variables sampled from the set $\{-1, 1\}$, and $l\in\{1,2\}$ accordingly to the training procedure.

Thus, by adapting Theorem 8 of \cite{Bartlett2003RademacherAG} to loss functions mapping between $[0,B_* \hat{R}]$ and using the fact that $\mathcal{R}_N(\mathcal{G})$ is greater than the Rademacher complexity term defined in Theorem 8, we obtain the expression from~\eqref{eq:rademachE}. \hfill $\square$

%We note that the variable $a$ from (\ref{eq:WDR1-SCNNT}) and (\ref{eq:WDR2-SCNNT}) is already an upper bound on the dual-norm of the weights for each problem, its optimal value post-training could then readily be used.

We remark that the expected out-of-sample error reduces to the training error as the number of training samples grows. We also remark that the tighter the bounds are on the sample space, i.e., $\hat{R}, \ R, \ S, \text{ and} \ T$, the tighter the guarantees are. \Red{The} proof of Proposition \ref{pr:osp2} also leads to a probabilistic guarantee similar to the one in Proposition \ref{pr:osp1}.

The theoretical guarantees provided in this section contain a probabilistic term and may be difficult to leverage empirically. Nonetheless, there is value in knowing that they exist when deploying complex ML models such as ours in critical applications because of the general lack of trust towards them. They also indicate how one can increase reliability by making these bounds tighter. 
\color{black}

\section{Physics-constrained shallow convex neural networks}\label{sec:pcscnn}
Compared to traditional physics-informed neural networks~\citep{cuomo2022scientific}, because the training problem is a convex program, knowledge of the system can be introduced as hard, convex constraints in the training phase. Let $\hat{y}_j = \left( \vc(\boldsymbol{\nu}),\vc(-\boldsymbol{\omega}) \right)^\top \hat{\mathbf{x}}_j, \ j \in \llbracket N \rrbracket$ \blue{be the prediction of the SCNN for a specific modified input vector}.  We formulate the physics-constrained SCNN (PCSCNN) training problem as:
\begin{alignat}{3}
   &\min_{\boldsymbol{\nu}, \boldsymbol{\omega}} && \sum_{j=1}^{N} \mathcal{L}\left(\hat{y}_j, y_j   \right) \tag{\texttt{PCSCNNT}} \label{eq:pcscnnt}\\
   & \;\;\text{s.t.} && \hat{y}_j \in \mathcal{Y}_j && \forall j \in \llbracket N \rrbracket \nonumber\\
   & && \boldsymbol{\nu}_i,\boldsymbol{\omega}_i \in \mathcal{K}_i \ &&\forall i \in \llbracket P \rrbracket, \nonumber
\end{alignat}
where $\mathcal{Y}_j \subseteq \mathbb{R}$ is the convex intersection of all physical constraints for sample $j \in \llbracket N \rrbracket$. The constraints are guaranteed to be respected during training, which is generally not \blue{true} in common physics-informed NNs. We illustrate some convex constraints with examples:
\begin{alignat}{2}
  & l_j \leq \hat{y}_j \leq u_j && \forall j \in \llbracket N \rrbracket \label{eq:const_bounds} \\
  & |\hat{y}_{j} - \hat{y}_{j-1}| \leq r_j && \forall j \in \llbracket N - 1 \rrbracket \label{eq:const_ramp} \\
  & \underline{\xi}_j \leq \sum_{i=1}^{j} \boldsymbol{\eta}_i^\top {\mathbf{x}}_i + \sum_{i=1}^{j} {\gamma}_i \hat{y}_i \leq \bar{\xi}_j \quad && \forall j \in \llbracket N \rrbracket, \label{eq:const_linear}
\end{alignat}
where in~\eqref{eq:const_bounds}, \eqref{eq:const_ramp}, and~\eqref{eq:const_linear}, with respect to sample $j$ and $\forall i \in \llbracket j \rrbracket$, $u_j \in \mathbb{R}$ is an upper bound, $l_j \in \mathbb{R}$ is a lower bound,  and $r_j \in \mathbb{R}$ is a ramping bound. Additionally, \eqref{eq:const_linear} is a bounded first-order relation between input and output history with $\underline{\xi}_j \in \mathbb{R}, \bar{\xi}_j \in \mathbb{R}, \boldsymbol{\eta}_i \in \mathbb{R}^d, \gamma_i \in \mathbb{R}$, e.g., a discrete-time dynamical equation \Red{or PDE}. \blue{These examples are linear or easily linearizable but we stress that any convex constraint can be included. In the case of non-differentiable convex functions, the numerical solver must be chosen accordingly or a subgradient method be implemented~\cite{boyd2014subgradient}.} We remark that the WaDiRo-SCNN training problem can be similarly modified to consider physical constraints by adding $\hat{y}_j \in \mathcal{Y}_j \ \forall j \in \llbracket N \rrbracket$ to the constraints of~\eqref{eq:WDR1-SCNNT} and \eqref{eq:WDR2-SCNNT}.\color{black}
\begin{proposition}\label{pr:pcscnn}
    Assume that mappings~\eqref{eq:map1} and~\eqref{eq:map2} are exact, i.e., $m \geq m^\star$, then \eqref{eq:pcscnnt} is exactly equivalent to adding convex physical constraints to~\eqref{eq:SNNT} for any additive convex loss function $\mathcal{L}$, e.g., $\|\cdot\|_1 $ and $\|\cdot\|_2^2$.
\end{proposition}
\textit{Proof:} To prove Proposition \ref{pr:pcscnn} we utilize the exact mappings of the original problem and show that the integration of physical constraints into~\eqref{eq:SNNT} yields to~\eqref{eq:pcscnnt}. We start by stating the physics-constrained non-convex problem:
\begin{alignat*}{2}
    p^\star_{\mathcal{Y}} = &\min_{\mathbf{W}_{1}, \mathbf{w}_2}  &&\sum_{j=1}^N \mathcal{L}\left( \sum_{i=1}^m \max\left\{\mathbf{x}_j^\top \mathbf{W}_{1i}, 0\right\}w_{2i}, y_j \right)\\
    &\;\;\;\text{s.t.} &&\sum_{i=1}^m \max\left\{\mathbf{x}_j^\top \mathbf{W}_{1i}, 0\right\}w_{2i} \in \mathcal{Y}_j \quad \forall j \in \llbracket N \rrbracket. \\
    \intertext{Substituting mappings~\eqref{eq:map1} and~\eqref{eq:map2}, and adding the cone constraints, we have:}
    p^\star_{\mathcal{Y}}= &\min_{\boldsymbol{\nu}, \boldsymbol{\omega}}  &&\sum_{j=1}^N \mathcal{L}\left( \sum_{i=1}^{m^\star/2} \max\left\{\mathbf{x}_j^\top \frac{\boldsymbol{\nu}_i}{\sqrt{\|\boldsymbol{\nu}_i\|_2}}, 0\right\}\sqrt{\|\boldsymbol{\nu}_i\|_2} + \max\left\{\mathbf{x}_j^\top \frac{\boldsymbol{\omega}_i}{\sqrt{\|\boldsymbol{\omega}_i\|_2}}, 0\right\}(-\sqrt{\|\boldsymbol{\omega}_i\|_2}), y_j  \right)\\
    &\;\;\text{s.t.} &&\sum_{i=1}^{m^\star/2} \max\left\{\mathbf{x}_j^\top \frac{\boldsymbol{\nu}_i}{\sqrt{\|\boldsymbol{\nu}_i\|_2}}, 0\right\}\sqrt{\|\boldsymbol{\nu}_i\|_2} + \max\left\{\mathbf{x}_j^\top \frac{\boldsymbol{\omega}_i}{\sqrt{\|\boldsymbol{\omega}_i\|_2}}, 0\right\}(-\sqrt{\|\boldsymbol{\omega}_i\|_2})  \in \mathcal{Y}_j \quad\forall j \in \llbracket N \rrbracket \\
    & &&(2\mathbbm{1}(\mathbf{x}_j^\top \mathbf{s}_i \geq 0) -1)\mathbf{x}_j^\top \boldsymbol{\nu}_i \geq 0 \quad \forall j \in \llbracket N \rrbracket, \forall i  \in \left\llbracket \frac{m^\star}{2}\right\rrbracket\\
    & &&(2\mathbbm{1}(\mathbf{x}_j^\top \mathbf{s}_i \geq 0) -1)\mathbf{x}_j^\top \boldsymbol{\omega}_i \geq 0 \quad \forall j \in \llbracket N \rrbracket, \forall i  \in \left\llbracket \frac{m^\star}{2}\right\rrbracket.
    \intertext{Using the hyperplane arrangement from which we obtained the possible activation patterns, we can show that if there is an activation for a specific $ij$, then $\max\left\{\mathbf{x}_j^\top \frac{\boldsymbol{\nu}_i}{\sqrt{\|\boldsymbol{\nu}_i\|_2}}, 0\right\}\sqrt{\|\boldsymbol{\nu}_i\|_2} = \mathbf{x}_j^\top \boldsymbol{\nu}_i,\text{ and} \ \max\left\{\mathbf{x}_j^\top \frac{\boldsymbol{\omega}_i}{\sqrt{\|\boldsymbol{\omega}_i\|_2}}, 0\right\} (-\sqrt{\|\boldsymbol{\omega}_i\|_2}) = -\mathbf{x}_j^\top \boldsymbol{\omega}_i$, and otherwise both equals 0. Thus, we can completely substitute the $\ReLU$ operator by $d_{ji} = \mathbbm{1}(\mathbf{x}_j^\top \mathbf{s}_i \geq 0)$, leading to:}
    p^\star_{\mathcal{Y}}= &\min_{\boldsymbol{\nu}, \boldsymbol{\omega}}  &&\sum_{j=1}^N \mathcal{L}\left( \sum_{i=1}^{m^\star/2} d_{ji} \mathbf{x}_j^\top(\boldsymbol{\nu}_i - \boldsymbol{\omega}_i), y_j  \right)\\
    &\;\; \text{s.t.} &&\sum_{i=1}^{m^\star/2} d_{ji} \mathbf{x}_j^\top(\boldsymbol{\nu}_i - \boldsymbol{\omega}_i)  \in \mathcal{Y}_j \quad\forall j \in \llbracket N \rrbracket \\
    & &&(2d_{ji} -1)\mathbf{x}_j^\top \boldsymbol{\nu}_i \geq 0 \quad\forall j \in \llbracket N \rrbracket, \forall i  \in \left\llbracket \frac{m^\star}{2}\right\rrbracket\\
    & &&(2d_{ji} -1)\mathbf{x}_j^\top \boldsymbol{\omega}_i \geq 0 \quad\forall j \in \llbracket N \rrbracket, \forall i  \in \left\llbracket \frac{m^\star}{2}\right\rrbracket,
\end{alignat*}
from which we find~\eqref{eq:pcscnnt}. We remark that a similar approach can be used for the regularized problem \Red{and that the converse of the proposition can be obtained following the same idea.}\hfill $\square$

\color{black}
\color{black}
\section{Post-Training Certification}\label{sec:post}
As mentioned \Red{in Section \ref{sec:intro}}, there is \Red{value} in having ML model architectures that are easily integrable in post-training verification frameworks. \Red{Indeed, PTV enhances trustworthiness in the ML pipeline by providing theoretical certifications for trained models.} \Red{In this section, we demonstrate} how to integrate SCNNs in such frameworks by proposing a local-Lipschitz\Red{-like} stability certification program to evaluate the stability induced by different SCNN training programs, viz.,~\eqref{eq:WDR1-SCNNT},~\eqref{eq:WDR2-SCNNT}, and~\eqref{eq:SCNNT}. \Red{Let $\mathcal{E}\subset \mathbb{R}^d$ be a convex bounded set of input perturbations.} \Red{Our goal is to} certify the worst-case output deviation of \Red{trained} SCNNs when subject to a small \Red{bounded} perturbation $\varepsilon \in \mathcal{E}$ in their feature space \Red{in a way that is agnostic to the training method}. \Red{This will allow us to compare the stability induced by different regularization paradigms in Section 8.3 without mapping the SCNN weights to their non-convex equivalent.} This methodology is inspired by the work of \cite{Huang2021_lipschitzNN} on $\ReLU$ feedforward deep neural networks, but tailored to SCNNs thanks to the decoupled formulation. 

Recall the definition of the Lipschitz constant for a function $f: \mathcal{X} \to \mathbb{R}$. A function is $K$-Lipschitz if there exists a constant $0<K<+\infty$ such that : \Red{$${|f(\mathbf{x}_1) - f(\mathbf{x}_2)|} \leq K {\|\mathbf{x}_1 - \mathbf{x}_2\|},$$}for any $\mathbf{x}_1, \mathbf{x}_2 \in \mathcal{X}$. In our case, we consider only neighbouring points by defining $\mathbf{x}_2 = \{(\mathbf{x}_1 + \boldsymbol{\varepsilon} )\Red{\ \in \mathcal{X}} \ |\ \boldsymbol{\varepsilon} \in \mathcal{E}, \mathbf{x}_1 \in \mathcal{X} \}$, thus the term local-Lipschitz. We define the $\varepsilon$-stability of a function $f$ by $L_\varepsilon$ such that:$$ L_\varepsilon = \max_{\mathbf{x}_1, \boldsymbol{\varepsilon}:\mathbf{x}_1 \in \mathcal{X}, \boldsymbol{\varepsilon}\in\mathcal{E}, \mathbf{x}_1 + \boldsymbol{\varepsilon}\in \mathcal{X}} |f(\mathbf{x}_1) - f(\mathbf{x}_1 + \boldsymbol{\varepsilon})|,$$
which is visibly linked to the local-Lipschitz bound \Red{as $L_\varepsilon$ englobes the product of $K$ and $\|\boldsymbol{\varepsilon}\|$, given the definition of $\mathbf{x}_2$ in the neighbourhood of $\mathbf{x}_1$}.

\begin{proposition}
Let $\boldsymbol{\beta}^{\star}$ be the optimal weights obtained by an SCNN training program and a set of sampling vectors $\mathcal{S}$ obtained before training. \Red{Assuming an SCNN is a finite mapping from a bounded feature space to the label space,} the worst possible output deviation $0 \leq L_\varepsilon < +\infty$ of the SCNN for any perturbation vector $\boldsymbol{\varepsilon}$, from a \Red{bounded} convex set $\mathcal{E}$, on any input from \Red{a bounded convex} feature space \Red{$\mathcal{X}$} can be obtained by a mixed-integer convex program (MICP).
\end{proposition}

\textit{Proof:} \Red{Recall the formulation of the modified samples from Example \ref{ex:mod_samples}.} Let $M\in\mathbb{R}_{>0}$ a large positive scalar\Red{, i.e., a big-$M$ \citep{williams1978model}}, $\mathbf{x} \in \mathcal{X}$ any possible input from feature space $\mathcal{X}\subseteq \mathbb{R}^d$, $\mathbf{x}_\varepsilon \in \mathcal{X}$ a perturbed input \Red{in the neighbourhood of} $\mathbf{x}$. The post-training verification problem can be formulated as:
\begin{alignat}{4}
    L_{\varepsilon} = &\max_{\hat{y}, \hat{y}_\varepsilon, \mathbf{x}, \boldsymbol{\varepsilon}, \mathbf{A}, \mathbf{B}, \boldsymbol{\gamma}, \boldsymbol{\eta}} &&\ &&| \hat{y} - \hat{y}_\varepsilon | &&\\
    &\qquad \; \text{s.t.} \ &&\hat{y} && = \boldsymbol{\beta}^{\star\top} \hat{\mathbf{x}} && \label{eq:begY}\\
    & &&\hat{\mathbf{x}} && = \left(\vc (\mathbf{A}), \vc (\mathbf{A})\right) \quad && \\
    & &&\mathbf{x}^\top \mathbf{s}_i &&\leq M \gamma_i && \forall i \in \llbracket P \rrbracket \label{eq:begd} \\
    & &&\mathbf{x}^\top \mathbf{s}_i && \geq -M(1 - \gamma_i) && \forall i \in \llbracket P \rrbracket \label{eq:endd} \\
    & &&x_k &&\leq a_{ki} + M(1 - \gamma_i) && \forall i \in \llbracket P \rrbracket, \ \forall k \in \llbracket d \rrbracket \label{eq:begA}  \\
    & &&x_k &&\geq a_{ki} - M(1 - \gamma_i) && \forall i \in \llbracket P \rrbracket, \ \forall k \in \llbracket d \rrbracket   \\
    & &&a_{ki} &&\leq M \gamma_i && \forall i \in \llbracket P \rrbracket, \ \forall k \in \llbracket d \rrbracket   \\
    & &&a_{ki} &&\geq -M \gamma_i && \forall i \in \llbracket P \rrbracket, \ \forall k \in \llbracket d \rrbracket \label{eq:endA} \\
    & &&\boldsymbol{\gamma} &&\in \{0,1\}^P \label{eq:endY}\\
    & &&\hat{y}_\varepsilon &&= \boldsymbol{\beta}^{\star\top} \hat{\mathbf{x}}_\varepsilon &&\\
    & &&\hat{\mathbf{x}}_\varepsilon && = \left(\vc (\mathbf{B}), \vc (\mathbf{B})\right) \quad && \label{eq:begYhat} \\
    & &&{\mathbf{x}}_\varepsilon &&= (\mathbf{x} + \boldsymbol{\varepsilon}) \\
    & &&\mathbf{x}_\varepsilon ^\top \mathbf{s}_i &&\leq M \eta && \forall i \in \llbracket P \rrbracket \label{eq:begd_eps}\\
    & &&\mathbf{x}_\varepsilon ^\top \mathbf{s}_i && \geq -M(1 - \eta_i) && \forall i \in \llbracket P \rrbracket \label{eq:endd_eps}\\
    & &&x_{\varepsilon,k} &&\leq b_{ki} + M(1 - \eta_i) && \forall i \in \llbracket P \rrbracket, \ \forall k \in \llbracket d \rrbracket  \label{eq:begB} \\
    & &&x_{\varepsilon,k} &&\geq b_{ki} - M(1 - \eta_i) && \forall i \in \llbracket P \rrbracket, \ \forall k \in \llbracket d \rrbracket   \\
    & &&b_{ki} &&\leq M \eta_i && \forall i \in \llbracket P \rrbracket, \ \forall k \in \llbracket d \rrbracket   \\
    & &&b_{ki} &&\geq -M \eta_i && \forall i \in \llbracket P \rrbracket, \ \forall k \in \llbracket d \rrbracket  \label{eq:endB}\\
    & &&\boldsymbol{\eta} &&\in \{0,1\}^P \label{eq:endYhat}\\
    & &&\mathbf{s}_i &&\in \mathcal{S}  && \forall i \in \llbracket P \rrbracket \\
    & &&\boldsymbol{\varepsilon} &&\in \mathcal{E} \\
   % & &&\mathbf{x}  &&\in {\Xi},\\
    & &&\mathbf{x}, {\mathbf{x}}_\varepsilon \ &&\in \mathcal{X} \label{eq:endend}
\end{alignat}
where $\mathbf{A}= (\mathbf{a}_1,\mathbf{a}_2, \ldots, \mathbf{a}_P)$ and $ \mathbf{B} = (\mathbf{b}_1,\mathbf{b}_2, \ldots, \mathbf{b}_P)$ are intermediate variables \Red{used to represent modified samples in their matrix form before vectorization, see Example \ref{ex:mod_samples}}. \Red{Here, $\mathcal{E}\subset \mathbb{R}^d$ is a bounded convex set that defines the neighbourhood around a given sample $\mathbf{x}$ from the feature space $\mathcal{X}$}, e.g., box constraints on each element or bounded \Red{Euclidean} norm. \Red{The magnitude of $\mathcal{E}$ should be chosen with respect to the application.}

Constraints~\eqref{eq:begY}$-$\eqref{eq:endY} and~\eqref{eq:begYhat}$ - $\eqref{eq:endYhat} are used to represent the $\ReLU$-like activation patterns of the SCNN for the unperturbed and perturbed output, respectively. Specifically, matrices~$\mathbf{A}$ and~$\mathbf{B}$ model \Red{the outer products} $\mathbf{xd}^\top$ and  $\mathbf{x_{\varepsilon}d_{\varepsilon}}^\top$ with constraints~\eqref{eq:begA}$-$\eqref{eq:endA} and~\eqref{eq:begB}$-$\eqref{eq:endB}, respectively. \Red{These constraints ensure that if the product $\mathbf{x}^\top \mathbf{s}_i$ is positive, i.e., $\gamma_i = 1$, then $a_{ki}=x_k, \ \forall k \in \llbracket d \rrbracket$. Otherwise, if $\gamma_i = 0$, then we have $a_{ki}=0, \ \forall k \in \llbracket d \rrbracket$. The process is similar for the product $\mathbf{x}_{\varepsilon}^\top \mathbf{s}_i$ but with matrix $\mathbf{B}$ and binary variable $\boldsymbol{\eta}$, instead.} Constraints~\eqref{eq:begd},~\eqref{eq:endd},~\eqref{eq:begd_eps}, and~\eqref{eq:endd_eps} embed the following definitions: $ \mathbf{d}^\top =\begin{bmatrix}
        \mathbbm{1}(\mathbf{x}^\top \mathbf{s}_1 \geq 0) & \mathbbm{1}(\mathbf{x}^\top \mathbf{s}_2\geq 0) & \ldots & \mathbbm{1}(\mathbf{x}^\top \mathbf{s}_P\geq 0) 
    \end{bmatrix}$ and $\mathbf{d}_\varepsilon^\top =  \begin{bmatrix}
        \mathbbm{1}(\mathbf{x}_\varepsilon^\top \mathbf{s}_1 \geq 0) & \mathbbm{1}(\mathbf{x}_\varepsilon^\top \mathbf{s}_2\geq 0) & \ldots & \mathbbm{1}(\mathbf{x}_\varepsilon^\top \mathbf{s}_P\geq 0) 
    \end{bmatrix}$\Red{, given a known set of sampling vectors $\mathcal{S}$}. \Red{In sum, we use the binary variables $\boldsymbol{\gamma}$ and $\boldsymbol{\eta}$ to model the result of the indicator function for samples $\mathbf{x}$ and $\mathbf{x}_\varepsilon$, respectively.} Finally, the non-concave objective function can be linearized using another binary variable, \Red{$\zeta$}, \Red{and disjunctive constraints, yielding:} \begin{alignat}{3}
    \Red{L_{\varepsilon} =} &\max_{\hat{y}, \hat{y}_\varepsilon, \mathbf{x}, \boldsymbol{\varepsilon}, \mathbf{A}, \mathbf{B}, \boldsymbol{\gamma}, \boldsymbol{\eta}}&& \alpha& \label{eq:alpha_linear_beg} \\ 
        &\qquad \;\text{s.t.} &&\hat{y}_\varepsilon - \hat{y} &&\leq M\zeta \\
        & &&\hat{y}_\varepsilon - \hat{y} &&\geq M(\zeta - 1)\\
        & &&\alpha &&\leq \hat{y}_\varepsilon - \hat{y} + M(1-\zeta)\\
        & &&\alpha &&\geq \hat{y}_\varepsilon - \hat{y} - M(1-\zeta)\\
        & &&\alpha &&\leq \hat{y} - \hat{y}_\varepsilon + M\zeta \\
        & &&\alpha &&\geq \hat{y} - \hat{y}_\varepsilon - M\zeta \\
        & &&\zeta &&\in \{0,1\} \label{eq:alpha_linear_end}\\ 
        & &&\quad \eqref{eq:begY} &&\text{ -- } \eqref{eq:endend}. \notag
    \end{alignat}
    \Red{In \eqref{eq:alpha_linear_beg}$-$\eqref{eq:alpha_linear_end}, we once again use a binary variable to ensure that if $\hat{y}_\varepsilon > \hat{y}$, then $\alpha = \hat{y}_\varepsilon - \hat{y} $; if $\hat{y}_\varepsilon < \hat{y}$, then $\alpha = \hat{y} - \hat{y}_\varepsilon$; and if $\hat{y}_\varepsilon = \hat{y}$, then $\alpha = 0$. Thus, $\alpha$ is equivalent to the absolute value objective. This concludes the proof.} \hfill $\square$

Besides pure certification, this formulation is interesting because it could be used in a pessimistic bilevel program, see \cite{dempe2002foundations}, that integrates SCNN training and certification into the same problem. \Red{We leave this idea as a direction for future research.} We remark that $M$, \Red{as in any big-$M$ formulation used to model disjunctive constraints,} can be tuned separately for each constraint to be as tight as possible \Red{such that each disjunctive constraint holds}. Recent works have greatly accelerated similar procedures for deep $\ReLU$ neural networks, e.g., see \cite{zhou2024scalable}.

\section{Outlier generation on benchmark functions}\label{sec:outliers}
\Red{In this section, we outline the methodology used to generate outlier-filled synthetic datasets for the numerical evaluation of WaDiRo-SCNNs in Section \ref{subsec:synth}. The proposed methodology is original as it leverages an optimal transport-based criterion for outlier generation and does not make any assumption on the data-generating function, allowing for a wide range of possible synthetic datasets.} 

First, to generate empirical distributions with complex patterns, we propose to use the mathematical expressions of traditional non-convex benchmark functions as a starting point. These functions are normally used to evaluate optimization solvers because of their interesting patterns and their generalizability to $n$ dimensions. 

We use a methodology inspired by the work of \cite{pallage2024sliced} \Red{and \cite{pallage2025sliced}} to generate outliers. They propose a method based on the sliced-Wasserstein (SW) distance to remove out-of-distribution outlier points from training sets. In this work, we do the opposite and use a SW-based method to ensure that introduced outliers are out-of-distribution with respect to the original data distribution. 

Computing the Wasserstein distance is generally of high computational complexity, but the one-dimensional Wasserstein distance is an exception as it possesses a closed-form solution. The SW distance \Red{(SWD)} exploits this special case by first computing, through linear projections, an \textit{infinite} amount of one-dimensional representations for each original $n$-dimensional distribution. It then computes the average Wasserstein distance between their one-dimensional projections \citep{NEURIPS2019_f0935e4c}. In practice, the order-$t$ SW distance's approximation under a Monte Carlo scheme of $L$ samples is defined as: 
\begin{equation*}S W_{\|\cdot\|,t}\left(\mathbbm{U}, \mathbbm{V} \right) \approx\left(\frac{1}{L} \sum_{l=1}^L W_{\|\cdot\|,t}\left(\mathfrak{R} \mathbbm{U}\left(\cdot, \theta_l\right), \mathfrak{R} \mathbbm{V}\left(\cdot, \theta_l\right)\right)^t\right)^{1 / t},\end{equation*} 
where $\mathfrak{R} \mathbbm{D}(\cdot, \boldsymbol{\theta}_l)$ is a single projection of the Radon transform of distribution function $\mathbbm{D}$ over a fixed sample $\boldsymbol{\theta}_l \in \mathbb{S}^{d-1} = \{\boldsymbol{\theta} \in \mathbb{R}^d | \theta_1^2 + \theta_2^2 + \cdots + \theta_d^2 = 1\}$ $\forall l \in \llbracket L \rrbracket$. Note that $\mathbb{S}^{d-1}$ is often referred to as the $(d-1)$-dimensional unit sphere as it encloses every $d$-dimensional unit-norm vector in a sphere of radius 1. It can be shown that the WD and SWD are closely related \cite[Theorem 5.1.5]{bonnotte2013unidimensional}. We refer interested readers to \cite{bonneel2015sliced} and \cite{NEURIPS2019_f0935e4c} as they offer deeper mathematical insights.  %Interestingly, it can be shown that:
%LINK BOUND SWD to WD 
%which links the equivalence of both metrics.

Our procedure is as follows. Consider an empirical distribution of $N$ samples $\hat{\mathbbm{P}}_N = \frac{1}{N} \sum_{i=1}^N \delta_{\mathbf{z}_i}(\mathbf{z})$ obtained by sampling uniformally a non-convex benchmark function \Red{on the domain of interest}. \Red{This domain is predetermined.} \Red{Each sample is constructed as a feature and label pair where the features are composed of the uniform samples and the labels are the respective evaluation of the benchmark function on each feature.} \Red{We denote the data-generating distribution as} $\mathbbm{P}$. We use the notation $\hat{\mathbbm{P}}_{N+m}$ to denote the empirical distribution with $m$ added samples from $\mathbbm{P}$ and $\hat{\mathbbm{P}}_N \cup \mathcal{O}_m$ to represent the union of the empirical distribution with a set of $m$ outlier samples $\mathcal{O}_m$. Let $\varphi_m > 0 $ represent a threshold SW distance for sample $m$. The SW distance deals with distributions of equal weights and must account for present outliers when generating a new one. Assuming we want to generate a total of $\overline{m} \in \mathbb{N}_{>0}$ out-of-sample outliers, the iterative procedure is based on:
$$\hat{\mathbbm{P}}_N \cup \mathcal{O}_m =  \left\{ \hat{\mathbbm{P}}_{N}\cup \mathcal{O}_{m-1}\cup \{\tilde{\mathbf{z}}\} \ \middle\vert \tilde{\mathbf{z}} \in \mathcal{Z}, \left( SW_{\|\cdot\|,t} (\hat{\mathbbm{P}}_{N+1}\cup \mathcal{O}_{m-1},\hat{\mathbbm{P}}_{N}\cup \mathcal{O}_{m-1}\cup \{\tilde{\mathbf{z}}\}) \geq \varphi_m \right) \right\} \  m \in \llbracket \overline{m}\rrbracket,
$$
where $\mathcal{O}_0 = \emptyset$ is the empty set. This procedure ensures that each new outlier creates a shift of at least $\varphi$ with respect to the empirical distribution and previous outliers $1, \ 2, \ldots, m-1$. We must consider previous outliers to \Red{avoid creating} any new involuntary patterns in the data. Because we do not make a single assumption on $\mathbbm{P}$, this methodology is adaptable to generate outliers on any function. \Red{This will allow us to test empirically the distributional robustness of different machine learning models against out-of-distribution outliers, with respect to the SW distance, on synthetic datasets with rich patterns.}

\color{black}

\section{Numerical study}\label{sec:numerical}
This section is split into two parts. We first benchmark our approach in a numerical study with data from a controlled environment before showcasing its performance in a real-world \blue{VPP} application.

\subsection{Synthetic experiments}\label{subsec:synth}
\subsubsection{Setting}
\begin{figure}[tb]
    \centering
    \begin{subfigure}{0.23\textwidth}
        \includegraphics[width=\textwidth, trim={2cm, 1cm, 2cm, 2.7cm}, clip]{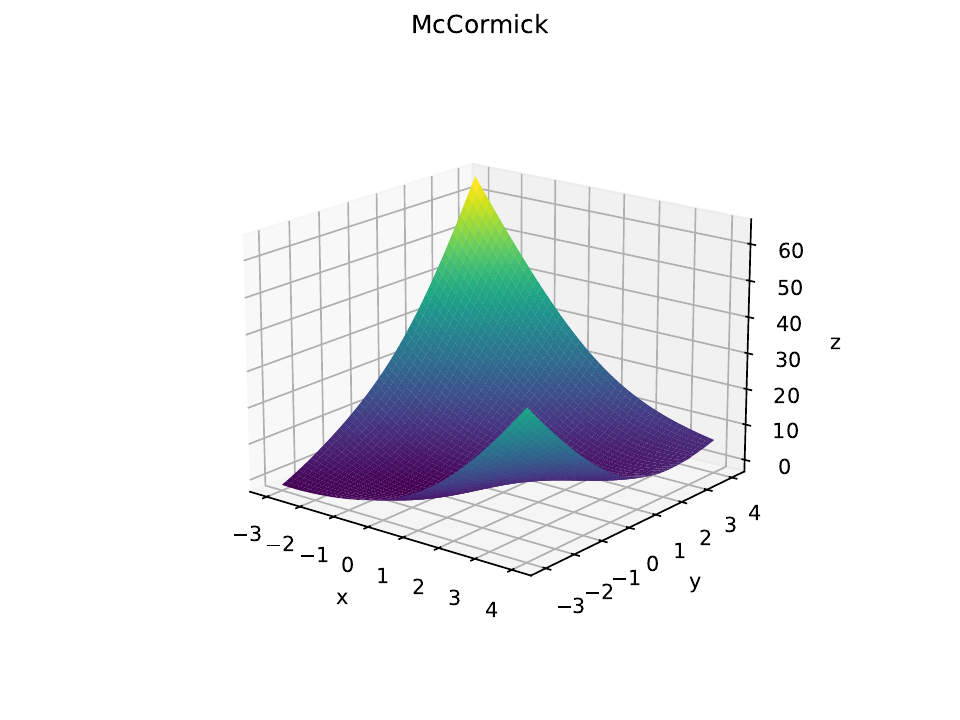}
        \caption{McCormick}
        %\label{fig:sub1}
    \end{subfigure}
    \hfill
    \begin{subfigure}{0.23\textwidth}
        \includegraphics[width=\textwidth, trim={2cm, 1cm, 2cm, 2.7cm}, clip]{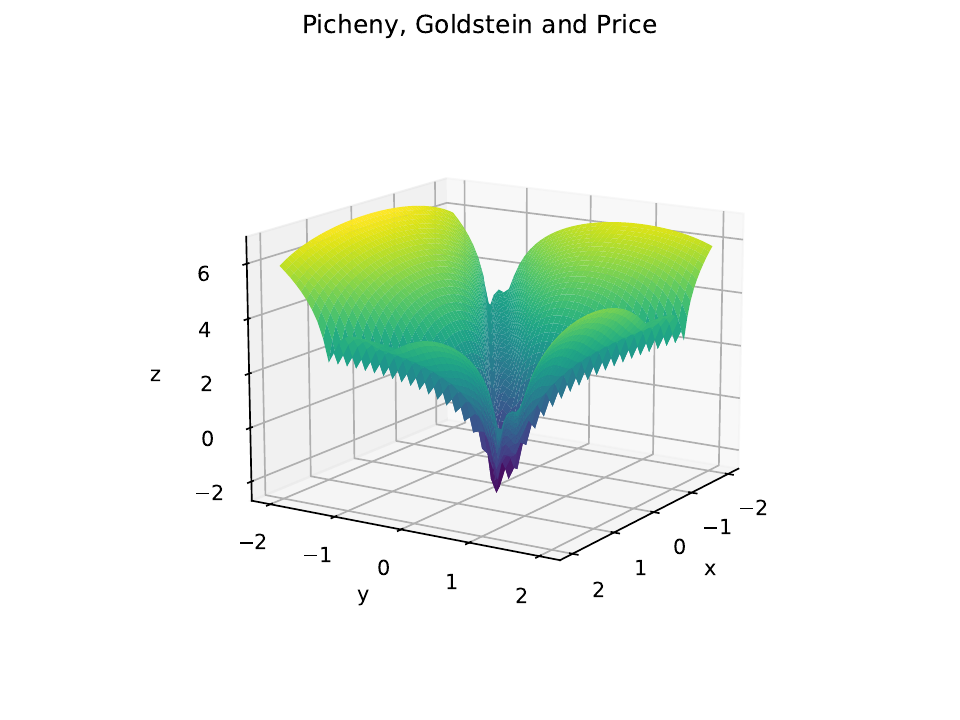}
        \caption{PGP}
        %\label{fig:sub2}
    \end{subfigure}
    \hfill
    \begin{subfigure}{0.23\textwidth}
        \includegraphics[width=\textwidth, trim={2cm, 1cm, 2cm, 2.4cm}, clip]{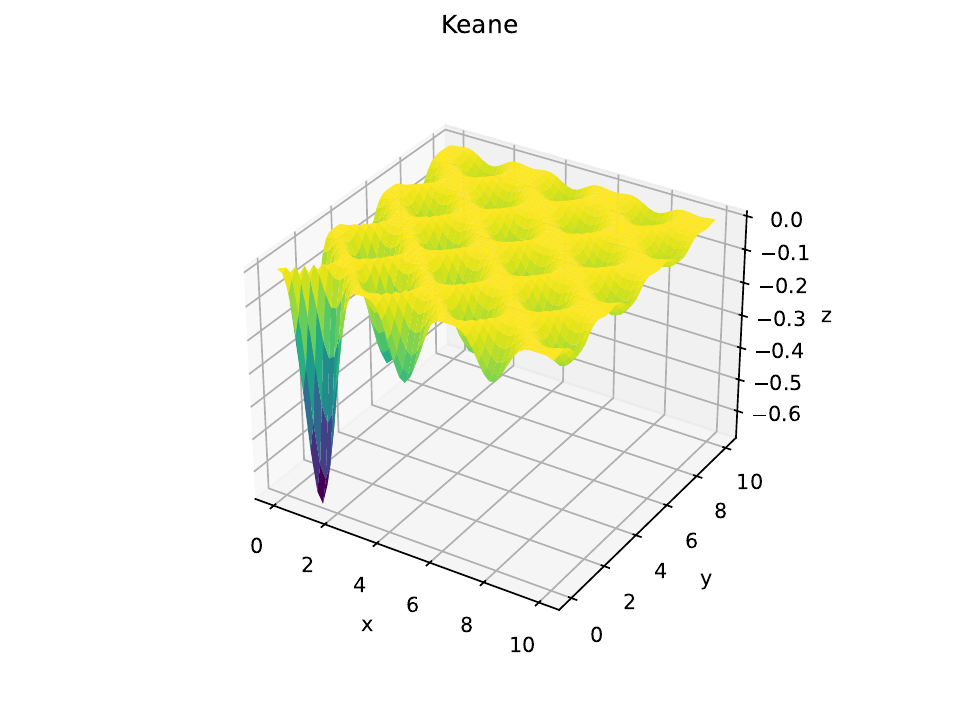}
        \caption{Keane}
        %\label{fig:sub3}
    \end{subfigure}
    \hfill
    \begin{subfigure}{0.23\textwidth}
        \includegraphics[width=\textwidth, trim={2cm, 1cm, 2cm, 2.7cm}, clip]{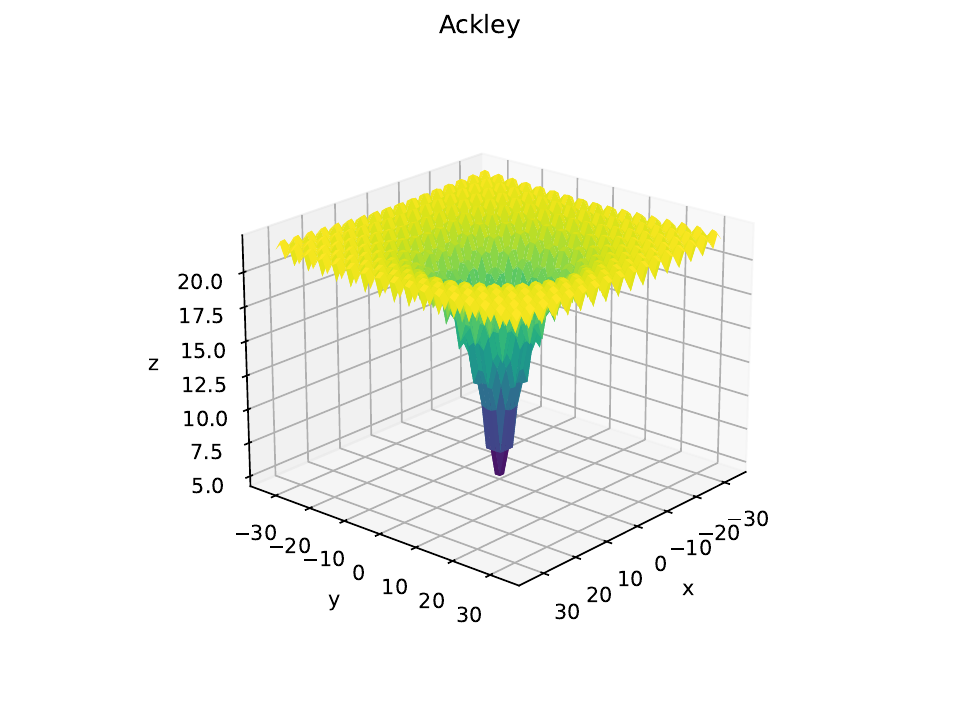}
        \caption{Ackley}
        %\label{fig:sub4}
    \end{subfigure}
    \caption{Benchmark functions}
    \label{fig:whole}
\end{figure}

We aim to assess the performance of our model at learning nonlinear functions in heavily corrupted settings, i.e., at being robust against outliers, with a limited amount of training data. To avoid biasing the underlying distribution, we introduce outliers with the SW distance {as presented in \Red{S}ection \ref{sec:outliers}} and a Gaussian noise, in both the training and validation datasets. We choose traditional non-convex benchmark functions: McCormick~\citep{Adorio2005MVFM_mccormick}, Picheny-Goldstein-Price (PGP)~\citep{picheny2013}, Keane~\citep{keane1994experiences}, and Ackley~\citep{ackley2012connectionist}. These functions strike a balance between complexity and tractability. Despite their non-convexity, they exhibit discernable trends and patterns that facilitate learning compared to some other benchmark functions, e.g., Rana or egg-holder~\citep{vanaret:hal-00996713}. The variation granularity within these functions is moderate ensuring that the learning process is not hindered by excessively fine and extreme fluctuations in the codomain. In a way, this is what we would expect of most real-world regression datasets, e.g., University of California Irvine's ML repository~\citep{markelleUCI}, with the added advantage of being able to control the degree of corruption of the datasets. These functions were generated with the Python library \texttt{benchmark functions}~\citep{baronti2024pythonbenchmarkfunctionsframework} and are illustrated in Figure \ref{fig:whole}. 

We compare our model to an SCNN, a regularized SCNN, a regularized linear regression, a WaDiRo linear regression, and a 4-layer deep $\ReLU$ feedforward neural network (FNN) with batch training and dropout layers. Each of these approaches uses a $\ell_1$-loss function. \blue{WaDiRo-SCNNs and SCNNs are modelled and optimized with \texttt{cvxpy}~\citep{diamond2016cvxpy} and with \texttt{Clarabel} \citep{Clarabel_2024} as the solver. We then extract the weights and encode them into \texttt{pytorch}'s neural network module~\citep{NEURIPS2019_9015}. Similarly, the WaDiRo linear regression and the regularized linear regression are optimized with \texttt{cvxpy} and \texttt{Clarabel}, while the deep FNN is fully implemented with \texttt{pytorch}.} We use simulated annealing~\citep{INGBER199329} for hyperparameter optimization with \texttt{hyperopt}~\citep{bergstra2015hyperopt}. We run 400 trials per ML model, per benchmark function, except for the deep FNN which we limit to 100 trials to mitigate its substantial training time. We split the data such that the training data accounts for 60\% of the whole dataset and we split the rest evenly between the validation and the testing phases. We note that the testing data is uncorrupted to measure the ability of each method to learn the true underlying functions without having access to it. We use the mean average error (MAE) for hyperparameter tuning. We also measure the \blue{root} mean squared error \blue{(RMSE)} for additional performance insights. Our whole experiment setup is available on our GitHub page\footnote{https://github.com/jupall/WaDiRo-SCNN} for reproducibility and is delineated in Appendix \ref{sec:AnnB}. In Experiment A, the outliers are introduced solely in the training data while they are split randomly between training and validation for all the other experiments. Table \ref{tab:noise} details the experiment contexts.

\begin{table}[tb]
\renewcommand{\arraystretch}{1.0}
    \centering
    \caption{Degree of corruption of each experiment}
    \begin{tabular}{c|c|c|c|c|c|c}
    \hline
        
        \hline
       \multirow{2}{*}{\textbf{Experiment}} & \multicolumn{3}{c}{ \textbf{Outlier percentage}} & \multicolumn{3}{c}{ \textbf{Gaussian noise}}  \\  \cline{2-7}
        & \textbf{ Train.}  & \textbf{Val.} & \textbf{Test} & \textbf{Train.} & \textbf{Val.} & \textbf{Test}\\\hline
       A & 0.4 & 0 & 0 & True & True & False \\ \hline
       B & \multicolumn{2}{c|}{ 0.1} & 0 & True & True & False \\ \hline
       C & \multicolumn{2}{c|}{ 0.4} & 0 & True & True & False \\
       \hline
        
        \hline
    \end{tabular}
    \label{tab:noise}
\end{table}

\subsubsection{Numerical Results}\label{subsec:synthres}
\begin{figure}[tb]
    \centering
    \begin{subfigure}{0.6\textwidth}
        \includegraphics[width=\textwidth]{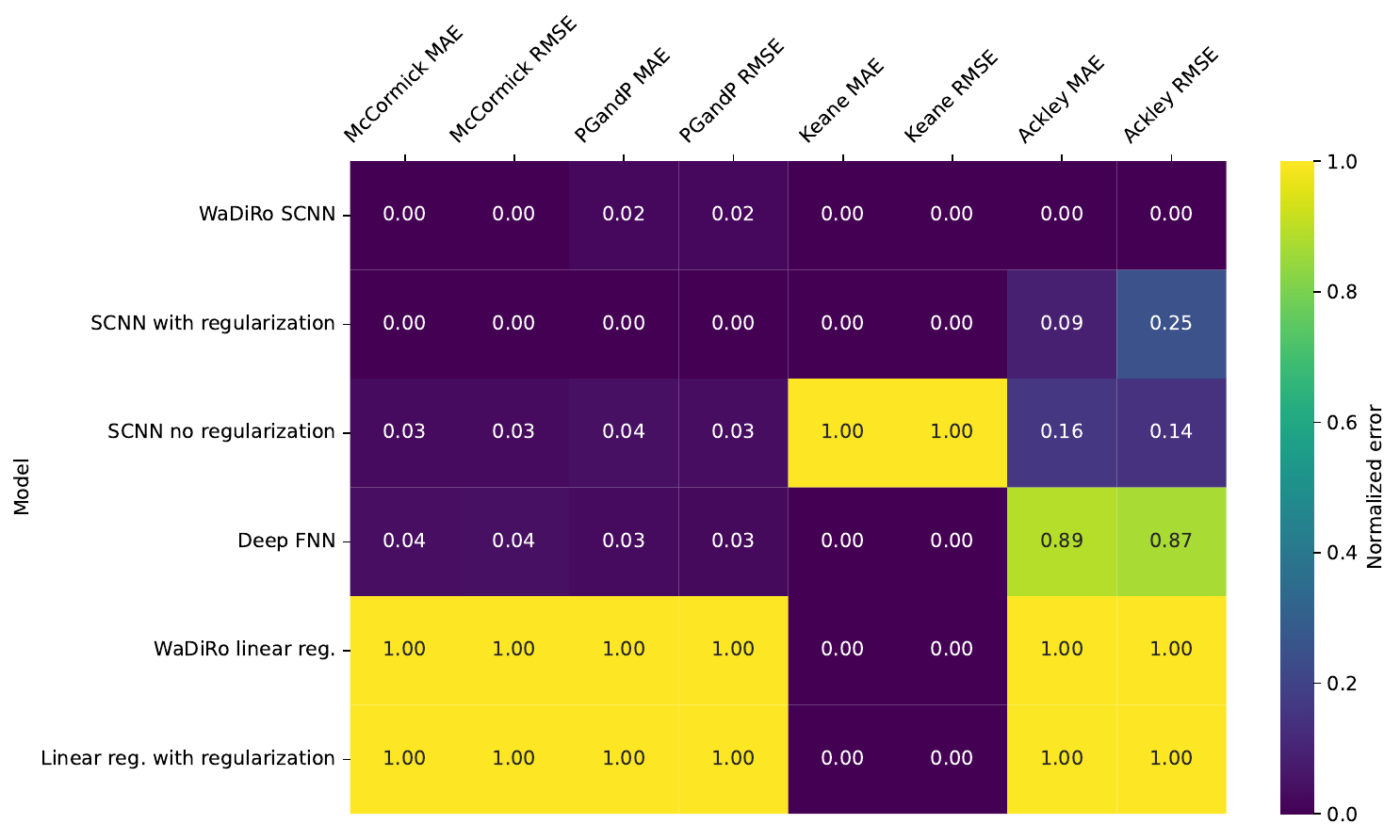}
        \caption{Experiment A}
        %\label{fig:sub1}
    \end{subfigure}
    \hfill
    \begin{subfigure}{0.6\textwidth}
        \includegraphics[width=\textwidth]{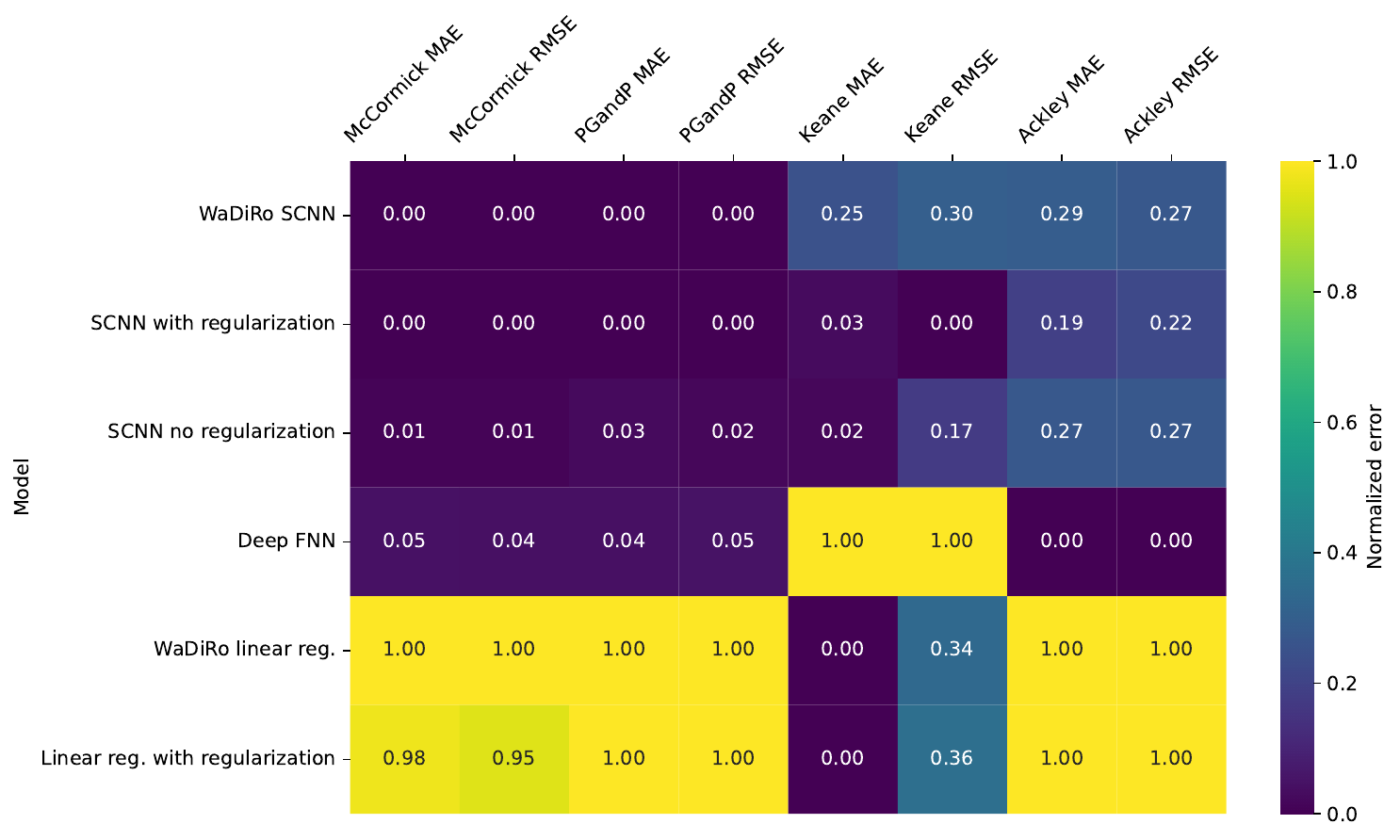}
        \caption{Experiment B}
        %\label{fig:sub2}
    \end{subfigure}
    \hfill
    \begin{subfigure}{0.6\textwidth}
        \includegraphics[width=\textwidth]{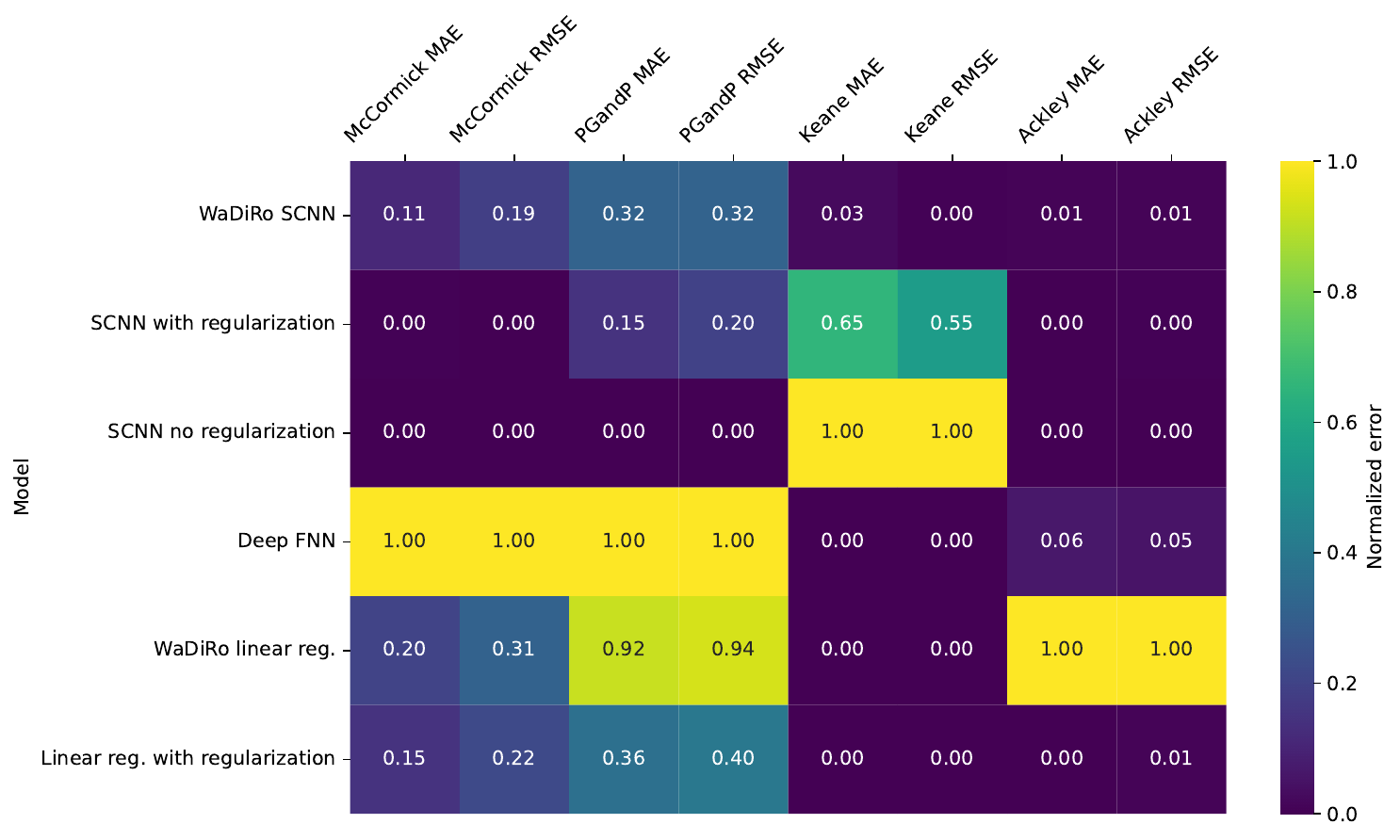}
        \caption{Experiment C}
        %\label{fig:sub3}
    \end{subfigure}
    
    \caption{Normalized error of each synthetic experiment}
    \label{fig:expsynthNorm}
\end{figure}

We now present and analyze the results of each synthetic experiment. Figure \ref{fig:expsynthNorm} shows the normalized error metrics of each model on each function. The normalization is done within each column so that 1.00 and 0.00 represent, respectively, the highest and lowest error between all models for a specific metric and function, e.g., the MAE on McCormick's function. \blue{We propose this representation because different functions have patterns of different complexity and it highlights how models relatively compare to each other when facing the same level of pattern complexity.  } \blue{Nonetheless,} the absolute values are also presented in Appendix \ref{sec:AnnB} as a complement.

The objective of Experiment A is to evaluate the ability of each model at learning over a highly adversarial training set while having the possibility to refine, indirectly through the validation phase, on a slightly out-of-sample distribution. This experiment was designed to see how different models' regularization techniques may prevent them from overfitting on training outliers. As we can see, the WaDiRo-SCNN, regularized SCNN, and deep FNN perform similarly except when trying to learn Ackley where the WaDiRo-SCNN has the lowest test error. In this experiment, the linear models perform poorly compared to other models. We hypothesize that the validation phase is uncorrupted enough to let nonlinear models adjust to the non-convex functions during hyperparameter optimization. As expected the non-regularized SCNN leads to performance lower than its regularization counterpart in this particular setting.

In Experiments B and C, outliers are randomly split between training and validation. This implies that models may overfit on outliers during the hyperparameter optimization. Interestingly, in a less corrupted setting, i.e., in Experiment B, we observe that the best models are the nonlinear ones. Yet, when increasing the corruption level, i.e., the ratio of outliers in the data, the linear models become competitive. In both experiments, the WaDiRo-SCNN's error is low while not necessarily being the lowest of all models. This is expected as the DRO framework is conservative by definition, maybe even more than other regularization methodologies. This conservatism is what makes it appealing for applications in critical sectors.

We remark that the deep FNN often excels but is subject to its highly stochastic training as there is no guarantee that a good model will be found within the 100 runs made for hyperparameter tuning. Other models, having low-stochasticity training, are easier to tune. We also remark that the non-regularized SCNN performs well but lacks the constancy of its regularized counterparts throughout the experiments.

\subsection{Baseline estimation of non-residential buildings}\label{subsec:real}
\subsubsection{Setting}
%In 2023, commercial and industrial sectors together represented 49.2\% of Québec's total energy consumption~\citep{whitmore2024energie}. To accelerate the deployment of Québec's virtual power plant (VPP), there is need for improving forecasting algorithms for individual non-residential buildings. These algorithms can be used for operations and planning decision-making but also for retroactively calculating rewards associated with shaved energy during demand response events~\citep{albadi2007demand}. Non-residential buildings comprise public buildings, e.g., public libraries, schools, and institutional buildings; commercial buildings, e.g., shops, cinemas, and offices; and industrial buildings, e.g., factories and warehouses.
%, yearly temperature roughly ranges from $-30^{\circ}C$ to $+30^{\circ}C$ ($-22^{\circ}F$ to $86^{\circ}F$). 

Montréal's continental humid climate delivers harsh winters and hot summers. As such, a significant amount of energy is consumed for both electric heating and air conditioning. Moreover, date and time features have a strong influence on non-residential buildings' consumption patterns as their activities may change within days and weeks, e.g., opening hours, occupancy, and seasonal activities.

In this experiment, we seek to predict the year-round hourly energy consumption \blue{in} kWh of 16 \blue{distinct} non-residential buildings located in three adjacent distribution substations in Montréal, Québec, Canada. \Red{These buildings are enrolled in a non-residential DR program. Being able to predict the baseline consumption of these assets under different contexts allows VPPs to better characterize grid needs through time and to retroactively estimate the relief given by each building during DR events.} As we do not have any information on building activity, this task is done using only hourly consumption history and historical meteorological features. We use a 12-week history to train each model to predict the following week. We have approximately 2.5 years of collected data on each building. We proceed with a 12-week sliding window and collect the predictions for each new week. We use the data of the first year and a half for hyperparameter tuning and the data of the last year for performance assessment. The features and label are listed in Table \ref{tab:features}.
%['day_of_week_sin', 'day_of_week_cos', 'hour_sin', 'hour_cos', 'solar_rad', 'wind_spd', 'temp', 'weekend_holiday_fix'] 
\begin{table}[tb]
\renewcommand{\arraystretch}{1.0}
    \centering
    \caption{Features and label used for the baseline prediction of non-residential buildings}
    \begin{tabular}{m{3em} | m{16em} | c}
    \hline 

    \hline 
       \textbf{Feature} & \textbf{Name} & \textbf{Domain} \\ \hline
       1 & Day of the week, cosine encoding & $[0,1]$ \\ %\hline
       2 & Day of the week, sine encoding & $[0,1]$\\ %\hline
       3 & Hour of the day, cosine encoding & $[0,1]$\\ %\hline 
       4 & Hour of the day, sine encoding & $[0,1]$\\ %\hline 
       5 & Week-end and holiday flag & $ \{0,1\}$\\ %\hline 
       6 & Average outside temperature [$^\circ$C] & $ \mathbb{R}$\\ %\hline
       7 & Average solar irradiance [W/m$^2$] & $ \mathbb{R}_+$\\ %\hline 
       8 & Average wind speed [m/s] & $ \mathbb{R}_+$\\ 
       \hline 

    \hline
    \end{tabular}\vspace{1em}
    \begin{tabular}{m{3em} | m{16em} | c}
    \hline 

    \hline
    \textbf{Label} & \textbf{Name} & \textbf{Domain} \\ \hline
    1 & Hourly energy consumption [kWh] & $ \mathbb{R}_+$ \\ 
    \hline 

    \hline
    \end{tabular}
    \label{tab:features}
\end{table}
We set cyclical features with a $\cos / \sin$ encoding similarly to~\citep{moon2018forecasting} and introduce a binary flag to indicate weekends and public holidays. \Red{The cyclical encoding consists of projecting each feature space onto the unit circle to model proximity between lower and higher values of a temporal feature, e.g., 23:00 is close to 1:00, but this is not represented numerically as it is.} The dataset is created from proprietary data obtained through our collaboration with Hydro-Québec.

To follow the literature, we compare the WaDiRo-SCNN to a Gaussian process (GP), a WaDiRo linear regression, and a linear support vector regression (SVR).  \blue{Both the GP and the linear SVR were implemented with \texttt{scikit-learn}~\citep{scikit-learn} while the WaDiRo-SCNN and WaDiRo linear regression have been implemented similarly to the previous experiment.} We refer readers to Appendix \ref{sec:AnnC} for details on each model's hyperparameter search space and the definition of the GP kernel. We limit the hyperparameter tuning to 15 runs per model to represent industrial computational constraints and use the tree-structured Parzen estimator (TPE) algorithm proposed by~\citep{bergstra2011algorithms} to navigate the search space. We have observed experimentally that TPE converges faster than simulated annealing to good solutions, which suits the limit on runs. The mean average error is used to guide the hyperparameter tuning. We also test a simple physics-constrained WaDiRo-SCNN in which we enforce that energy consumption must be greater or equal to zero.

\blue{With this experiment, we first aim to showcase how our models perform on real-world VPP data compared to the forecasting models actually used in critical industrial VPP applications. We also seek to demonstrate how the inclusion of hard physical constraints in the SCNN training affects the solution.  We acknowledge that these models are applied naively and that more sophisticated methods could be preferable for this specific application. }

\subsubsection{Numerical Results}

\begin{figure}[tb]
    \centering
    \begin{subfigure}{0.48\textwidth}
        \includegraphics[width=\textwidth]{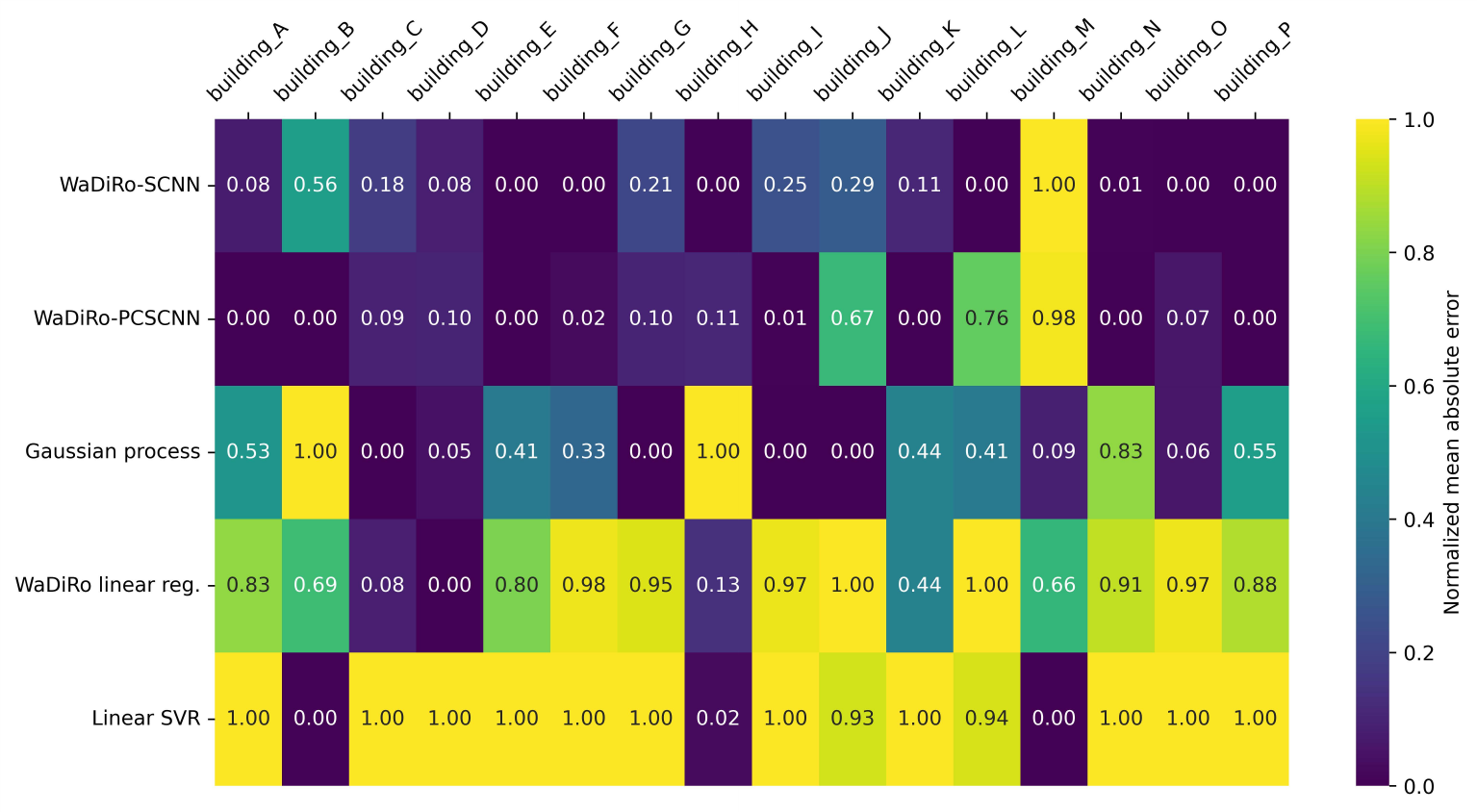}
        \caption{Normalized MAE}
        %\label{fig:sub1}
    \end{subfigure}
    \hfill
    \begin{subfigure}{0.48\textwidth}
        \includegraphics[width=\textwidth]{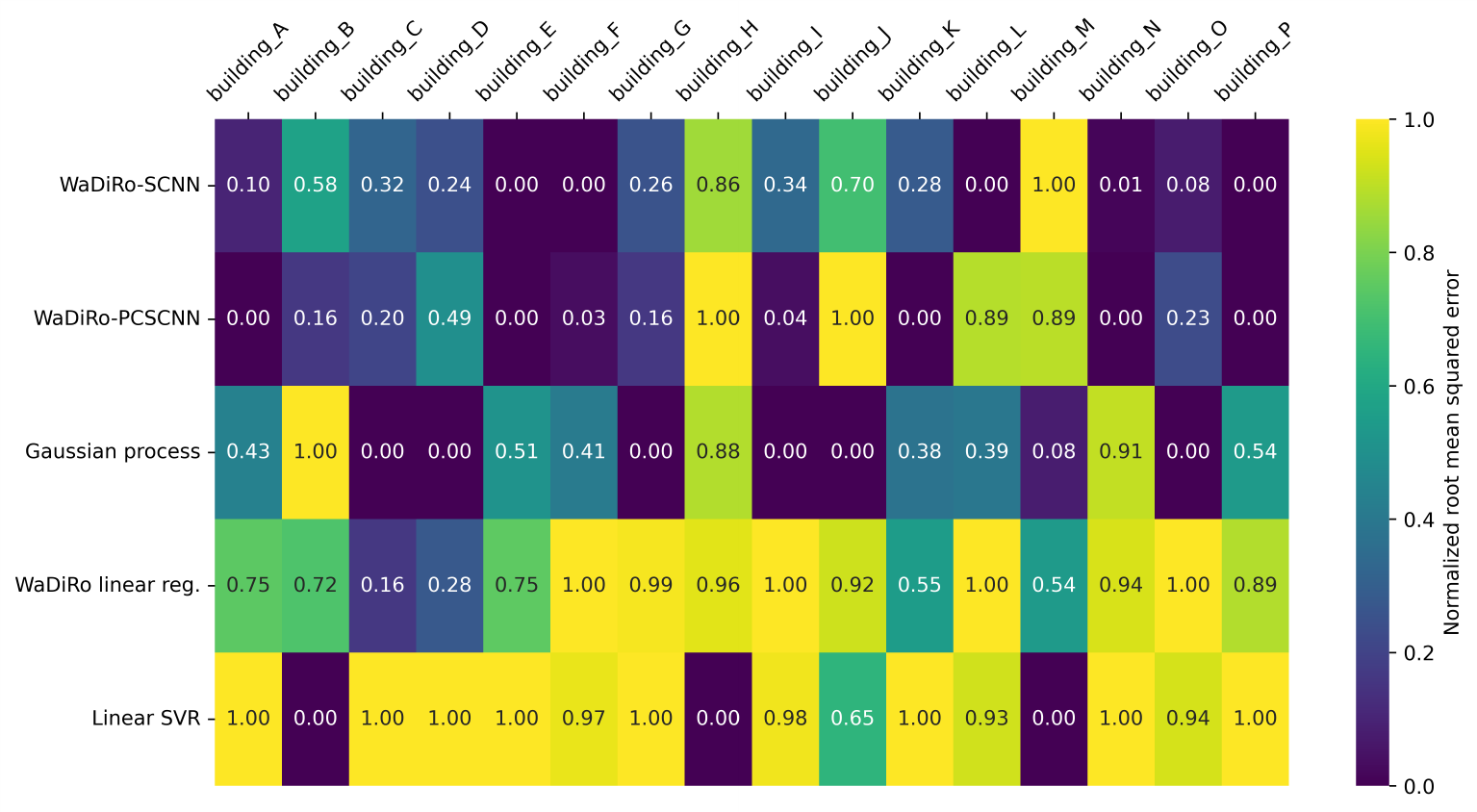}
        \caption{Normalized RMSE}
        %\label{fig:sub2}
    \end{subfigure}
    \caption{Normalized errors of the hourly baseline prediction of non-residential buildings}
    \label{fig:expbuildNorm}
\end{figure}

We now proceed with the analysis of our numerical experiments on the prediction of buildings' consumption patterns. Figure \ref{fig:expbuildNorm} presents the normalized testing errors of each algorithm on each of the 16 tested buildings. 

Table \ref{tab:violations} presents the number of test samples violating the constraint, on non-negative energy consumption, for each model over all buildings and the percentage of deviation from the best model.
\begin{table}[tb]
\renewcommand{\arraystretch}{1.0}
    \centering
    \caption{Number of constraint violations in the testing phase for each model}
    \begin{tabular}{r|ccccc}
    \hline 

    \hline 
       \textbf{Model} & \textbf{WaDiRo-PCSCNN} & \textbf{WaDiRo-SCNN} & \textbf{GP} & \textbf{SVR} & \textbf{WaDiRo lin. reg.}\\ \hline
       \textbf{Violations} & \textbf{1431} & 2277 & 2404 & 4625 & 3495 \\
       \textbf{Deviation [\%]} & \textbf{0.00} &  59.12 & 67.99 & 223.20 & 144.23  \\
    \hline 

    \hline 
    \end{tabular}
    \label{tab:violations}
\end{table}
We observe that the WaDiRo-SCNN performs on par with the GP \blue{as both models seem adequate to model nonlinear patterns}. Still, \Red{its performance exceeds that of the GP} as it has the lower MAE between the two on 10 buildings out of 16 while having similar RMSEs. \blue{Note that trained GPs may only converge to local optima.} We observe that the linear SVR has the worst performance of all models as it has a consistently lower performance than the WaDiRo linear regression. We observe no significant performance gap between the unconstrained and physics-constrained WaDiRo-SCNNs. \blue{We also remark, as shown in Table \ref{tab:violations}, that the WaDiRo-PCSCNN has the lowest number of constraint violations throughout the experiment and that the WaDiRo-SCNN has the second lowest amount with yet 59.12\% more.} Additional figures \Red{of} the absolute errors and \Red{the} test prediction plots, are presented in Appendix \ref{sec:AnnC} for completeness.

We remark that both the GP and WaDiRo-SCNNs have different strengths. Our model can include hard physical constraints easily, does not require making assumptions on the type of data seen in training, and is conservative by design. GPs offer empirical uncertainty bounds and can be defined for multiple outputs.
\color{black}
\subsection{Stability certification}\label{subsec:stability}
We now design a final experiment to evaluate the stability of SCNNs trained with varying numbers of neurons and degrees of regularization.
\subsubsection{Setting}  We propose a grid search over the maximal number of neurons and the regularization parameter. We proceed on different popular real-world datasets from the UCI ML repository, namely the energy efficiency dataset from \cite{energy_efficiency_242} and the concrete compressive strength dataset from \cite{concrete_compressive_strength_165}. Additional tests on other real-world datasets are found on our GitHub page. The random seed for the sampling of $\mathcal{S}$ is fixed throughout the experiments, meaning that any instability that may arise from the sampling will be highlighted similarly for each model. We collect predictions in the SCNN form without mapping back to the non-convex SNN. We provide testing MAEs as well. Both the stability bounds and the MAEs are in the scale of the original dataset, but a standard scaler is used during training. As such, the inverse transform from the normalized outputs to the original scale may have an additional effect on the size of $L_\varepsilon$. 

\subsubsection{Numerical results}
\begin{figure}[tb]
    \centering
    \begin{subfigure}{0.48\textwidth}
        \includegraphics[width=\textwidth]{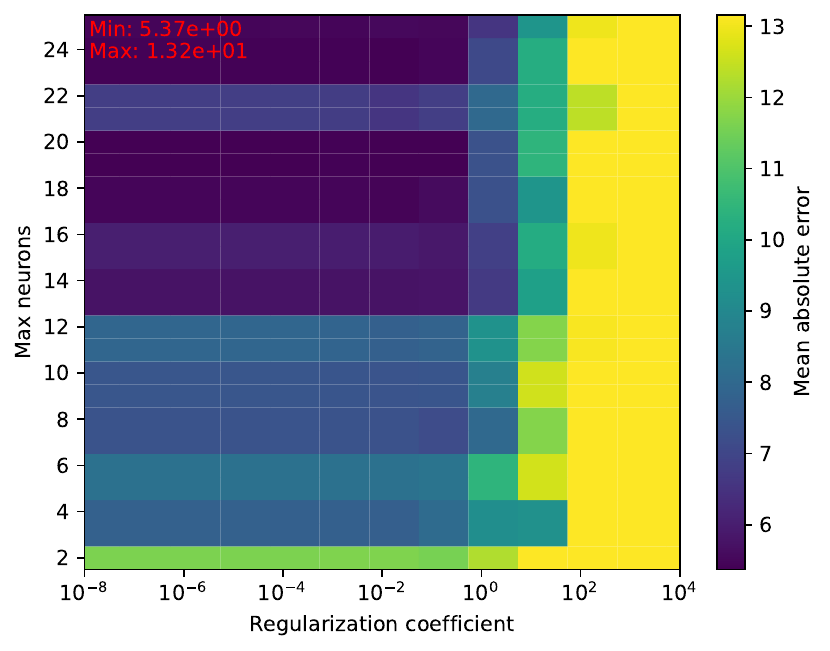}
        \caption{MAE test: LASSO-SCNN}
        %\label{fig:sub1}
    \end{subfigure}
    \hfill
    \begin{subfigure}{0.48\textwidth}
        \includegraphics[width=\textwidth]{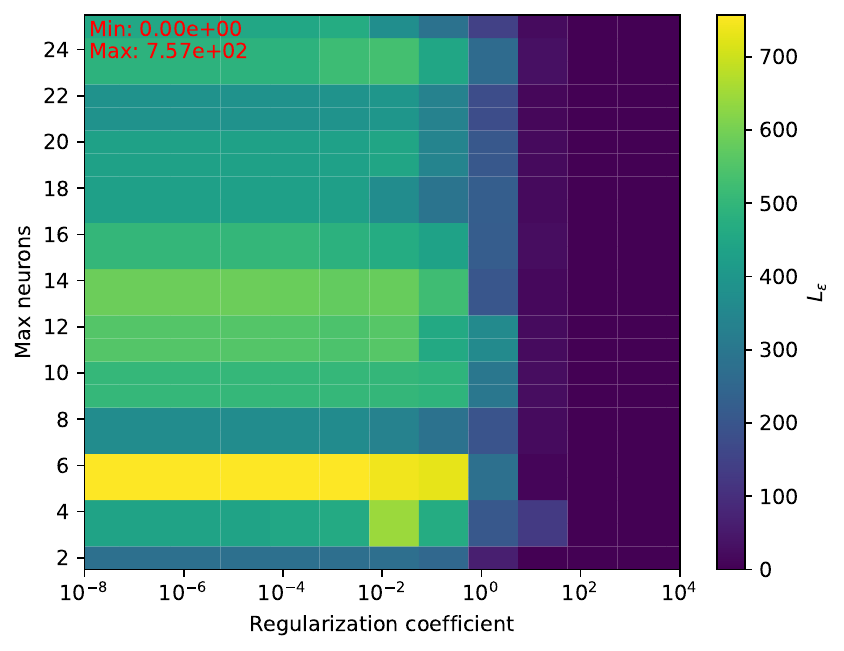}
        \caption{$L_\varepsilon$: LASSO-SCNN}
        %\label{fig:sub2}
    \end{subfigure}
    \hfill
    \begin{subfigure}{0.48\textwidth}
        \includegraphics[width=\textwidth]{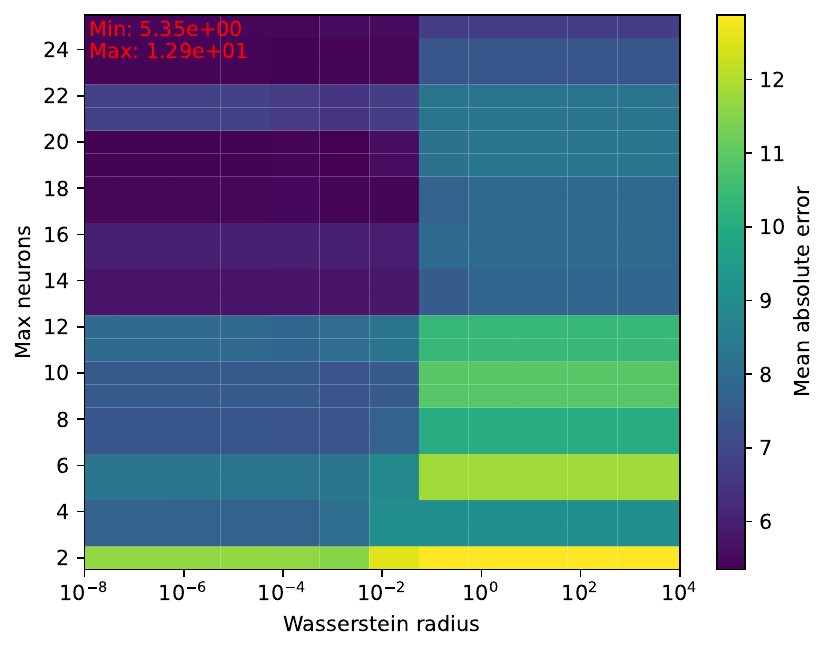}
        \caption{MAE test: WaDiRo-SCNN $\left(W_{\|\cdot\|_1,1}\right)$}
        %\label{fig:sub3}
    \end{subfigure}
    \hfill 
    \begin{subfigure}{0.48\textwidth}
        \includegraphics[width=\textwidth]{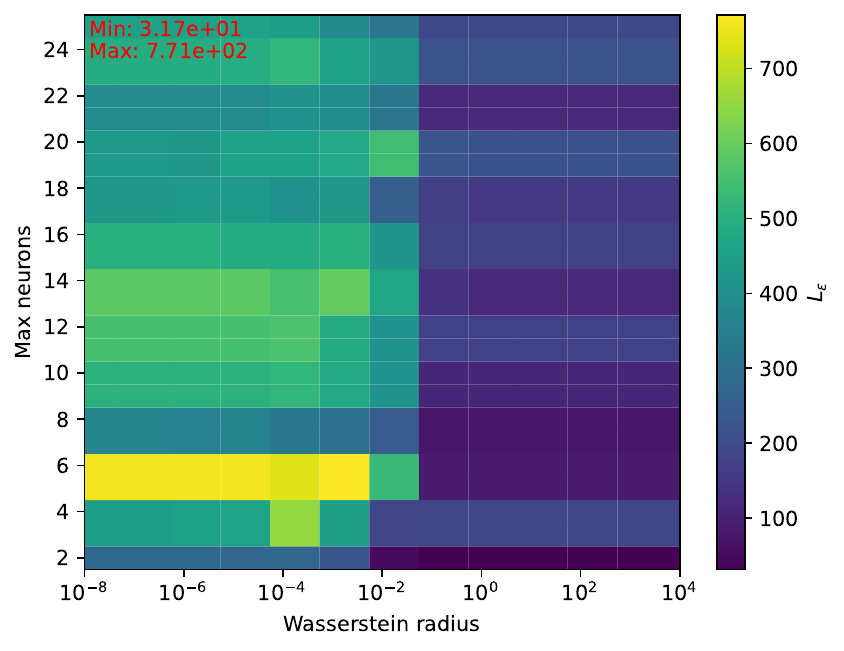}
        \caption{$L_\varepsilon$: WaDiRo-SCNN $\left(W_{\|\cdot\|_1,1}\right)$}
        %\label{fig:sub3}
    \end{subfigure}
        \hfill 
    \begin{subfigure}{0.48\textwidth}
        \includegraphics[width=\textwidth]{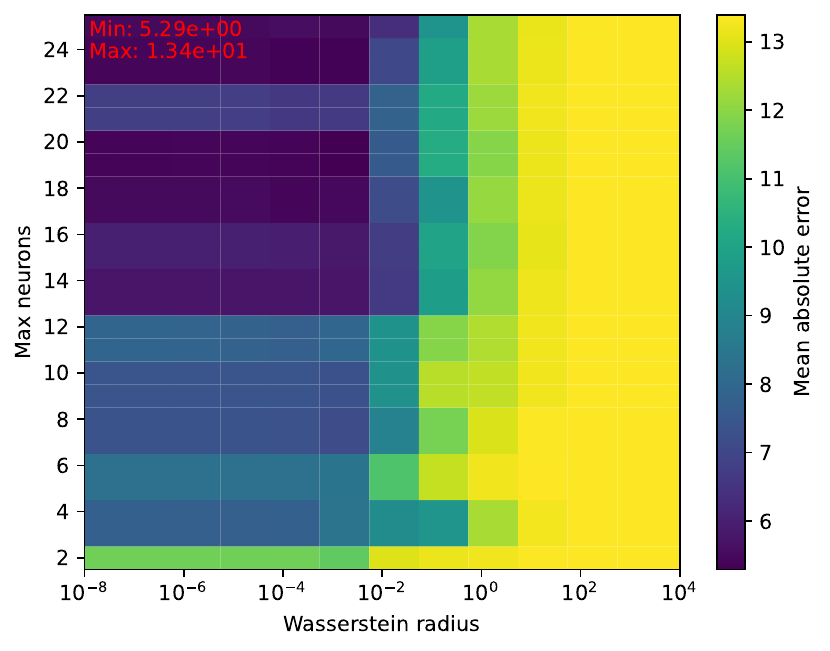}
        \caption{MAE test: WaDiRo-SCNN $\left(W_{\|\cdot\|_2,1}\right)$}
        %\label{fig:sub3}
    \end{subfigure}
           \hfill 
    \begin{subfigure}{0.48\textwidth}
        \includegraphics[width=\textwidth]{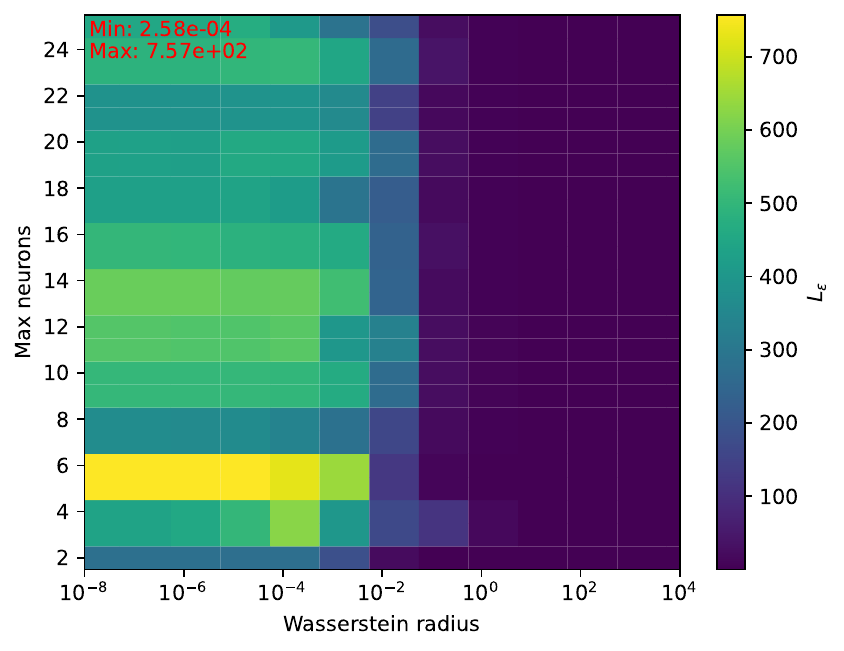}
        \caption{$L_\varepsilon$: WaDiRo-SCNN $\left(W_{\|\cdot\|_2,1}\right)$}
        %\label{fig:sub3}
    \end{subfigure}
    
    \caption{Experiments on \texttt{concrete}}
    \label{fig:concrete}
\end{figure}

\begin{figure}[tb]
    \centering
    \begin{subfigure}{0.48\textwidth}
        \includegraphics[width=\textwidth]{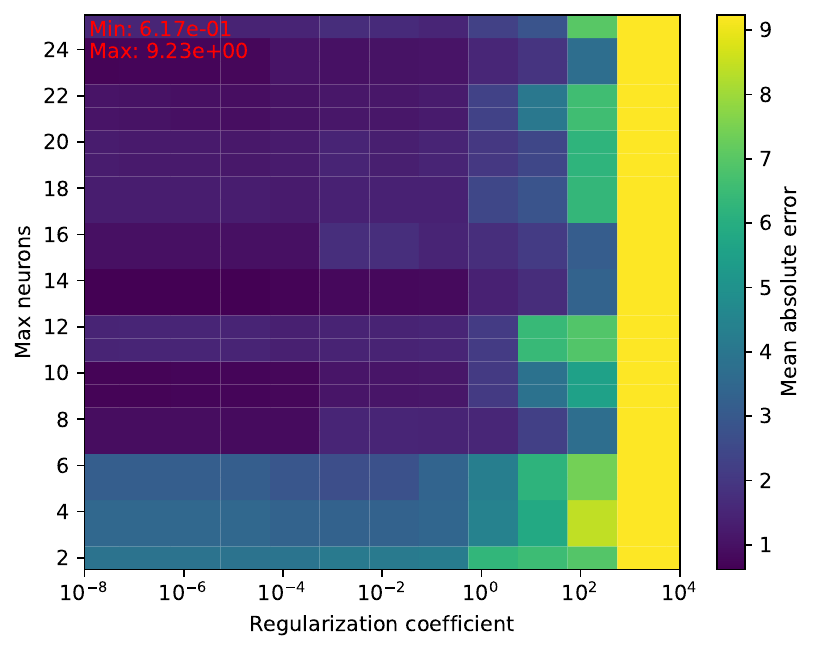}
        \caption{MAE test: LASSO-SCNN}
        %\label{fig:sub1}
    \end{subfigure}
    \hfill
    \begin{subfigure}{0.48\textwidth}
        \includegraphics[width=\textwidth]{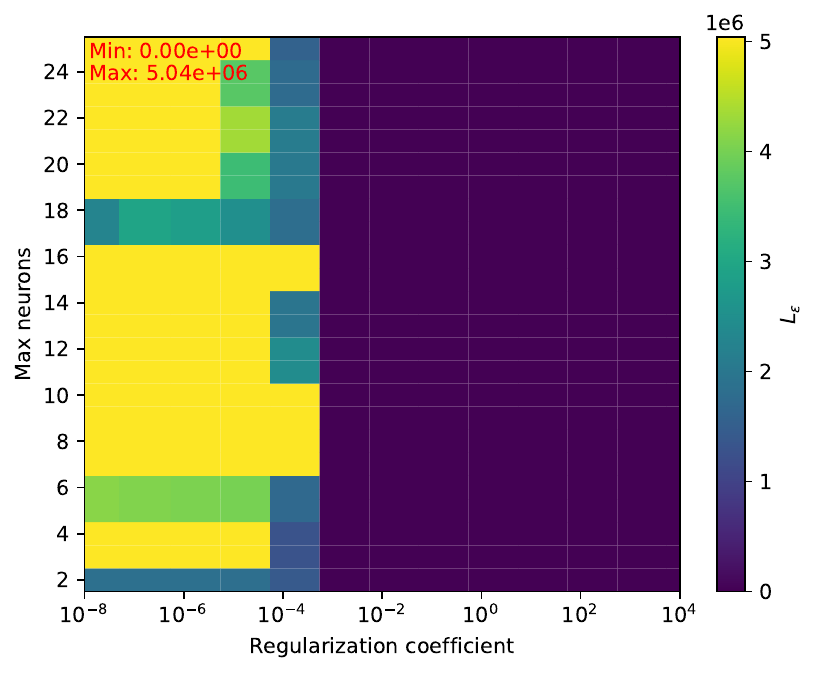}
        \caption{$L_\varepsilon$: LASSO-SCNN}
        %\label{fig:sub2}
    \end{subfigure}
    \hfill
    \begin{subfigure}{0.48\textwidth}
        \includegraphics[width=\textwidth]{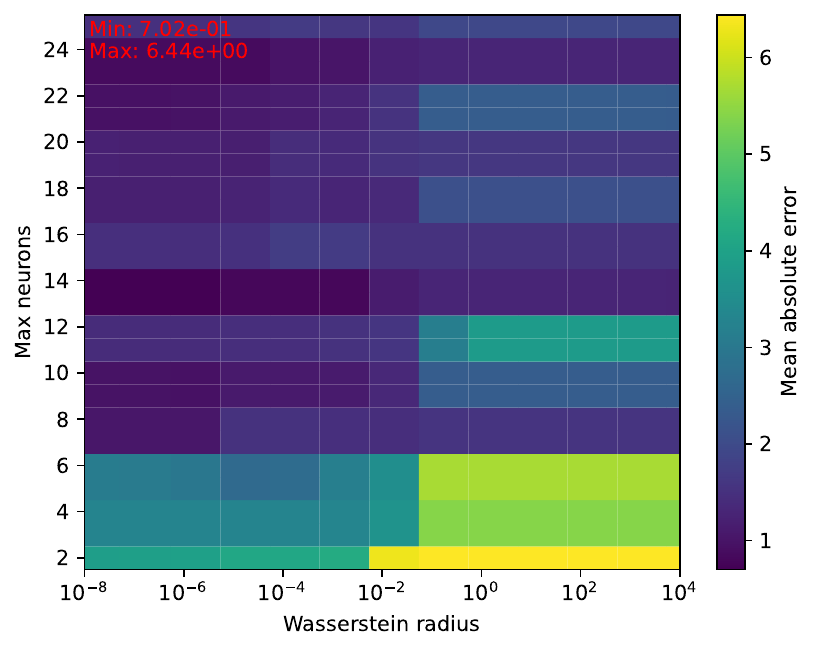}
        \caption{MAE test: WaDiRo-SCNN $W_{\|\cdot\|_1,1}$}
        %\label{fig:sub3}
    \end{subfigure}
    \hfill 
    \begin{subfigure}{0.48\textwidth}
        \includegraphics[width=\textwidth]{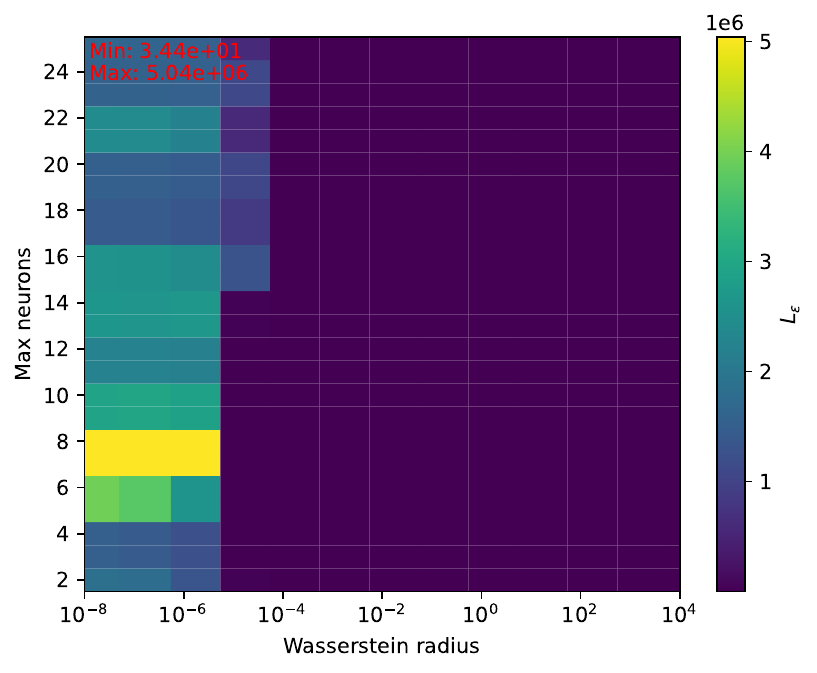}
        \caption{$L_\varepsilon$: WaDiRo-SCNN $W_{\|\cdot\|_1,1}$}
        %\label{fig:sub3}
    \end{subfigure}
        \hfill 
    \begin{subfigure}{0.48\textwidth}
        \includegraphics[width=\textwidth]{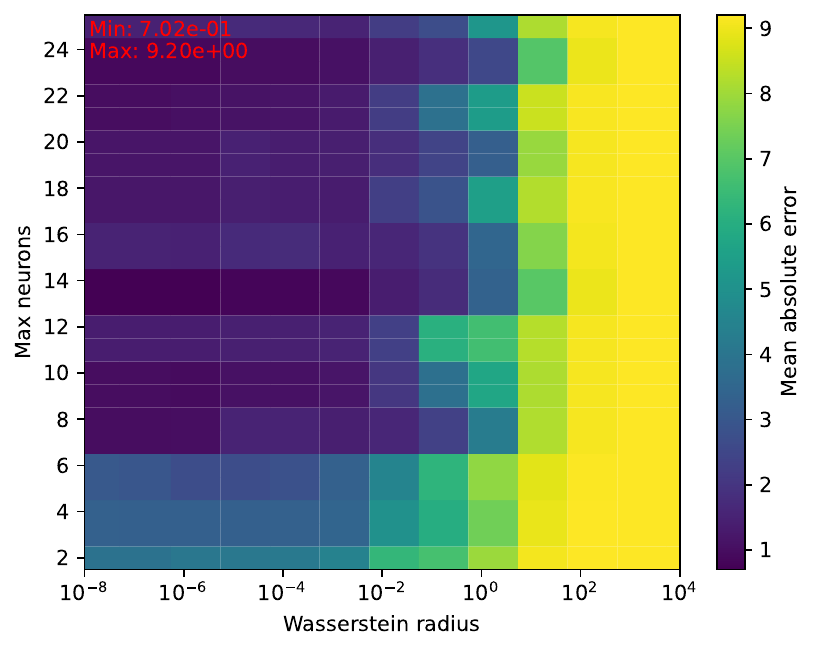}
        \caption{MAE test: WaDiRo-SCNN $W_{\|\cdot\|_2,1}$}
        %\label{fig:sub3}
    \end{subfigure}
           \hfill 
    \begin{subfigure}{0.48\textwidth}
        \includegraphics[width=\textwidth]{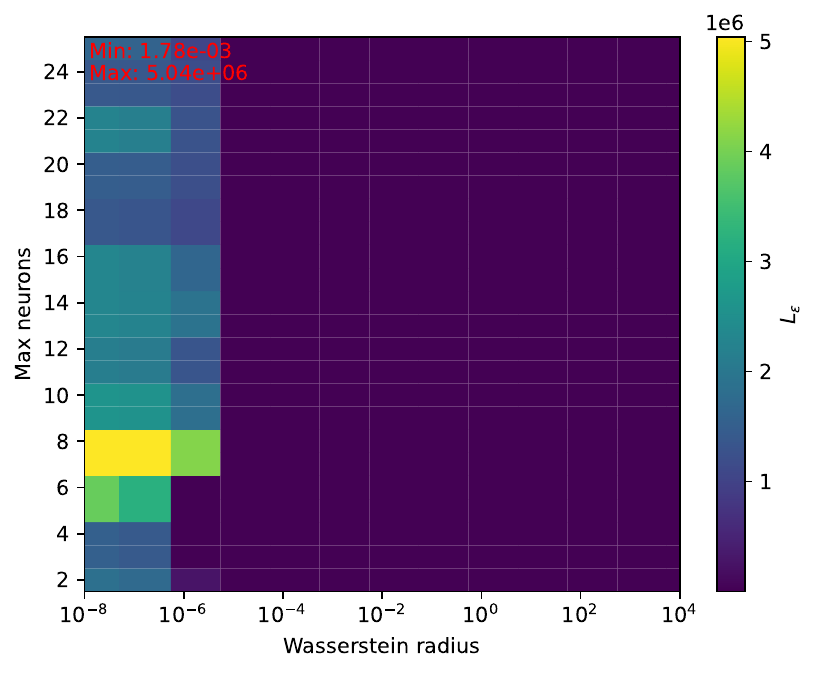}
        \caption{$L_\varepsilon$: WaDiRo-SCNN $W_{\|\cdot\|_2,1}$}
        %\label{fig:sub3}
    \end{subfigure}
    
    \caption{Experiments on \texttt{energy}}
    \label{fig:energy}
\end{figure}

Experiment results are presented in Figures \ref{fig:concrete} and \ref{fig:energy}.  We observe some general tendencies across the two datasets and the different models. The minimum and maximum of each scale are identified in the top left corner of each subfigure. First, the lowest test errors are achieved for high neuron counts and low regularization. Instability seems inversely correlated with the size of the regularization parameter. Thus, there seems to be a trade-off between precision and stability. We note that some neurons count are \textit{anomalously} less stable than others, e.g., 6 neurons in Figure \ref{fig:concrete}\Red{. W}e believe that this is related to the population of sampling vectors used \Red{because} the same outlying patterns emerge similarly for all models. 

The experiment on the \texttt{energy} dataset demonstrates that neural networks can be highly unstable with lower regularization. Indeed, with the added effect of the scaling, $L_\varepsilon$ reaches the millions. For a \Red{large} number of neurons and an intermediate regularization parameter, we note a \Red{favourable balance between low test errors and model stability.}

Overall, this experiment validates experimentally that our WaDiRo training methodology operates as a regularizer.

\color{black}
\section{Closing remarks}\label{sec:conclusion}
% what we have done, highlight some limitations, next steps
In this work, we reformulate the $\ReLU$-SCNN training problem into its tractable convex order-1 Wasserstein distributionally robust counterpart by using the $\ell_1$-loss function and by conditioning on the set of sampling vectors required by the SCNN. We demonstrate that the training problem has inherent out-of-sample performance guarantees. We show that our settings allow the inclusion of convex physical constraints, which are then guaranteed to be respected during training\blue{, and easy integration in mixed-integer convex post-training verification programs. These results showcase the strength and potential of convex learning for critical applications.} We showcase the conservatism of our model in \blue{three} experiments: a synthetic experiment with controlled data corruption, the real-world prediction of non-residential buildings' hourly energy consumption\blue{, and two datasets from the UCI ML repository on which we certify model stability}.

This work does have some limitations that must be highlighted. The WaDiRo-SCNN training problem is only defined for one output neuron, it is not yet GPU parallelizable, and, as most convex optimization problems, it takes an increasing amount of time to solve when augmenting its dimensionality, e.g., the maximal numbers of neurons in the hidden layers, the size of the training set, and the number of features. As such, adapting its formulation for GPU acceleration or adapting the algorithm to a more efficient solving method, similarly to work done by~\cite{mishkin2022scnn}, \cite{bai2023efficient} \blue{and more recently by \cite{feng2024cronos}}, would be highly interesting.  We remark that the integration of physical constraints has not been covered much as it will be investigated further in future works \blue{on nonlinear optimal control}.

% Acknowledgments here
\paragraph{Acknowledgements}{This work was possible thanks to funding from the Fonds de recherche du Québec (FRQ), the Natural Sciences and Engineering Research Council of Canada (NSERC), Mitacs Accelerate, and Hilo by Hydro-Québec under scholarships CGRS-M and B1X, as well as Grants RGPIN-2023-04235 and IT35303 \& IT38517. Special thanks to Salma Naccache and Bertrand Scherrer for their active support and the enriching discussions, as well as Steve Boursiquot, Ahmed Abdellatif, and Odile Noël from Hilo for making this project possible.}

% References here (outcomment the appropriate case)

% CASE 1: BiBTeX used to constantly update the references
%   (while the paper is being written).
\bibliographystyle{informs2014} % outcomment this and next line in Case 1
\bibliography{ref.bib} % if more than one, comma separated

%\THEEndNotes
%\begingroup \parindent 0pt \parskip 0.0ex \def\enotesize{\normalsize} \theendnotes \endgroup

% Appendix here
% Options are (1) APPENDIX (with or without general title) or
%             (2) APPENDICES (if it has more than one unrelated sections)
% Outcomment the appropriate case if necessary
%
% \begin{APPENDIX}{<Title of the Appendix>}
% \end{APPENDIX}
%
%   or
%
\appendix
\section*{Appendices}
\addcontentsline{toc}{section}{Appendices}
\renewcommand{\thesubsection}{\Alph{subsection}}
% \section{<Title of Section A>}
% \section{<Title of Section B>}
% etc

\normalsize
\subsection{Proof of Proposition \ref{pr:l1}}\label{sec:AnnA}

Our objective is to rewrite~\eqref{eq:W-DRO SCNN} in the form of~\eqref{eq:CWDRO}. This proof is derived from results of~\cite{Chen2020_DRO} and is adapted here for completeness. Consider the $\ell_1$-norm loss function. We have $h_{\boldsymbol{\beta}}(\hat{\mathbf{z}}) = |\hat{\mathbf{z}}^\top \boldsymbol{\beta}|$ and
$$h_{\boldsymbol{\beta}}^*(\boldsymbol{\theta}) < +\infty \iff \sup_{\hat{\mathbf{z}}: \hat{\mathbf{z}}^\top \boldsymbol{\beta} \geq 0} \{\boldsymbol{\theta}^\top \hat{\mathbf{z}} - {\boldsymbol{\beta}}^\top \hat{\mathbf{z}} \} < +\infty \ \ \text{and} \  \sup_{\hat{\mathbf{z}}: \hat{\mathbf{z}}^\top \boldsymbol{\beta} \leq 0} \{\boldsymbol{\theta}^\top \hat{\mathbf{z}} + {\boldsymbol{\beta}}^\top \hat{\mathbf{z}} \} < +\infty. $$
Considering the two linear optimization problems, we can find the equivalent dual problems A and~B with dual variables $\lambda_A$ and $\lambda_B$, respectively:
\begin{alignat}{3}
   &\max \ (\boldsymbol{\theta} - {\boldsymbol{\beta}})^\top \hat{\mathbf{z}} && \xrightarrow{\text{dual}} \ && \min \ 0 \cdot \lambda_A \tag{\texttt{Dual-A}}\label{eq:dual-a}\\
   &\text{s.t.} \ \hat{\mathbf{z}}^\top \boldsymbol{\beta} \geq 0 && \ && \text{s.t.} \ {\boldsymbol{\beta}} \lambda_A = \boldsymbol{\theta} - {\boldsymbol{\beta}},\ \lambda_A \leq 0, \notag
  \intertext{and,}
    &\max (\boldsymbol{\theta} + {\boldsymbol{\beta}})^\top \hat{\mathbf{z}} && \xrightarrow{\text{dual}} \ && \min \ 0 \cdot \lambda_B \tag{\texttt{Dual-B}}\label{eq:dual-b}\\
   &\text{s.t. }\ \hat{\mathbf{z}}^\top \boldsymbol{\beta} \leq 0 && \ && \text{s.t.} \ {\boldsymbol{\beta}} \lambda_B = \boldsymbol{\theta} + {\boldsymbol{\beta}},\ \lambda_B \geq 0. \notag
\end{alignat}
Because the objective function of both dual problems is 0, to have finite optimal values for both primals,~\eqref{eq:dual-a} and~\eqref{eq:dual-b} need to have a non-empty feasible set.
This implies that these two constraints must be respected:
\begin{alignat*}{2}
   & \exists \lambda_A \leq 0, \quad \text{s.t.} \quad &&{\boldsymbol{\beta}} \lambda_A = \boldsymbol{\theta} - {\boldsymbol{\beta}}\\
   & \exists \lambda_B \geq 0, \quad \text{s.t.} \quad &&{\boldsymbol{\beta}} \lambda_B = \boldsymbol{\theta} + {\boldsymbol{\beta}}.
\end{alignat*}
From this we find that,  \begin{equation}|\theta_i|\leq |{\beta}_i| \ \forall i, \label{eq:theta_beta}\end{equation} because
\begin{alignat*}{1}
{\boldsymbol{\beta}}(1 - \lambda_B) &= \boldsymbol{\theta} \ \text{where} \ \lambda_B \geq 0\\
{\boldsymbol{\beta}}(1 + \lambda_A) &= \boldsymbol{\theta} \ \text{where} \ \lambda_A \leq 0.
\end{alignat*}
If~\eqref{eq:theta_beta} holds, then both primals are finite. Thus, we have
$$\kappa({\boldsymbol{\beta}}) = \sup\{ \|{\boldsymbol{\theta}}\|_* : |\theta_i|\leq {\beta_i}, \forall i\},$$
and, 
$$\kappa({\boldsymbol{\beta}}) = \| {\boldsymbol{\beta}}\|_* = \sup_{\|\mathbf{k}\|\leq 1} {\boldsymbol{\beta}}^\top \mathbf{k}. $$
For the $\ell_1$-loss, $\| {\boldsymbol{\beta}}\|_* =\|{\boldsymbol{\beta}} \|_{\infty} $, and the problem becomes:
\begin{alignat}{2}
    &&\inf_{\boldsymbol{\beta}}  \epsilon \|{\boldsymbol{\beta}} \|_{\infty} + \frac{1}{N} \sum_{i=1}^{N} h_{\boldsymbol{\beta}}(\hat{\mathbf{z}}_i). \label{eq:l1dro}
\end{alignat}
Using the slack variables $a \in \mathbb{R}$ and $\mathbf{c}\in \mathbb{R}^N$ in, respectively, the first and second terms of the objective, yields the equivalent problem:
\begin{alignat*}{3}
   &\min_{\boldsymbol{\nu}, \boldsymbol{\omega}} \quad &&\epsilon  a  \ &&+ \frac{1}{N} \sum_{j=1}^{N} c_j \\
   &\text{s.t.}  &&{\boldsymbol{\beta}}_i &&\leq a \quad \forall i \in \llbracket 2P d + 1 \rrbracket\\
    & &&-{{\beta}}_i &&\leq a \quad \forall i \in \llbracket 2P  d + 1 \rrbracket\\
   & &&{{\beta}}^\top \hat{\mathbf{z}}_j &&\leq c_j \quad \forall j \in \llbracket N\rrbracket\\
   & &&-({\boldsymbol{\beta}}^\top \hat{\mathbf{z}}_j) &&\leq c_j \quad \forall j \in \llbracket N\rrbracket\\
   & && \blue{\boldsymbol{\beta}} &&\blue{= \left( \vc(\boldsymbol{\nu}),\vc(-\boldsymbol{\omega}),  -1 \right) }\\
    & &&\boldsymbol{\nu}_i, \boldsymbol{\omega}_i &&\in \mathcal{K}_i \quad  \forall i \in \llbracket P\rrbracket.
\end{alignat*}
For the $\ell_2$-loss, $\| {\boldsymbol{\beta}}\|_* $ is equal to the $\ell_2$-norm~\citep{boyd2004convex}. By the same process and noting that $a\geq 0$ because a norm is non-negative, we obtain the second equivalent problem: 
\begin{alignat*}{3}
   &\min_{\boldsymbol{\nu}, \boldsymbol{\omega}} \quad && \epsilon  a && + \frac{1}{N} \sum_{j=1}^{N} c_j \\
   &\text{s.t.} &&\|{\boldsymbol{\beta}}\|_2^2  &&\leq a^2 \quad \quad \\
   & \ &&a &&\geq 0  \\
   & \ &&{\boldsymbol{\beta}}^\top \hat{\mathbf{z}}_j &&\leq c_j \quad \forall j \in \llbracket N\rrbracket\\
   & \ &&-({\boldsymbol{\beta}}^\top \hat{\mathbf{z}}_j) &&\leq c_j \quad \forall j \in \llbracket N\rrbracket\\
   & && \blue{\boldsymbol{\beta}} &&\blue{= \left( \vc(\boldsymbol{\nu}),\vc(-\boldsymbol{\omega}),  -1 \right) }\\
    &\ &&\boldsymbol{\nu}_i, \boldsymbol{\omega}_i &&\in \mathcal{K}_i \quad  \forall i \in \llbracket P\rrbracket,
\end{alignat*} 
which completes the proof. \hfill $\square$

\subsection{Complementary materials for the synthetic experiment}\label{sec:AnnB}
\normalsize
In this section, we detail the synthetic experiments of Section \ref{subsec:synth} and we present additional figures. 

\subsubsection{Hyperparameters and setting}
The hyperparameters of each model are presented in Table \ref{tab:hypersynth}. Note that all models are formulated with bias weights by default and that different regularization methodologies are set as hyperparameters in this experiment.

\begin{table}[tb]
\renewcommand{\arraystretch}{1.0}
    \centering
    \caption{Hyperparameters of the synthetic experiment}
    \begin{tblr}{colspec={m{3cm}|lll},hline{1,8} = {1.5pt}, hline{2,3,4,5,6,7}={1pt}} 
    \textbf{Model} & \textbf{Hyperparameter} & \textbf{Description} & \textbf{Possible values} \\
    WaDiRo-SCNN & {\texttt{radius} \\ \texttt{max\_neurons} \\ \texttt{wasserstein} }  & { Radius of the Wasserstein ball \\ Max number of neurons \\ Norm \blue{of the} Wasserstein metric} & { $[\mathrm{e}^{-8}, \mathrm{e}^1]$ \\ $\{10, 11, \ldots, 300\}$ \\ $\ell_1$ or $\ell_2$ }\\
    Regularized SCNN & {\texttt{lambda\_reg} \\ \texttt{max\_neurons} \\ \texttt{regularizer} }  & {Regularizing parameter \\ Max number of neurons \\ Regularization framework } & { $[\mathrm{e}^{-8}, \mathrm{e}^1]$ \\$ \{10, 11, \ldots, 300\}$ \\  Lasso or ridge }\\
    Non-regularized SCNN & {\texttt{max\_neurons}}  & { Max number of neurons } & { $ \{10, 11, \ldots, 300\}$ }\\
    WaDiRo lin. reg. & {\texttt{radius} \\ \texttt{wasserstein} }  & { Radius of the Wasserstein ball \\ Norm \blue{of the} Wasserstein metric } & { $[\mathrm{e}^{-8}, \mathrm{e}^1]$ \\ $\ell_1$ or $\ell_2$ }\\
    Regularized lin. reg. & {\texttt{lambda\_reg} \\ \texttt{regularizer} }  & {Regularizing parameter  \\ Regularization framework } & { $[\mathrm{e}^{-8}, \mathrm{e}^1]$ \\ Lasso or ridge }\\
    Deep FNN & {\texttt{batch\_size} \\ \texttt{n\_hidden} \\ \texttt{learning\_rate} \\ \texttt{n\_epochs}\\ \texttt{dropout\_p}}  & { Size of each batch \\ Number of neurons per layer \\ Learning rate \\ Number of epochs \\ Dropout probability at each layer } & { $\{2, 3, \ldots, 100\}$ \\$ \{10, 11, \ldots , 300\}$ \\ $[\mathrm{e}^{-8}, \mathrm{e}^1]$ \\ $ \{2, 3, \ldots, 10^3\}$ \\ $[0.01, 0.4]$} \\
\end{tblr}
    \label{tab:hypersynth}
\end{table}
We sample 2000 data points for each benchmark function before splitting them between training, validation, and \blue{testing}. For the benchmark functions that can be defined in dimensions higher than two, Ackley and Keane, we impose a dimension of four.

\subsubsection{Additionnal figures}
\begin{figure}[tb]
    \centering
    \begin{subfigure}{0.6\textwidth}
        \includegraphics[width=\textwidth]{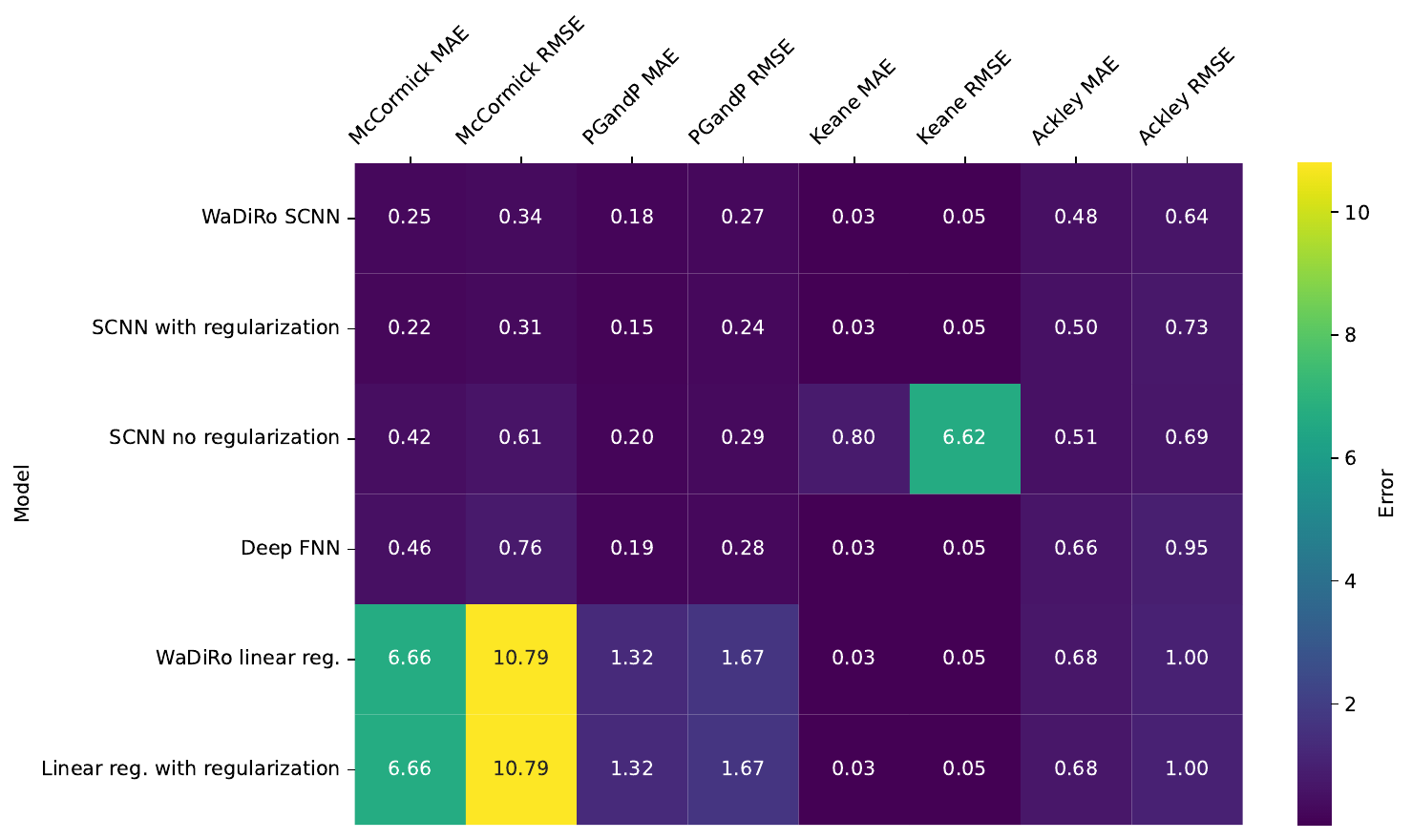}
        \caption{Experiment A}
        
    \end{subfigure}
    \hfill
    \begin{subfigure}{0.6\textwidth}
        \includegraphics[width=\textwidth]{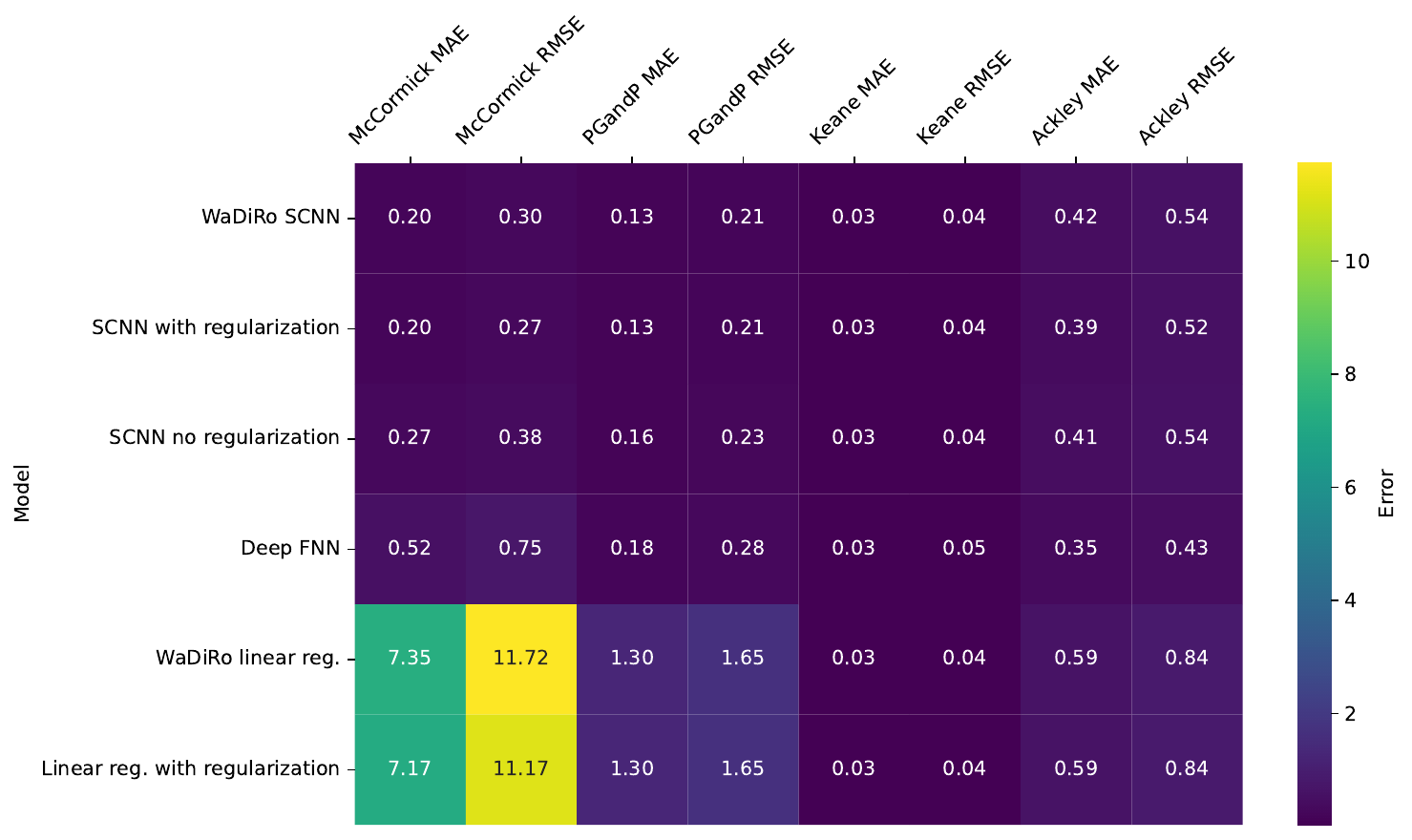}
        \caption{Experiment B}
        %\label{fig:sub2}
    \end{subfigure}
    \hfill
    \begin{subfigure}{0.6\textwidth}
        \includegraphics[width=\textwidth]{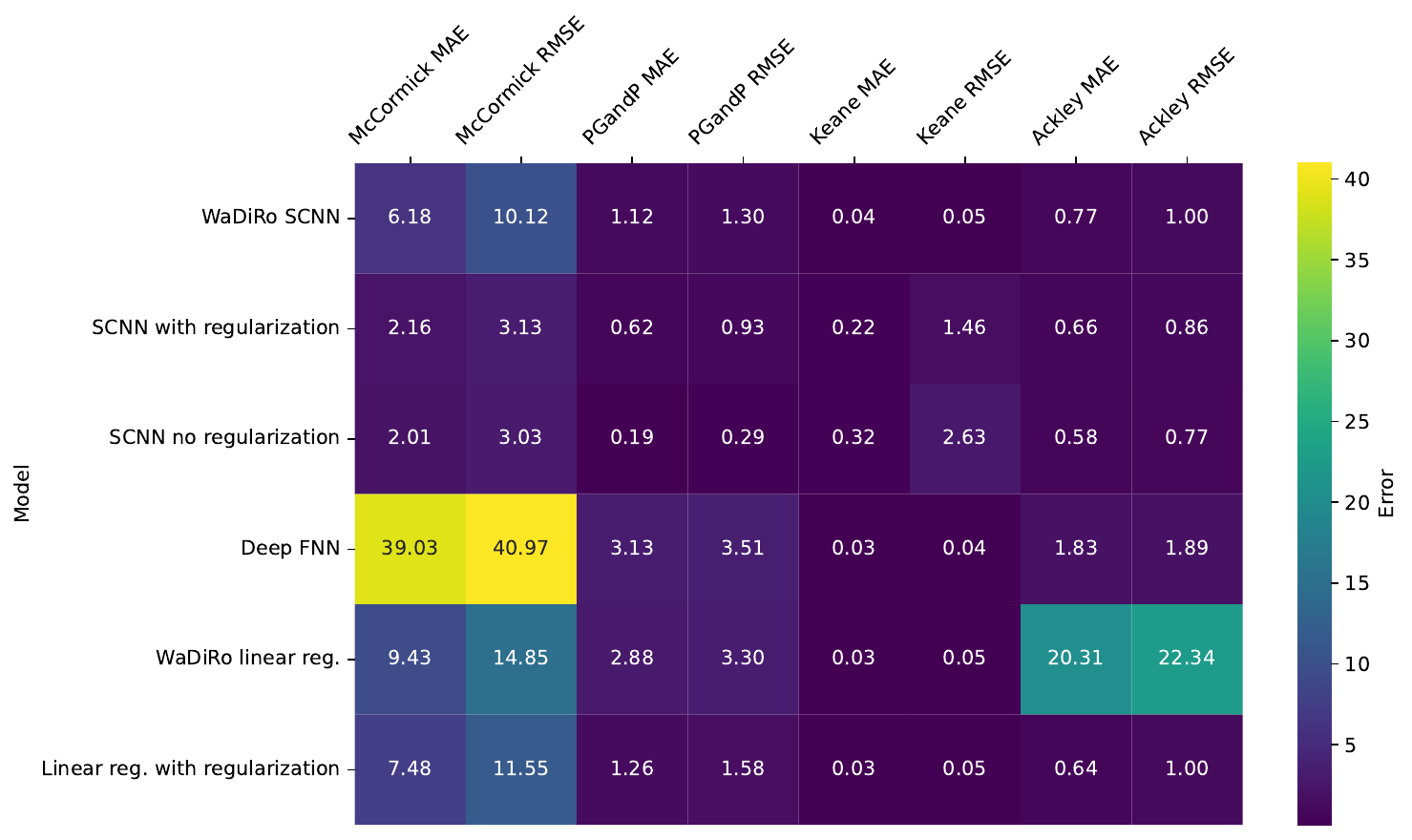}
        \caption{Experiment C}
        %\label{fig:sub3}
    \end{subfigure}
    \caption{Absolute errors of each synthetic experiment}
    \label{fig:expsynth}
\end{figure}

Figure \ref{fig:expsynth} presents the absolute values of the errors presented in Section \ref{subsec:synthres}. \color{black} Label values for each function on the feature domain are needed to better understand the figure. We proceed to identify the max and min sampled values. For McCormick, the minimal value is -1.91 and the maximal value is 65.32. For PGandP, the minimal value is -3.12 and the maximal value is 6.78. For Keane, the minimal value is -9.13 and the maximal value is 0. For Ackley, the minimal value is 0 and the maximal value is 22.29. 

Compared to the normalized results, McCormick is extremely polarized: some models have low nominal test errors while others have bigger ones. While the WaDiRo-SCNN performs well in Experiments A and B, it sees a medium performance in Experiment C. This is hidden by the extremely poor performance of the deep FNN in Experiment C. Figure \ref{fig:whole} shows that McCormick has linear tendencies in the centre of its domain but steep increases on its edges that linear models fail to capture. 

Keane, with its small codomain, has a small error range. We observe that the SCNN with no regularization is the only one to fail in Experiment A, while both standard SCNN formulations fail comparatively to the other models in Experiment C. 

One interesting takeaway from Figure \ref{fig:expsynth} is that WaDiRo-SCNN, at its worst performance, is on par with linear regression models. We do see that the standard SCNN formulations have a poorer performance than the linear models on Keane, and the deep FNN has consistently worse performances in the corrupted setting from Experiment C.\color{black}

\color{black}
\subsubsection{Computational time}\label{subsec:compute}
\begin{figure}[tb]
    \centering
    \begin{subfigure}{0.48\textwidth}
        \includegraphics[width=\textwidth]{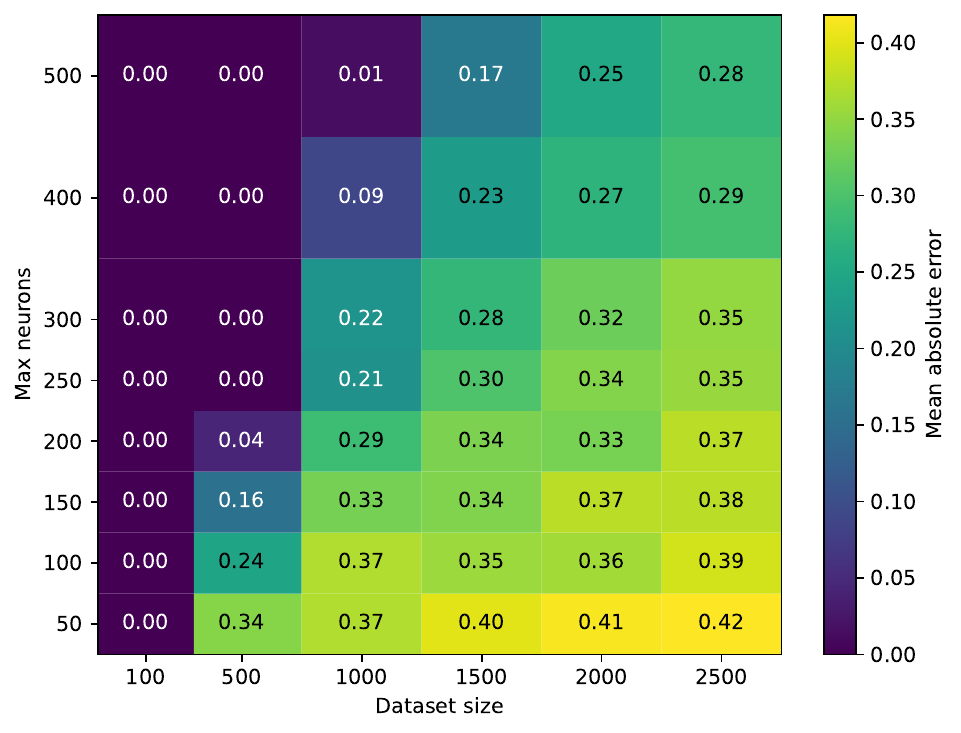}
        \caption{Training MAE}
        
    \end{subfigure}
    \hfill
    \begin{subfigure}{0.48\textwidth}
        \includegraphics[width=\textwidth]{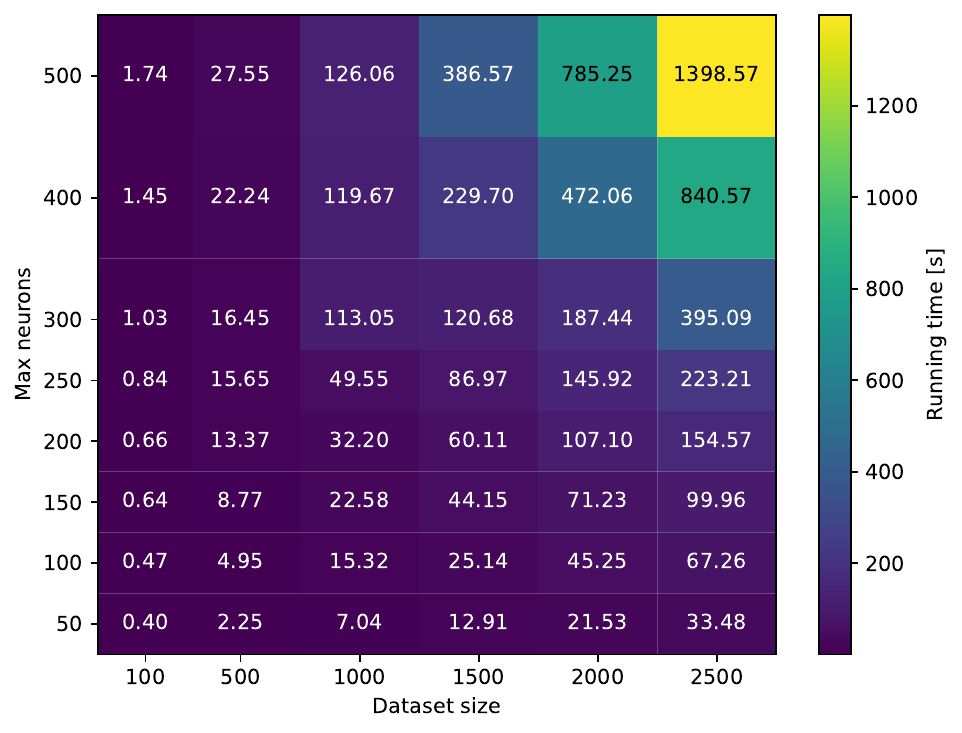}
        \caption{Training time}
        %\label{fig:sub2}
    \end{subfigure}
    \caption{Training error and time on Ackley}
    \label{fig:time}
\end{figure}

The tests presented in Figure \ref{fig:time} were conducted on a computer with 64GB of RAM DDR5 and an AMD Ryzen 9 7950x3d CPU. The training was performed with the open-source solver \texttt{Clarabel} and was formulated in Python with \texttt{cvxpy}. As we can see, more neurons lead to smaller training errors, while a greater dataset size increases the training error. Indeed, sampling more data points increases the difficulty of fitting them all correctly. 

To \Red{keep} the \Red{training} time \Red{low}, there is a \Red{trade-off} between the maximal number of neurons and the size of the dataset. This is expected partly because we do not yet have a GPU acceleration method for this training procedure. We also remark that a more direct implementation, e.g., by using a lower-level programming language \Red{such as C++} and avoiding an interface like \texttt{cvxpy} \Red{to formulate the optimization problem}, could also \Red{greatly} diminish the training time. \Red{Our implementation is thus relatively inefficient.} \Red{As such, we do not compare the training times with other models from the literature compared in Section \ref{sec:numerical}, because most of them are developed in highly computationally optimized or GPU-enabled libraries, i.e., \texttt{scikit-learn} and \texttt{pytorch}. The comparison would be unfair when dealing with large datasets without the proper computation time optimization of our method. This is left for future work as discussed in Section \ref{sec:conclusion}. Nonetheless, as can be seen in Figure \ref{fig:time}, our method can be relatively fast under some dimensionality constraints.} \color{black}
\subsection{Complementary materials for the building experiment}\label{sec:AnnC}

In this section, we detail the experiments on non-residential buildings' baseline prediction from Section \ref{subsec:real} and we showcase some additional figures of interest. 
\subsubsection{Hyperparameters and setting}
First, we define the kernel used for the Gaussian process regression. Following our inspection of the data, we construct the GP kernel by adding a radial basis function (RBF) kernel to an exponential sine kernel to model periodicity and a white noise kernel to capture the possible noise.

Similarly to the previous section, we present, in Table \ref{tab:hyperreal}, the hyperparameters used in the experiments.

\begin{table}[h]
    \centering
    \caption{Hyperparameters of the building experiment}
    \begin{tblr}{colspec={m{2.5cm}|lll}, hline{1,8} = {1.5pt}, hline{2,3,4,5,6,7}={1pt}} 
    \textbf{Model} & \textbf{Hyperparameter} & \textbf{Description} & \textbf{Possible values} \\
    WaDiRo-SCNN & {\texttt{radius} \\ \texttt{max\_neurons} \\ \texttt{wasserstein} }  & { Radius of the Wasserstein ball \\ Max number of neurons \\ Norm used in Wasserstein metric} & { $[\mathrm{e}^{-6}, \mathrm{e}^1]$ \\ $\{10, 11, \ldots, 70\}$ \\ $\ell_1$ or $\ell_2$ }\\
    WaDiRo lin. reg. & {\texttt{radius} \\ \texttt{wasserstein} }  & { Radius of the Wasserstein ball \\ Norm used in Wasserstein metric } & { $[\mathrm{e}^{-6}, \mathrm{e}^1]$ \\ $\ell_1$ or $\ell_2$ }\\
    SVR & {\texttt{C} \\ \texttt{fit\_intercept} \\ \texttt{loss}}  & {Regularizing parameter  \\ Bias \\ Loss function used} & { $[\mathrm{e}^{0.1}, \mathrm{e}^7]$ \\ True or False \\ $\ell_1$ or $\ell_2$   }\\
    GP & {\texttt{length\_scale} \\ \texttt{periodicity} \\ \texttt{lenght\_scale\_sine} \\ \texttt{noise\_level}}  & { Length-scale of the RBF kernel \\ Periodicity of the exp.-sine kernel \\ Lenght-scale of the exp.-sine kernel \\Noise level of the white noise kernel} & { $[\mathrm{e}^{0.1}, \mathrm{e}^5]$ \\ $[\mathrm{e}^{0.1}, \mathrm{e}^5]$ \\ $[\mathrm{e}^{0.1}, \mathrm{e}^5]$ \\ $[\mathrm{e}^{0.001}, \mathrm{e}^2]$ } \\
\end{tblr}
    \label{tab:hyperreal}
\end{table}

\subsubsection{Additional figures}
\begin{figure}[h]
    \centering
    \begin{subfigure}{0.48\textwidth}
        \includegraphics[height=10em]{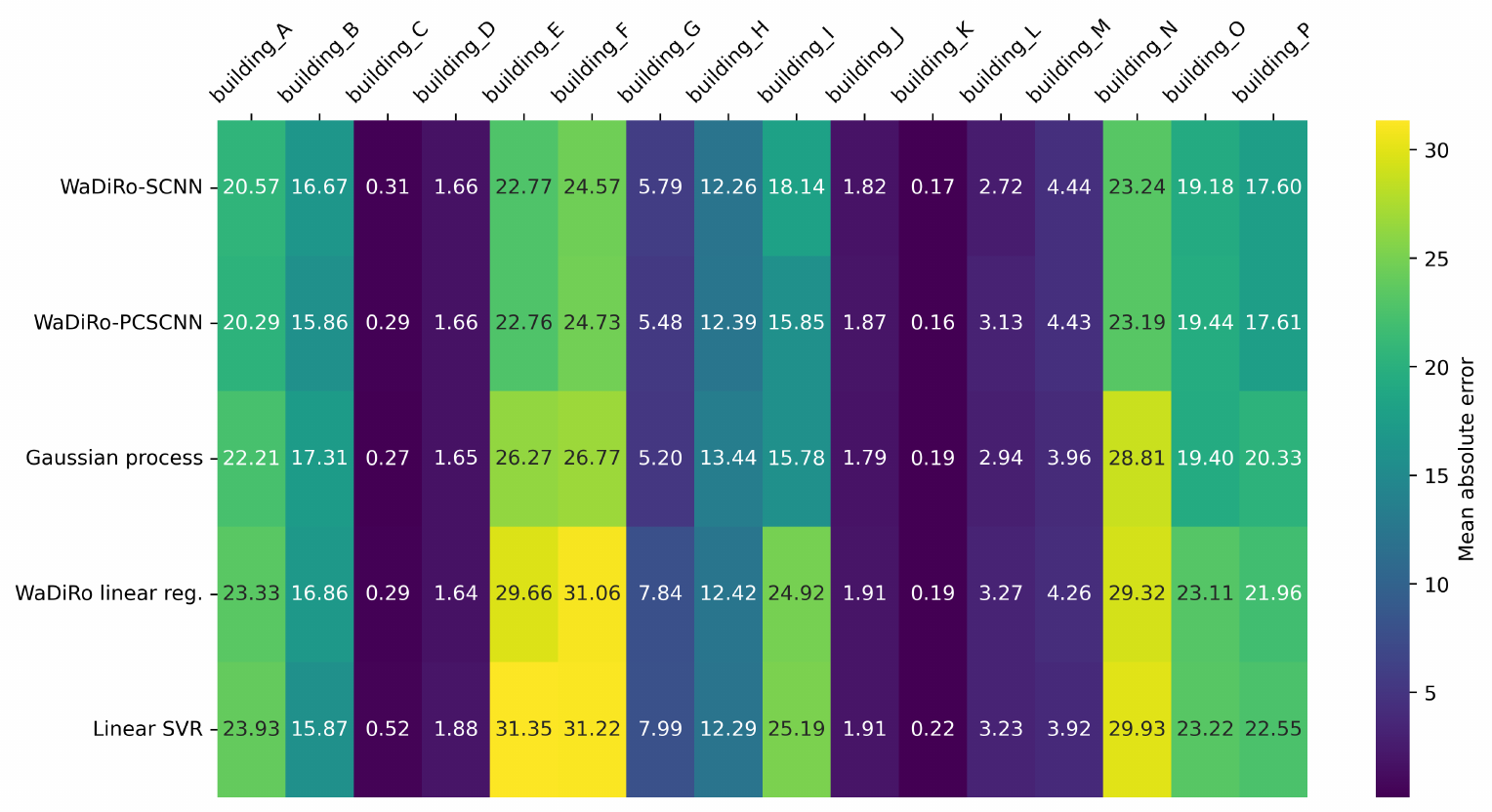}
        \caption{MAE}
        %\label{fig:sub1}
    \end{subfigure}
    \hfill
    \begin{subfigure}{0.48\textwidth}
        \includegraphics[height=10em]{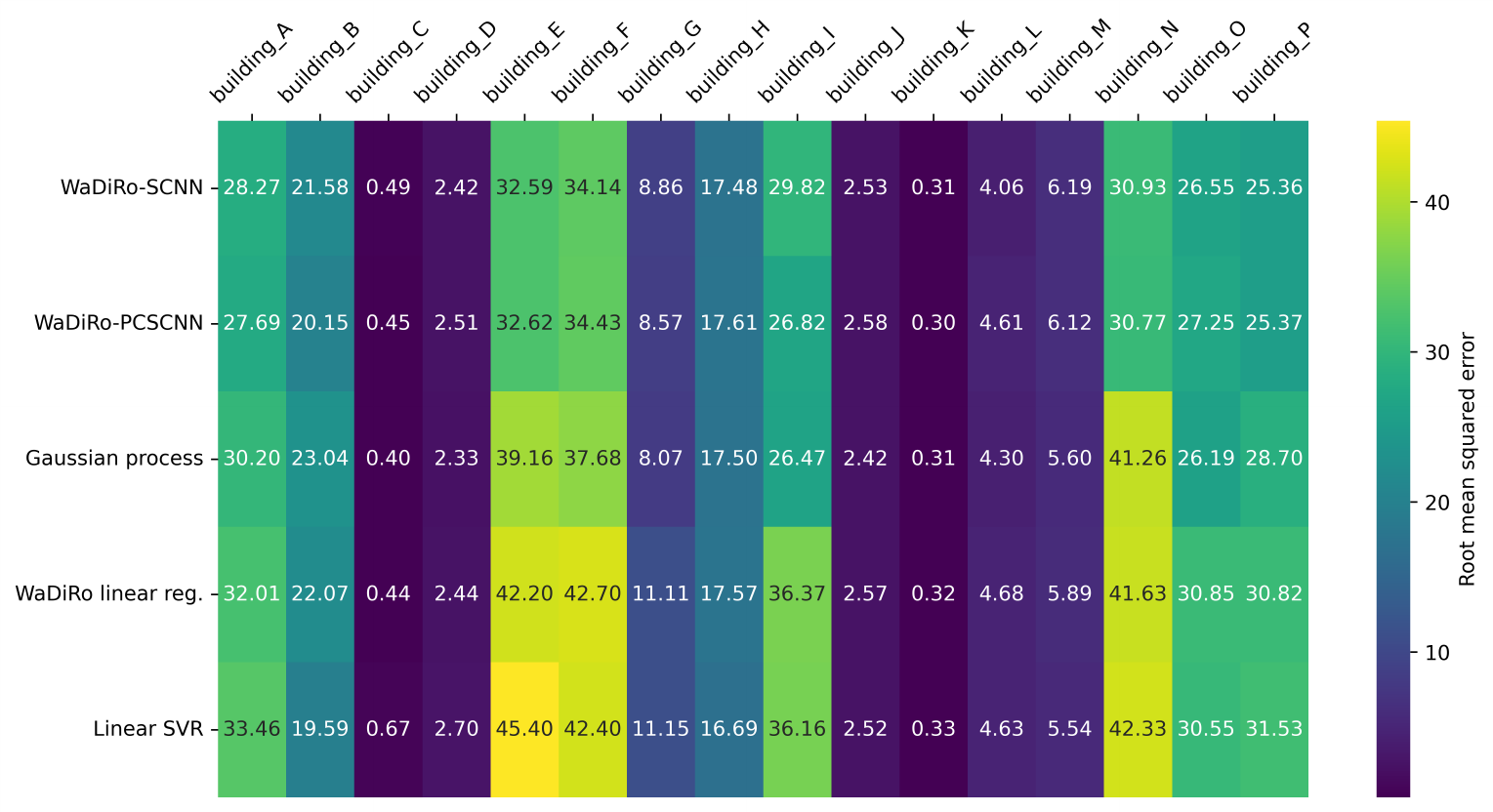}
        \caption{RMSE}
        %\label{fig:sub2}
    \end{subfigure}
    \caption{Absolute errors of the hourly baseline prediction of non-residential buildings}
    \label{fig:expbuild}
\end{figure}

In Figure \ref{fig:expbuild}, we show the absolute values of the errors corresponding to Figure \ref{fig:expbuildNorm}. \color{black} We observe that errors have the same magnitude for each building. This is to be expected as each building has its own consumption range and its own level of pattern complexity. Because we use a high number of features and test for the winter season only, linear regressions, while having worse performances than nonlinear models in most cases, are tested in a setting close to their training set. \color{black}

\begin{figure}[h]
    \centering
    \begin{subfigure}{\textwidth}
     \centering
        \includegraphics[width=0.65\textwidth]{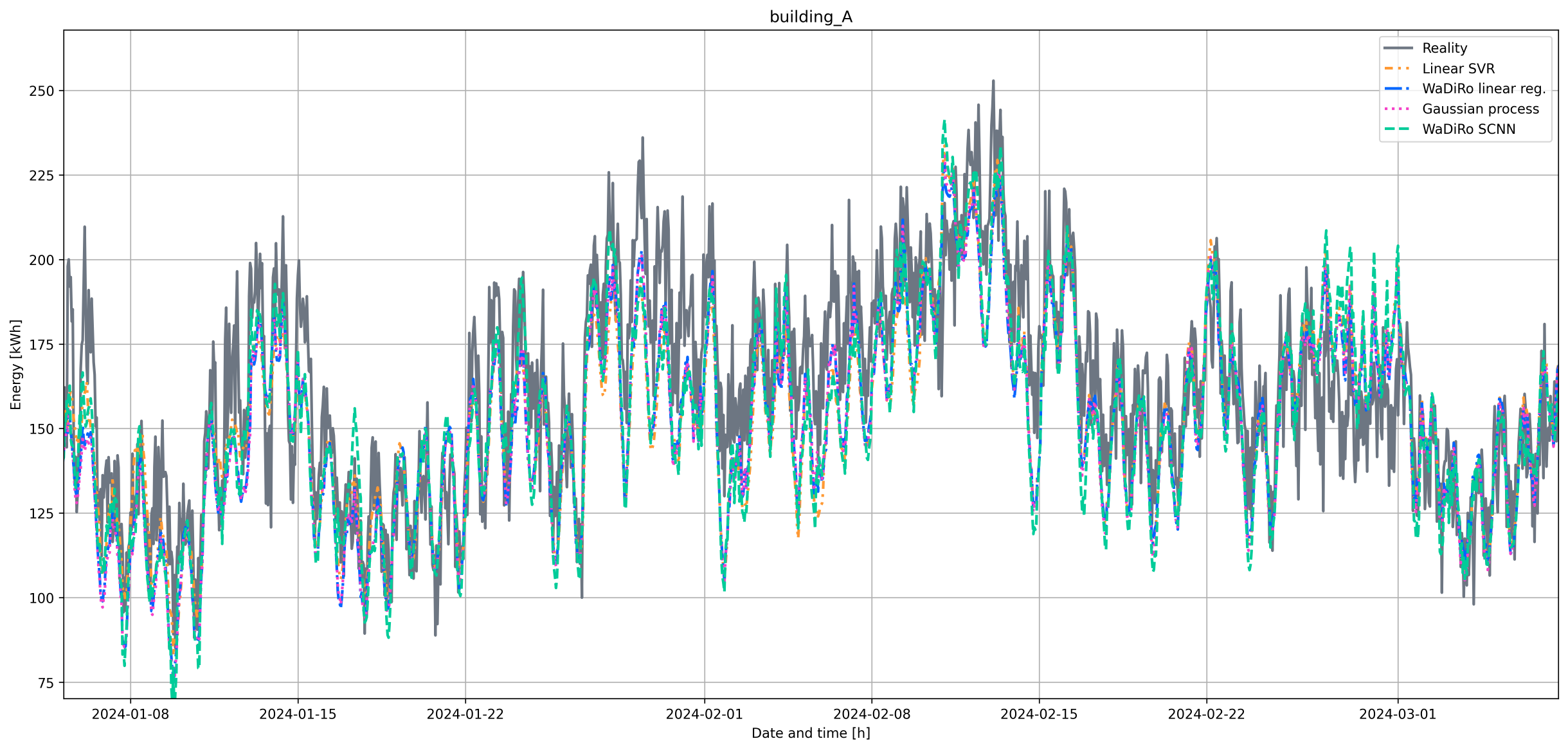}
        %\caption{Experiment A}
        %\label{fig:sub1}
    \end{subfigure}
    \hfill
    \begin{subfigure}{\textwidth}
     \centering
        \includegraphics[width=0.65\textwidth]{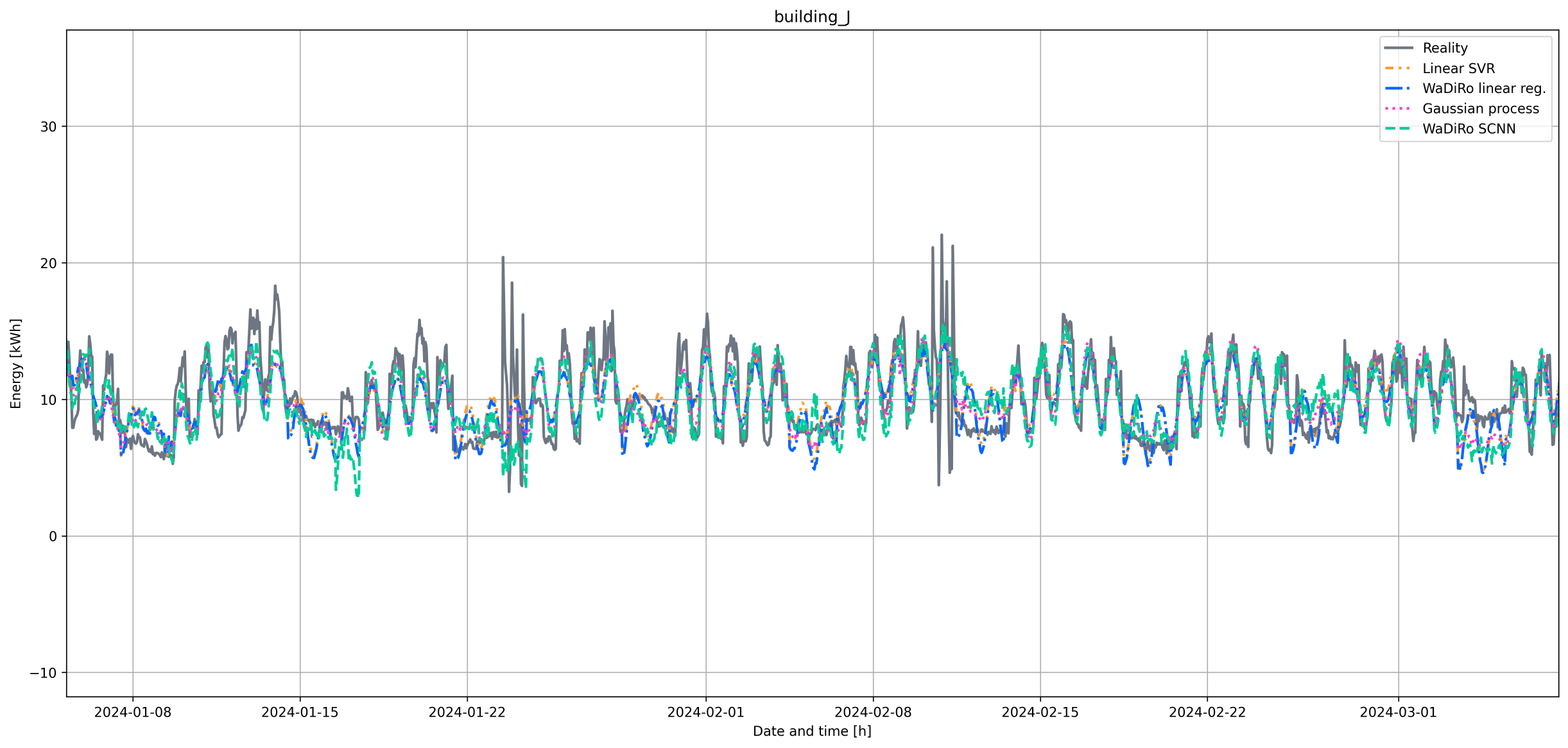}
        %\caption{Experiment B}
        %\label{fig:sub2}
    \end{subfigure}
    \hfill
    \begin{subfigure}{\textwidth}
     \centering
        \includegraphics[width=0.65\textwidth]{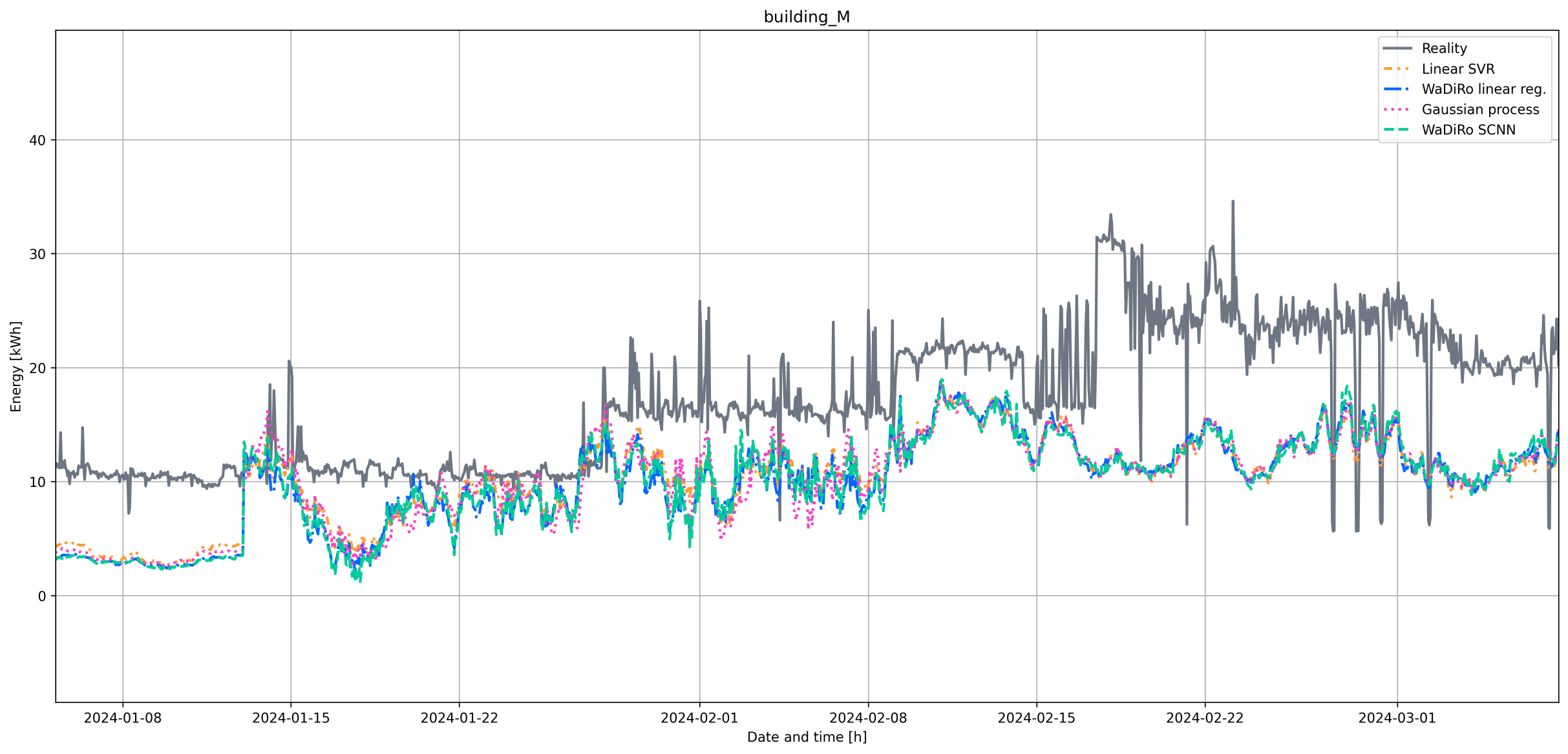}
        %\caption{Experiment C}
        %\label{fig:sub3}
    \end{subfigure}
    \hfill
    \begin{subfigure}{\textwidth}
     \centering
        \includegraphics[width=0.65\textwidth]{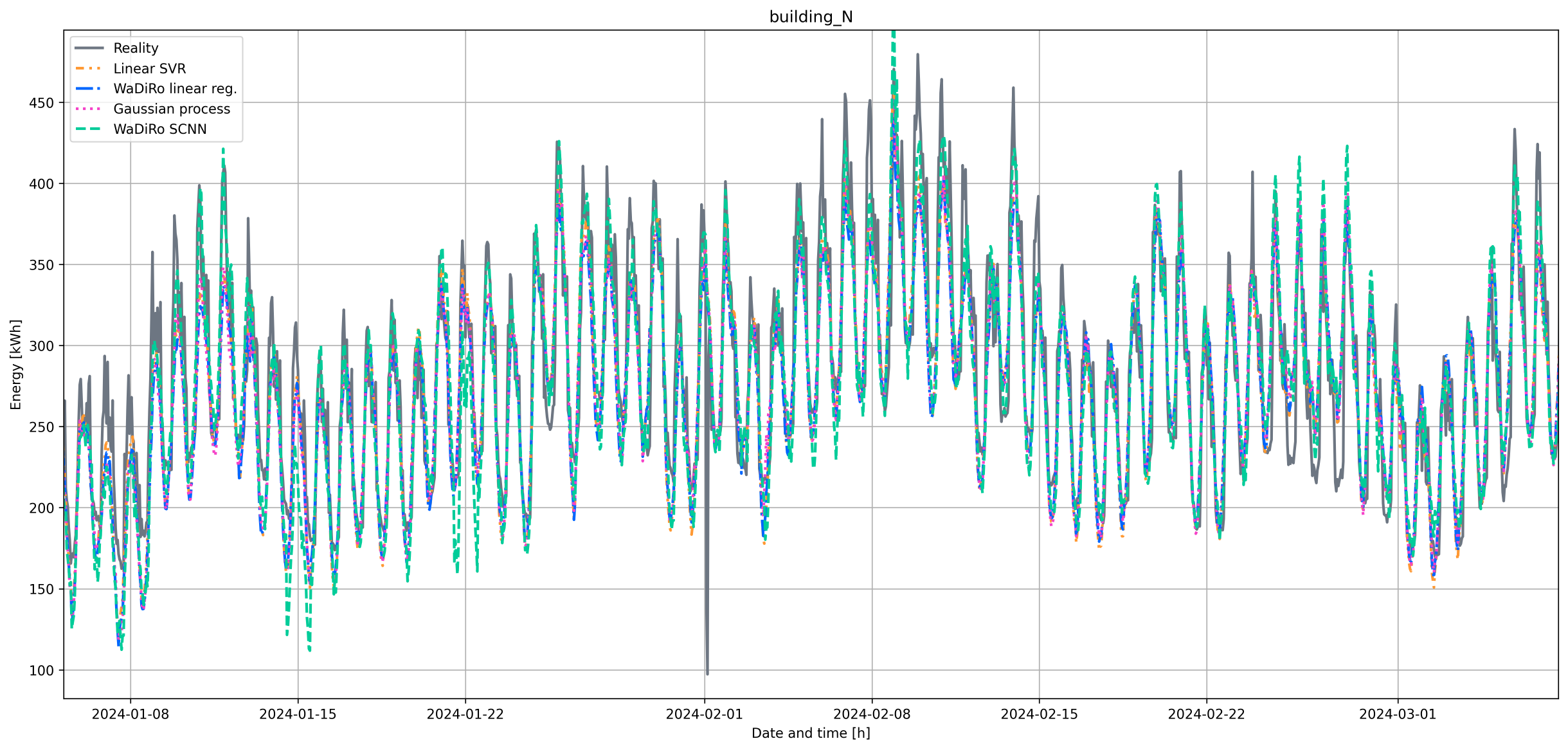}
        %\caption{Experiment C}
        %\label{fig:sub3}
    \end{subfigure}
    \caption{Winter predictions of the testing phase for four buildings}
    \label{fig:buildseries}
\end{figure}

In Figure \ref{fig:buildseries}, we present hourly winter predictions from the testing phase on a subset of buildings. \blue{This figure has qualitative purposes more than anything else and is better examined with colors.} We observe that applying models naively does not always work for every building, e.g., building~M, while it may perform quite well on others, e.g., building~N.

%\bibliographystyle{informs2014} % outcomment this and next line in Case 1
%\bibliography{sample} % if more than one, comma separated

% CASE 2: BiBTeX used to generate mypaper.bbl (to be further fine tuned)
%\input{mypaper.bbl} % outcomment this line in Case 2

%If you don't use BiBTex, you can manually itemize references as shown below.

%\bibliographystyle{nonumber}

%%%%%%%%%%%%%%%%%
\end{document}